\documentclass[a4paper,fleqn]{article}
\usepackage{a4wide}
\usepackage{amsmath,amssymb,amsthm}
\usepackage{thmtools}
\usepackage{hyperref}
\usepackage[utf8]{inputenc}
\usepackage[T1]{fontenc}
\usepackage{xifthen} 
\usepackage{xspace}
\usepackage{booktabs}
\usepackage{multirow}
\usepackage{graphicx}
\usepackage{rotating}
\usepackage{siunitx}

\usepackage[noend]{algpseudocode}
\usepackage{algorithm,algorithmicx}

\declaretheorem[numberwithin=section,refname={Theorem,Theorems},Refname={Theorem,Theorems}]{theorem}

\declaretheorem[sibling=theorem,style=definition,refname={Lemma,Lemmas},Refname={Lemma,Lemmas}]{lemma}

\newcommand{\signal}[2][]{#2\ifthenelse{\isempty{#1}}{}{(#1)}}
\newcommand{\sut}{\mathcal{M}}
\newcommand{\Z}{\mathbb{Z}}
\newcommand{\R}{\mathbb{R}}
\newcommand{\I}{\mathcal{I}}
\newcommand{\inputspace}{\mathcal{I}}
\newcommand{\outputspace}{\mathcal{O}}
\newcommand{\latentspace}{\mathcal{H}}
\newcommand{\generator}{\mathcal{G}}
\newcommand{\discriminator}{\mathcal{D}}

\newcommand{\always}{\square}
\newcommand{\eventually}{\lozenge}
\newcommand{\until}{\mathbin{\mathcal{U}}}

\newcommand{\pyes}{\ensuremath{\checkmark}}
\newcommand{\pno}{\ensuremath{\boldsymbol{\times}}}

\newcommand{\AFC}{\textup{AFC27}}
\newcommand{\ATI}{\textup{AT1}_{20}}
\newcommand{\ATXII}{\textup{ATX2}}
\newcommand{\ATVIA}{\textup{AT6}_{4,35,3000}}
\newcommand{\ATVIB}{\textup{AT6}_{8,50,3000}}
\newcommand{\ATVIC}{\textup{AT6}_{20,65,3000}}
\newcommand{\ATXVIA}{\textup{AT6}_{30,80,4500}}
\newcommand{\ATXVIB}{\textup{AT6}_{30,50,2700}}
\newcommand{\CCIII}{\textup{CC3}}
\newcommand{\CCIV}{\textup{CC4}}
\newcommand{\FA}{\textup{F16}}
\newcommand{\NN}{\textup{NN}}
\newcommand{\PM}{\textup{PM}}

\title{Requirement falsification for cyber-physical systems using generative models}

\author{
Jarkko Peltomäki\footnote{Corresponding author.} \hphantom{ }and Ivan Porres\\
Faculty of Science and Engineering, \\ Åbo Akademi University \\ Turku, Finland \\
\texttt{name.surname@abo.fi}
}

\begin{document}

\maketitle

\begin{abstract}
We present the OGAN algorithm for automatic requirement falsification of cyber-physical systems.
System inputs and outputs are represented as piecewise constant signals over time while requirements are expressed in signal temporal logic.
OGAN can find inputs that are counterexamples for the correctness of a system revealing design, software, or hardware defects before the system is taken into operation.
The OGAN algorithm works by training a generative machine learning model to produce such counterexamples. It executes tests offline and does not require any previous model of the system under test.
We evaluate OGAN using the ARCH-COMP benchmark problems, and the experimental results show that generative models are a viable method for requirement falsification. OGAN can be applied to new systems with little effort, has few requirements for the system under test, and exhibits state-of-the-art CPS falsification efficiency and effectiveness.
\end{abstract}

\section{Introduction}
Cyber-physical systems (CPS)~\cite{DBLP:conf/wcsp/ShiWYS11} integrate complex hardware and software components to interact autono\-mously with their physical environment in real time. Examples of CPS range from smart home systems to autonomous vehicles. Most CPS are considered safety-critical systems~\cite{DBLP:conf/icse/Knight02} since they may cause injuries, environmental damage, or economic losses if they fail. It is therefore paramount to assess the correctness of a CPS before it is taken into operation. This requires rigorous verification and validation methods to identify and remove potential defects.

In this article, we present a novel algorithm named OGAN for validation by requirement falsification~\cite{DBLP:journals/jair/CorsoMKLK21} based on generative machine learning and generative adversarial networks~\cite{goodfellow2014generative}. Requirement falsification searches for faults in the form of system inputs that yield outputs that violate a given system requirement. It can be understood as a form of runtime verification, i.e., search-based test design and execution guided by formal requirements \cite{DBLP:series/lncs/10457}.   Due to the real-time nature of CPS, execution traces are usually represented as signals over time, and requirements are expressed using  signal temporal logic (STL) \cite{maler2004monitoring}. A requirement falsification algorithm iteratively generates a system input, executes it on the system under test (SUT), observes the system output with a runtime monitor, and evaluates if the requirement is satisfied or not. If the algorithm finds an input for which the given requirement is not satisfied, then that input is a counterexample against the claim that the system is safe with respect to the requirement.

The main benefit of requirement falsification as a verification method is that it can be fully automated, while its main drawback is that it is sound but it is not complete, i.e., it can miss faults. Therefore research on requirement falsification focuses on algorithms that can find faults effectively and efficiently within a limited testing budget, given as a wall-clock time limit or a number of tests to be executed.

The search for faults in OGAN is driven by a robustness metric~\cite[Sec.~3.2]{DBLP:conf/rv/FainekosH019}. A robustness metric is a function from an execution trace (a pair of system inputs and outputs) to a single real value. We require that the robustness metric has the following two properties. First, it is zero if the trace does not satisfy the requirement. Second, if the requirement is satisfied, then the robustness is positive and indicates how close the trace is from being falsifying. These properties allow a search towards inputs that do not satisfy a given requirement.  Tools and algorithms using robustness metrics include~\cite{DBLP:conf/icse/MenghiNBP20, 2023:search_based_software_testing_driven, 2020:falsification_of_cyber_physical_systems_with_robustness, 2021:psy_taliro_a_python_toolbox_for_search, 2010:breach_a_toolbox_for_verification_and_parameter_synthesis, 2021:falsification_of_hybrid_systems_using_adaptive, staliro-tool-paper,2021:stochastic_optimization_with_adaptive_restart_a_framework}.

The novelty of the OGAN algorithm lies in learning two models: one model serves as a surrogate of the composition of the SUT and the robustness metric while the other serves as a query strategy for the surrogate model. The latter model is a generative model, the \emph{generator}, and it generates system inputs from random noise. The surrogate model, the \emph{discriminator}, models the system response to inputs and is used to predict the robustness of generated system inputs. We implement both models as neural networks and train them in a fashion inspired by Generative Adversarial Networks (GAN)~\cite{goodfellow2014generative}. The OGAN algorithm is generic since it applies to any deterministic black-box system with requirements specified in STL. 

OGAN differs from existing requirement falsification algorithms by using both a surrogate model and a generator model. The existing algorithms frequently use direct search and optimization to find falsifying inputs \cite{2021:effective_hybrid_system_falsification_using_monte,2010:breach_a_toolbox_for_verification_and_parameter_synthesis,2021:falsification_of_hybrid_systems_using_adaptive,staliro-tool-paper}, and they do not perform any modeling. Some approaches, such as ARIsTEO \cite{DBLP:conf/icse/MenghiNBP20} or methods that use Bayesian optimization \cite{2021:stochastic_optimization_with_adaptive_restart_a_framework,DBLP:journals/tecs/DeshmukhHJMP17}, refine a surrogate model, but they do not use generative modeling to find the falsifying input. Instead, they use direct search on the surrogate model. In OGAN, this direct search is replaced by the generator model, which learns to generate falsifying inputs.

OGAN is an online algorithm in the sense that it generates test inputs in a closed loop with the system under test. However, all the signals comprising a single test are generated offline, before the start of the test execution. This allows us to use OGAN in systems with strict real-time requirements (e.g. due to hardware-in-the-loop subsystems). In addition, OGAN does not require the use of previously trained models or data. OGAN learns tabula rasa, and the effort to do that is included in the effort required for the falsification task.

We evaluate the effectiveness and efficiency of OGAN experimentally on benchmarks derived from the falsification track of the ARCH-COMP competition \cite{ARCH21,ARCH22,ARCH23}, which is a friendly research competition on the verification of continuous and hybrid systems. We present evidence that OGAN's effectiveness and efficiency are on par with the state-of-the-art requirement falsification algorithms of the ARCH-COMP competition.

The OGAN algorithm has been used previously in the context of performance testing \cite{ogan} and the falsification of multiple requirements in combination with multi-armed bandits \cite{mab}. In this work, we provide for the first time a complete description of OGAN including all necessary theoretical considerations as well as a comprehensive evaluation of OGAN. The main contributions of this article, not present in \cite{ogan,mab},  are as follows:

\begin{itemize}
\item A requirement falsification algorithm based on generative models that uses offline test execution, supports arbitrary STL requirements, and does not require any previous model or dataset on the system. The hyperparameter setup of the described implementation provides competitive results on a wide range of problems.

\item A scaled robustness metric for STL. This is proposed to improve the representation of STL robustness in the neural network models used by OGAN.

\item A novel methodology to compare the effectiveness and efficiency of requirement falsification algorithms based on survival analysis.

\item A comprehensive evaluation of the OGAN algorithm. This includes evaluating the discriminator accuracy and the generator Monte Carlo sampling strategy.
\end{itemize}

We proceed as follows. \autoref{sec:problem_description} describes the problem of requirement falsification and presents the overall idea on how to use generative machine learning to solve it. We then present the OGAN algorithm as a concrete realization of these ideas in \autoref{sec:ogan}. \autoref{sec:fr} presents how system inputs and outputs are represented by OGAN and its machine learning models, including how to automatically obtain scaled robustness metrics for STL requirements. We then proceed to the evaluation of OGAN. The main research questions and the methodology to answer them are found in \autoref{sec:evaluation_methodology} while the actual experimental results are discussed in \autoref{sec:results}. We conclude by presenting related work (\autoref{sec:related_work}) and conclusions (\autoref{sec:conclusions}).

\section{Problem Description}\label{sec:problem_description}

\subsection{System Under Test}\label{ssec:sut}
We aim to analyze cyber-physical systems as diverse as the automatic transmission of a car~\cite{2015:benchmarks_for_temporal_logic_requirements_for_automotive}, the ground collision avoidance system of a fighter jet~\cite{2018:verification_challenges_in_f16_ground_collision_avoidance}, and a pacemaker~\cite{2022:two_simulink_models_with_requirements_for_a_simple}. To this end, we need an abstraction that is general enough to describe many different systems.

We view the system under test (SUT) as a deterministic function $\sut$ from an input space $\inputspace$ to an output space $\outputspace$. We assume that the spaces $\inputspace$ and $\outputspace$ are either compact subsets of a Euclidean vector space or function spaces containing signals $[0, T] \to \R^n$ for some $T > 0$. For convenience, we represent vector inputs as constant signals.

For example, the inputs in the automatic transmission benchmark are a vehicle's throttle and brake signals over a time interval of $30 \, \textup{s}$ taking values respectively in $[0, 100]$ and $[0, 325]$. In the ground collision avoidance system benchmark, the input is a vector of length three representing the initial roll, pitch, and yaw of the jet. The output is a signal describing the altitude of the jet during the $15 \, \textup{s}$ simulation.

\subsection{System Requirements}\label{sec:stl}
The signal temporal logic (STL) \cite{maler2004monitoring} allows us to express complex requirements unambiguously. STL is used to express the requirements for the benchmarks selected for the evaluation of the OGAN algorithm (see \autoref{sec:results}). 

For example, given a signal $\textup{SPEED}\colon [0,30] \to [0,\infty)$, the STL formula
\begin{equation*}
  \always_{[0,20]} \, \textup{SPEED} < 120
\end{equation*}
expresses that the signal should never exceed $120$ during the first $20$ time units. 

A brief definition of STL follows (see \cite{maler2004monitoring} for more details). Let $\mathbb{T}$ be a discrete-time domain, that is, $\mathbb{T} = [A, B] \cap \Z$ with $A, B \in \Z$ and $A < B$. In
what follows, we simply write $[A, B]$ for $[A, B] \cap \Z$. A signal $s$ is a function $s\colon \mathbb{T} \to \R$.
For convenience, we represent several signals with a common time domain as a single vector-valued signal $\mathbf{s}$.
Let $P = \{p_1, \ldots, p_n\}$ be a set of predicates
$p_i\colon \mathbb{T} \to \{\top, \bot\}$, $i = 1, \ldots, n$, whose truth value depends on both time
$t$ and a signal $\mathbf{s}_i$. The grammar of the \emph{signal temporal logic} (STL) is defined recursively by
\begin{equation*}
  \varphi := p \, | \, \lnot \varphi \, | \, \varphi \land \psi \, | \, \varphi \until_{\I} \psi
\end{equation*}
where $\varphi$, $\psi$ are STL formulas, $p \in P$, and $\I$ is an interval in $\mathbb{T}$ (open, closed, or
half-open). As usual, we let $\varphi \lor \psi$ be shorthand for
$\lnot(\lnot \varphi \land \lnot \psi)$ and $\varphi \rightarrow \psi$ for $\lnot \varphi \lor \psi$. In addition, we
define the temporal operators \emph{eventually} $\eventually_\I \varphi := \top \until_{\I} \varphi$ (here $\top$ is a
predicate which always evaluates to true) and \emph{always} $\always_{\I} \varphi := \lnot \eventually_\I \lnot \varphi$. The
operator $\until_\I$ is called the \emph{until} operator.

We write $(\mathbf{s},t) \models \varphi$ to indicate that a signal $\mathbf{s}$ satisfies the formula $\varphi$ at
time $t$. This relation is defined inductively as follows:
\begin{alignat*}{4}
  (\mathbf{s},t) &\models p_{i,\mathbf{s}_i}     &&\iff p_{i,\mathbf{s}}(t) = \top, \\
  (\mathbf{s},t) &\models \lnot \varphi          &&\iff (s,t) \not\models \varphi, \\
  (\mathbf{s},t) &\models \varphi \land \psi     &&\iff (s,t) \models \varphi \text{ and }  (s,t) \models \psi, \quad \text{and} \\
  (\mathbf{s},t) &\models \varphi \until_\I \psi &&\iff \exists t' \in t + \I\colon (\mathbf{s}, t'') \models \varphi \enspace \forall t'' \in [t, t'] \text{ and } (\mathbf{s}, t') \models \psi.
\end{alignat*}
The interpretation for the until operator is that the formula $\varphi$ evaluates to be true until the formula $\psi$
evaluates to true. Notice that $\varphi$ needs to evaluate to true from the time point $t$ which might be outside of $t
+ \I$. The interpretations of the remaining operators are obvious.

It follows from above that
\begin{alignat*}{4}
  (\mathbf{s},t) &\models \eventually_\I \varphi &&\iff \exists t' \in t + \I\colon (\mathbf{s}, t') \models \varphi \quad \text{and} \\
  (\mathbf{s},t) &\models \always_\I \varphi &&\iff \forall t' \in t + \I\colon (\mathbf{s}, t') \models \varphi.
\end{alignat*}
Above, we silently assume that the signal $\mathbf{s}$ is long enough so that it makes sense to talk about the signal
value at a time $t$, that is, we assume that $t + \I \subseteq \mathbb{T}$.

\subsection{Falsification of Formal Requirements}\label{ssec:falsification_of_stl}
Let $\sut\colon \inputspace \to \outputspace$ be a SUT. A \emph{falsification problem} for a requirement represented as a formula $\varphi$ means producing an input (a test) $x$ in $\inputspace$ such that the behavior of the SUT for $x$ violates $\varphi$. If such an $x$ is found, the claim that system $\sut$ is correct with respect to $\varphi$ is falsified with the input $x$ as a counterexample.

More precisely, the falsification problem is to find
\begin{alignat*}{1}
 x \in \inputspace \text{ such that } (\sut(x), 0) \not\models \varphi.
\end{alignat*}
Here $(\sut(x), 0) \not\models \varphi$ means that the signal $\sut(x)$ does not satisfy $\varphi$ when evaluated at time $0$. We abuse the notation and consider $\sut(x)$ to contain both output and input signals, that is, $\sut(x)$ is the execution trace. This is needed as a requirement might involve both inputs and outputs. We evaluate the truth value at time $0$ simply for convenience, and the discussion is easily adjusted to cover a more general setting. 

\subsection{Robustness-Based Falsification}\label{sec:falsification_robustness}
Our central assumption is that the SUT is a black-box function meaning that we cannot solve for falsifying inputs, and all we can do is to search for them. We assume that associated to a requirement $\varphi$ there is a \emph{scaled robustness metric} $\overline{\rho}$ taking values in the interval $[0,1]$. The scaled robustness metric has the following properties:
\begin{align*}
\overline{\rho}(\sut(x)) > 0 &\text{ if } (\sut(x), 0) \models \varphi,  \\
\overline{\rho}(\sut(x)) = 0 &\text{ if } (\sut(x), 0) \not\models \varphi.
\end{align*}
Moreover, we also expect $\overline{\rho}$ to have the following property: the smaller the values of $\overline{\rho}(\sut(x))$ are, the closer the input $x$ is to falsifying $\varphi$. This property will be made precise in \autoref{ssec:traditional_stl_robustness} when we introduce a scaled STL robustness metric in detail.

We can now define the \emph{robustness-based falsification problem} as follows: given a scaled robustness metric $\overline{\rho}$, find
\begin{alignat*}{1}
 x \in \inputspace \text{ such that } \overline{\rho}(\sut(x)) = 0.
\end{alignat*}

\subsection{Falsification as a Generative Machine Learning Problem }\label{ssec:falsification_generative_models}
We propose to solve the robustness-based falsification problem by learning a model to generate falsifying inputs. For this, we define $\mathcal{F}$ as the set of falsifying inputs, that is, let
\begin{equation*}
  \mathcal{F} = \{x \in \inputspace : \overline{\rho}(\sut(x)) = 0\}.
\end{equation*}

\begin{figure}
\includegraphics[width=\columnwidth]{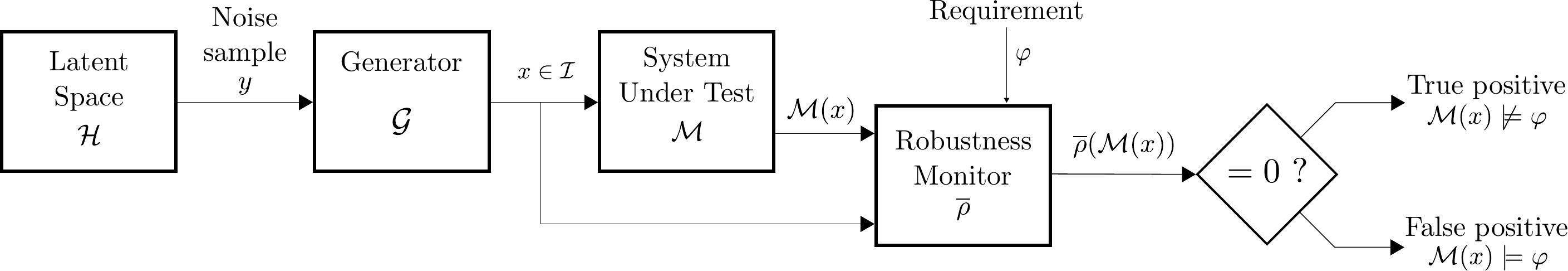}
\caption{Overview of falsification using a generative machine learning. A generator model is used to sample inputs to execute against the SUT.}\label{fig:overview1}
\end{figure}

We can then solve the falsification problem by training a generator $\generator\colon \latentspace \to \inputspace$ which maps the uniform distribution on a latent space $\latentspace$ to the uniform distribution supported on $\mathcal{F}$ (see \autoref{ssec:ogan_setup} for a concrete choice for $\latentspace$). That is, we propose to learn a function $\generator$ that transforms random noise into falsifying inputs, so that $\overline{\rho}(\sut(\generator(y))) = 0$ for all $y \in \latentspace$. This can be achieved by representing the generator as a neural network and training its weights so that it outputs tests with robustness $0$. This process is illustrated in \autoref{fig:overview1}. We present a concrete algorithm to achieve this in the next section.

\section{The OGAN Algorithm for Robustness-Based Falsification Using Generative Models}\label{sec:ogan}

\subsection{Sketch of the Proposed Solution}
As discussed in the previous section, we aim to train the generator $\generator$ to generate falsifying tests. However, we do not possess a dataset of falsifying inputs before attempting to solve a falsification problem (and having such a dataset would render falsification pointless). We address this issue by introducing a second model called the \emph{discriminator}\footnote{The word discriminator is used here because our research is inspired by generative adversarial networks. However, the discriminator defined here does not play the same role as discriminators in generative adversarial networks.} $\discriminator\colon \inputspace \to \mathbb{R}$
which we train to learn the mapping $x \mapsto \overline{\rho}(\sut(x))$, $x \in \inputspace$. That is, we use $\discriminator$ as a proxy for the robustness value of the system trace obtained by executing the input $x$.

\begin{figure}
\includegraphics[width=\columnwidth]{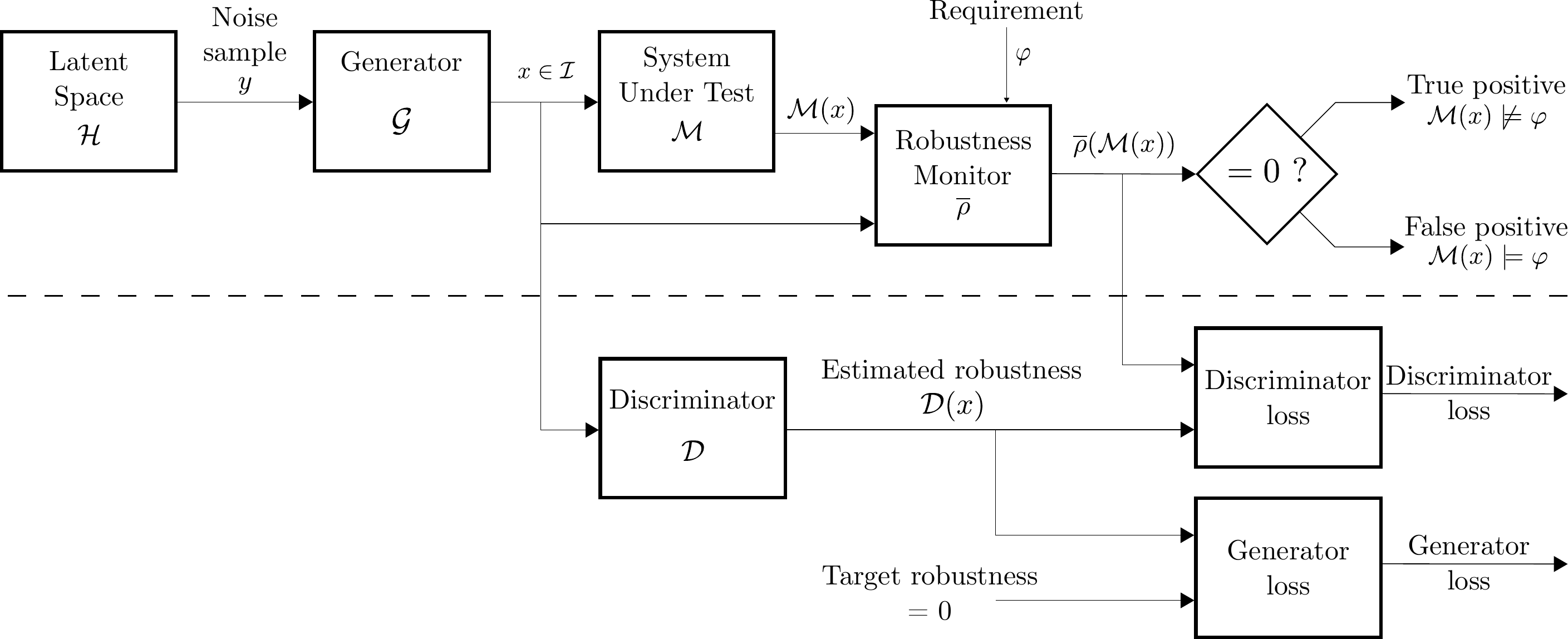}
\caption{Training a generative model for falsification. The discriminator predicts the robustness of inputs sampled from the generator. The generator is trained with a loss function to generate tests that have estimated target robustness $0$.}\label{fig:overview2}
\end{figure}

If we assume that there is such a discriminator $\discriminator$, we can train
the generator $\generator$ by minimizing
\begin{equation}\label{eq:loss}
  \frac{1}{|\mathcal{B}|} \sum_{y \in \mathcal{B}} \mathcal{L}(\discriminator(\generator(y)), 0)
\end{equation}
for a loss function $\mathcal{L}$ and a batch $\mathcal{B}$ of elements (noise) sampled from $\latentspace$. We illustrate this process in \autoref{fig:overview2}. We remark that, in order for the training with this loss to be successful, the robustness metric must have the properties listed in \autoref{sec:falsification_robustness}.

Since we wish to avoid using any precollected data or pretrained models, we do not have a priori a discriminator $\discriminator$, so we need to train it online. We propose to approach this as follows. First, we obtain a small amount of initial training data. This can be done by uniform random sampling or a similar Monte-Carlo method. Based on this initial data, a discriminator $\discriminator$ can be trained and consequently the generator $\generator$ can be trained, too, as discussed above. In practice, it is unlikely that this yields a good generator especially if the initial training data is small or contains few tests with low robustness. It is therefore crucial to augment the discriminator's training data. For the augmentation, there are essentially two approaches: 1) continue the Monte-Carlo sampling, 2) sample the generator $\generator$ for tests. The options can be even combined by alternating between them.

Let us then discuss the latter approach of using $\generator$ for training data augmentation. Once $\generator$ has been trained against $\discriminator$, we can sample $\generator$ for tests by sampling the uniform distribution on the latent space as described in \autoref{ssec:falsification_generative_models}. Let $x$ be such a test, and execute it on the SUT to learn $\sut(x)$. If $\overline{\rho}(\sut(x)) = 0$ for several tests $x$, we can consider that the required generator has been found. If this is not the case, then we can improve the discriminator's accuracy by introducing pairs $(x, \overline{\rho}(\sut(x)))$ into its training data. The generator $\generator$ can then be retrained on the improved discriminator, and the process can be repeated.

Intuitively, the process described above eventually leads to finding a generator $\generator$ that has the desired properties. In practice, the process can get stuck due to local minima. However, we may still be able to sample at least one element from $\mathcal{F}$ during the training, and this suffices for falsification.

Above, we focused on using the partially trained generator as a mechanism to obtain more training data for the discriminator. As stated previously, this can also be done using Monte-Carlo sampling. However, we expect that using the generator is more efficient. An input sampled from $\generator$ with high robustness provides $\discriminator$ immediate feedback about its inaccuracies whereas a test obtained via Monte-Carlo sampling can be unhelpful in improving $\discriminator$.

\subsection{The OGAN Algorithm}
The pseudo code for our proposal to implement the ideas of \autoref{ssec:falsification_generative_models} is given in \autoref{alg1}. The code omits many practical details like the neural network structure and the loss function. These choices are made concrete in \autoref{ssec:ogan_setup} where the setup for our evaluation of OGAN is described. 

\begin{algorithm}[t]
  \caption{OGAN requirement falsification algorithm.}\label{alg1}
  \begin{algorithmic}[1]
    \Require{SUT $\sut$, execution budget $E$, number $N$ of tests to be sampled initially, sampling probability $P$, batch size $B$, discriminator training epochs $E_\discriminator$, generator training epochs $E_\generator$, multiplier $\alpha$.}
    \Statex
    \State $T \gets$ \Call{sample}{$\inputspace$; $N$}
    \State $F \gets $ \Call{execute}{$\sut; T$}
    \While{$\lnot$ stop condition}
      \State $\generator, \discriminator \gets$ \Call{initialize\_ogan\_models}{\! }
      \For{$i \in \{1, \ldots, E_\discriminator\}$}
        \State \Call{train\_discriminator}{$\discriminator; T, F$} \vspace{-0.2em}
      \EndFor
      \For{$i \in \{1, \ldots, E_\generator\}$}
        \State $X \gets$ \Call{sample}{$\latentspace; B$}
        \State \Call{train\_generator}{$\generator; \discriminator(\generator(X)), 0$} \vspace{-0.2em}
\EndFor
      \State test $\gets$ \Call{sample\_test}{$\generator, \discriminator; P$}
      \State test\_robustness $\gets$ \Call{execute}{$\sut;$ test}
      \State $T \gets T \, \cup \, \{$test$\}$
      \State $F \gets F \, \cup \, \{$test\_robustness$\}$
    \EndWhile
    \Statex
    \Function{sample\_test}{$\generator, \discriminator; P$}
        \If{\Call{random}{\! } $\leq P$}
            \State test $\gets$ \Call{sample}{$\inputspace; 1$}
        \Else
            \State $Q \gets$ \Call{initialize\_priority\_queue}{\! }
            \State accept\_threshold $\gets 0$
            \Repeat
                \State candidate\_test $\gets$ \Call{sample}{$\generator; 1$}
                \State estimated\_robustness $\gets$ \Call{estimate\_robustness}{$\discriminator; $ candidate\_test}
                \State \Call{add\_to\_queue}{$Q; $ estimated\_robustness, candidate\_test}
                \State accept\_threshold $\gets 1 - \alpha \cdot (1 -$ accept\_threshold$)$
           \Until{\Call{min}{$Q$} $\leq$ accept\_threshold}
           \State test $\gets$ \Call{test\_with\_min\_robustness}{$Q$}
        \EndIf \vspace{-0.3em}
        \State \Return test
    \EndFunction
\end{algorithmic}
\end{algorithm}

Let us describe the pseudo code in detail. On line $1$, the test suite $T$ is populated by a sample of size $N$ via the chosen Monte-Carlo sampling algorithm. For example, uniform random sampling on $\latentspace$ can be used for this purpose. Line $2$ executes the tests on the SUT, and saves the corresponding scaled robustness function values into the structure $F$ (OGAN uses the SUT output only to compute the robustness). Line $3$ begins the main loop of the algorithm, and its execution resumes until the execution budget (number of allowed executions) has been exhausted. On lines $4$--$9$, the models $\generator$ and $\discriminator$ are initialized and trained. The model $\discriminator$ is trained using the training data $(T, F)$ collected so far for a certain amount of training epochs $E_\discriminator$. Lines $8$--$9$ correspond to training $\generator$ against $\discriminator$, as in \eqref{eq:loss}, for a certain number of epochs $E_\generator$. On lines $10$--$13$, a new test is sampled, executed, and it and its robustness are added to the discriminator's training data. After the algorithm has executed, the test suite $T$ can be inspected for falsifying tests.

Let us next look at the sampling of a new test, i.e., the function on line $14$. This function samples a new test via Monte-Carlo sampling with probability $P$; otherwise $\generator$ is sampled (see below). The reason for this is the following. If the falsification of the requirement is of unknown difficulty, then it is not obvious how to set the initial random sampling budget $N$. If the problem is easy, then a small $N$ should suffice for OGAN to falsify, and setting $N$ too large can use an excessive amount of executions before a falsification is observed. If the problem is not easy, then a larger $N$ can help the discriminator to learn the mapping better globally. Monte-Carlo sampling with probability $P$ after the initial phase allows $N$ to be small: tests independent of the generator will be executed throughout the whole execution of the algorithm. This probability $P$ thus strikes a balance between exploration and exploitation. More sophisticated sampling strategies based on, e.g., multi-armed bandits, could be used here, but we have not explored this.

Let us finally describe the test sampling using $\generator$, which corresponds to lines $18$--$26$ of \autoref{alg1}. The loop starting on line $20$ first samples $\generator$ (i.e., samples the uniform distribution on $\latentspace$ and maps the samples via $\generator$) for a test. Then it estimates its robustness using $\discriminator$. This is a crucial step as this avoids expensive execution of tests on the SUT. If the estimated robustness is low enough, then the test is accepted and returned. We sample many tests and check them against an acceptance threshold because, due to the random nature of the sampling, we are not guaranteed to immediately sample a test with the lowest possible estimated robustness. The acceptance threshold is progressively increased on each round of the loop because i) we want to accept tests reasonably fast, ii) the estimates of $\discriminator$ might have a large lower bound. The threshold increases from $0$ to $1$ like $1$ decreases to $0$ when repeatedly multiplied by the number $\alpha$. We use a priority queue $Q$ to select the candidate with the lowest estimated robustness.

We can use different criteria as the stop condition for the main OGAN loop, such as stopping when the first falsifying input is found, stopping after a fixed wall time has elapsed, or after a number of tests have been executed, or a combination of these criteria. We expect that a falsifying test is executed on the SUT during the execution of OGAN. There are no theoretical guarantees for this, but the evaluation of OGAN presented in \autoref{sec:results} shows that this happens in practice.

\section{Feature Representation}\label{sec:fr}
In this section, we discuss how we normalize and represent system inputs and the robustness of system outputs since machine learning models require inputs and outputs to be given in an explicit form. More importantly, it is often recommended to normalize the inputs of machine learning models as it improves convergence. Indeed, in the particular case of neural networks, with constant-sign inputs and activation functions, the convergence, while not impossible, can be very slow
\cite{2012:efficient_backprop}. For us, convergence is especially important as the training data is obtained on the fly and new training data is collected using models trained with little data. An additional benefit of normalization is that the generative models can be agnostic about the actual ranges of the system inputs and outputs.

\subsection{Representation of System Inputs}
We represent system inputs as piecewise constant signals of equal piece length. Such a one-dimensional signal consisting of $D$ pieces, taking values in an interval $[A, B]$, $A < B$, can be fully described by $D$ numbers. We map these numbers to the interval $[-1, 1]$ using the linear map $x \mapsto (-2x + A + B)/(A - B)$. We thus take the input space  $\I$ to be $[-1, 1]^D$ and map its vectors to system inputs using the inverse transformation. If the input is a vector, not a signal, then we do the same but omit the sampling of the inverse images into a signal. In the case of a multidimensional signal, we simply concatenate the vectors representing each signal.

We have assumed here that the signal ranges (which can be different between components) are known in advance. Ranges must be specified as OGAN cannot be used otherwise. 

\subsection{Representation of Requirement Robustness}
The OGAN algorithm uses system outputs to compute a robustness metric normalized to the interval $[0,1]$. The rest of this section describes how the normalization is achieved for STL requirements.

\subsubsection{Traditional STL Robustness Metric}\label{ssec:traditional_stl_robustness}
Our  aim is to transform an STL formula $\varphi$ into a real-valued function $\rho$ such that the following
implications are satisfied:

\begin{align*}
  \rho(\varphi; \mathbf{s}, t) > 0 &\Longrightarrow (\mathbf{s}, t) \models \varphi \Longrightarrow \rho(\varphi; \mathbf{s}, t) \geq 0, \\
  \rho(\varphi; \mathbf{s}, t) < 0 &\Longrightarrow (\mathbf{s}, t) \not\models \varphi \Longrightarrow \rho(\varphi; \mathbf{s}, t) \leq 0.
\end{align*}
In other words, the sign of $\rho(\varphi; \mathbf{s}, t)$ determines the truth value of $\varphi$ whenever
$\rho(\varphi; \mathbf{s}, t) \neq 0$. There are infinitely many ways to accomplish this, but we focus here on
what we call the traditional robustness metric \cite{donze2010robust}. Other alternative STL robustness metrics can be found in~\cite{2020:on_robustness_metrics_for_learning_stl_tasks,2021:a_smooth_robustness_measure_of_signal_temporal,2021:effective_hybrid_system_falsification_using_monte}. 

First of all, we assume that a robustness metric $\rho$ is given when $\varphi = p$ for a predicate $p$. In our
applications, the predicates take the form of a simple inequality like $\textup{SPEED} \leq 120$. We set
\begin{equation*}
  \rho(X \geq Y; \mathbf{s}, t) = \rho(X; \mathbf{s}, t) - \rho(Y; \mathbf{s}, t),
\end{equation*}
and we let
\begin{equation*}
  \rho(v; \mathbf{s}, t) = v(t)
\end{equation*}
whenever $v$ refers to a component of $\mathbf{s}$. We interpret constants in the inequalities as signals with constant
value. Our approach cannot distinguish the predicates $X \geq Y$ and $X > Y$ in the case of equality. In order to work
around this, we can set $\rho(X \geq Y)$ to be a nonzero constant with appropriate sign in the case of equality. When
$X$ and $Y$ refer to continuous quantities, the problem with the value $0$ does not often matter in practice. If $X = Y$,
then the requirements $X > Y$ and $X \geq Y$ are almost falsified as even the tiniest
perturbation of the signals leads to falsification. It is thus natural to count both requirements as falsified.

The robustness metric for more complex STL formulas are found as follows:
\begin{align}\label{eq:trad_robustness}
  \rho(\lnot \varphi; \mathbf{s}, t)          &= -\rho(\varphi; \mathbf{s}, t), \nonumber \\
  \rho(\varphi \land \psi; \mathbf{s}, t)      &= \min\{\rho(\varphi; \mathbf{s}, t), \rho(\psi; \mathbf{s}, t)\}, \quad \text{and} \nonumber \\
  \rho(\varphi \until_\I \psi; \mathbf{s}, t) &= \max_{t' \in t + \I}\left( \min\left\{ \rho(\psi; \mathbf{s}, t'), \min_{t'' \in [t, t')} \rho(\varphi; \mathbf{s}, t'') \right\} \right).
\end{align}
From this, we deduce that
\begin{align*}
  \rho(\eventually_\I \varphi; \mathbf{s}, t) &= \max_{t' \in t + \I} \rho(\varphi; \mathbf{s}, t') \quad \text{and} \\
  \rho(\always_\I \varphi; \mathbf{s}, t)     &= \min_{t' \in t + \I} \rho(\varphi; \mathbf{s}, t').
\end{align*}
Notice that given $\varphi$, $\mathbf{s}$, and $t$, the function $\rho$ can be efficiently computed.

It is straightforward to verify that the robustness metric is computed as follows when $\varphi = \always_{[0,20]} \, \textup{SPEED} < 120$:
\begin{equation*}
  \rho(\varphi; \textup{SPEED}, t) = \min_{t' \in [t, t+20]} \left( 120 - \textup{SPEED}(t') \right).
\end{equation*}

It is an easy exercise to prove that $(\mathbf{s}, t) \models \varphi$ implies that $\rho(\varphi; \mathbf{s}, t) \geq 0$
(whenever the robustness of the predicates satisfies this property as well) and conversely that
$\rho(\varphi; \mathbf{s}, t) > 0$ implies that $(\mathbf{s}, t) \models \varphi$ \cite{donze2010robust,2009:robustness_of_temporal_logic_specifications_for_continuous}. Analogous statements apply for the
opposite-signed statements. Moreover, it is straightforward to see that the following result is true (see \cite[Sect.~3.4]{2009:robustness_of_temporal_logic_specifications_for_continuous}).\\

\begin{lemma}
  Let $\pi_i(\mathbf{v})$ be the projection of a vector $\mathbf{v}$ to its $i$th coordinate. If
  $\rho(\varphi; \mathbf{s}, t) = \varepsilon$ and $\mathbf{s}'$ is a signal such that
  $|\pi_i(\mathbf{s}) - \pi_i(\mathbf{s'})| < |\varepsilon|$ for all $i$, then $(\mathbf{s}, t) \models \varphi$ if and
  only if $(\mathbf{s}', t) \models \varphi$ given that this property is true for predicates.
\end{lemma}
\vspace{1em}

The preceding lemma shows why the name robustness metric was chosen: if $\rho(\varphi; \mathbf{s}, t)$ is large in
absolute value, then the truth value is robust to large changes in the signal $\mathbf{s}$.

\subsubsection{Scaled STL Robustness Metric}\label{sssec:effective_range}
In this section, we study how to scale the traditional robustness metric $\rho$.
If $\I$ is an interval, then we denote respectively by $\I^-$ and $\I^+$ its left and right endpoints.

In what follows, we define the \emph{effective range} $\I(\varphi; \mathbf{s}, t)$ of an STL formula $\varphi$ for a
signal $\mathbf{s}$ at time $t$. Once such a range is defined, we obtain a \emph{scaled robustness metric}
$\overline{\rho}$, taking values in $[0, 1]$, by setting
\begin{equation*}
  \overline{\rho}(\varphi; \mathbf{s}, t) = \begin{cases}
                                                0,                                                             & \text{if $\rho(\varphi; \mathbf{s}, t) \leq 0$}, \\
                                                \rho(\varphi; \mathbf{s}, t) / \I^+(\varphi; \mathbf{s}, t), & \text{otherwise}.
                                              \end{cases}
\end{equation*}
Notice that the value $0$ does not strictly imply falsification here since it is possible that
$\rho(\varphi; \mathbf{s}, t) = 0$. Observe that the effective range of $\varphi$ depends also on the signal
$\mathbf{s}$ and time $t$ not only on the ranges of the respective signals.

Suppose that $\mathbf{s}$ and $\mathbf{s}'$ are two signals and that $\overline{\rho}(\varphi; \mathbf{s}, t) \leq \overline{\rho}(\varphi; \mathbf{s}', t)$. Then it is not necessarily true that $\rho(\varphi; \mathbf{s}, t) \leq \rho(\varphi; \mathbf{s}', t)$, that is, the signal $\mathbf{s}$ is not closer to falsifying $\varphi$ than the signal $\mathbf{s}'$ in the absolute sense. It is, however, closer to falsifying in the proportional sense.

Let $\pi_i(\mathbf{v})$ be the projection of a vector $\mathbf{v}$ to its $i$th coordinate. We assume that, for all
$i$, there exists two real-valued mappings $t \mapsto \I_i^-(t)$ and $t \mapsto \I_i^+(t)$ such that
$\pi_i(\mathbf{s}(t)) \in [\I_i^-(t), \I_i^+(t)]$ for all $t \in \mathbb{T}$. In practice, we fix a range $[A_i, B_i]$
for the $i$th signal component, that is, we assume that the mappings $\I_i^-$ and $\I_i^+$ are constant.

Our aim is to build mappings $\I^-, \I^+\colon \mathbb{T} \to \R$ such that
$\rho(\varphi; \mathbf{s}, t) \in [\I^-(\varphi; t), \I^+(\varphi; t)]$ for all $t \in \mathbb{T}$. We then define the
effective range $\I(\varphi; \mathbf{s}, t)$ of an STL formula $\varphi$ to be $[\I^-(\varphi; t), \I^+(\varphi; t)]$.

Let $\I^-, \I^+\colon \mathbb{T} \to \R$ be functions such that
$\rho(\varphi; \mathbf{s}, t) \in [\I^-(\varphi; t), \I^+(\varphi; t)]$ for all $t \in \mathbb{T}$. For our
applications, we need to discuss how to define $\I^-$ and $\I^+$ appropriately for simple inequalities $X \geq Y$. Here
we simply set
\begin{align*}
  \I^-(X \geq Y; \mathbf{s}, t) &= \I^-(X; \mathbf{s}, t) - \I^+(Y; \mathbf{s}, t) \quad \text{and} \\
  \I^+(X \geq Y; \mathbf{s}, t) &= \I^+(X; \mathbf{s}, t) - \I^-(Y; \mathbf{s}, t).
\end{align*}
This obviously satisfies our requirements.

Let us then describe how to define $\I^-$ and $\I^+$ for general STL formulas. It is natural to set
\begin{align*}
  \I^-(\lnot \varphi; \mathbf{s}, t) &= -\I^+(\varphi; \mathbf{s}, t) \quad \text{and} \\
  \I^+(\lnot \varphi; \mathbf{s}, t) &= -\I^-(\varphi; \mathbf{s}, t).
\end{align*}
It is clear that this achieves our aim for the negation $\lnot \varphi$.

Let us then consider the until operator and the formula $\varphi \until_\I \psi$. Inspired by
\eqref{eq:trad_robustness}, let $h(t')$ be the least integer in $[t, t')$ such that
\begin{equation*}
  \rho(\varphi; \mathbf{s}, h(t')) = \min_{t'' \in [t,t')} \rho(\varphi; \mathbf{s}, t'').
\end{equation*}
Similarly $u(t)$ be the least integer in $t + \I$ such that
\begin{equation*}
  \rho(\varphi \until_\I \psi; \mathbf{s}, t) = \min\left\{ \rho(\psi; \mathbf{s}, u(t)), \rho(\varphi; \mathbf{s}, h(u(t)) \right\}.
\end{equation*}
If $\rho(\psi; \mathbf{s}, u(t)) \leq \rho(\varphi; \mathbf{s}, h(u(t)))$, then we set
\begin{equation*}
  \I^-(\varphi \until_\I \psi; \mathbf{s}, t) = \I^-(\psi; \mathbf{s}, u(t)) \quad \text{and} \quad \I^-(\varphi \until_\I \psi; \mathbf{s}, t) = \I^-(\varphi; \mathbf{s}, h(u(t)))
\end{equation*}
otherwise. $\I^+(\varphi \until_\I \psi; \mathbf{s}, t)$ is defined analogously by substituting $\I^-$ for $\I^+$. With
the convention that the effective range for the true signal $\top$ is $[\infty, \infty]$, we obtain that
\begin{equation*}
  \I^-(\eventually_\I \varphi; \mathbf{s}, t) = \I^-(\varphi; \mathbf{s}, v_\lozenge(t)) \quad \text{and} \quad \I^+(\eventually_\I \varphi; \mathbf{s}, t) = \I^+(\varphi; \mathbf{s}, v_\lozenge(t))
\end{equation*}
where $v_\lozenge(t)$ is is the least integer in $t + \I$ such that
$\rho(\eventually_\I \varphi; \mathbf{s}, t) = \rho(\varphi; \mathbf{s}, v_\lozenge(t))$. Analogously
\begin{equation*}
  \I^-(\always_\I \varphi; \mathbf{s}, t) = \I^-(\varphi; \mathbf{s}, v_\square(t)) \quad \text{and} \quad \I^+(\always_\I \varphi; \mathbf{s}, t) = \I^+(\varphi; \mathbf{s}, v_\square(t))
\end{equation*}
where $v_\square(t)$ is is the least integer in $t + \I$ such that
$\rho(\always_\I \varphi; \mathbf{s}, t) = \rho(\varphi; \mathbf{s}, v_\square(t))$.

Consider finally the conjunction $\varphi \land \psi$ of two formulas. Let $\zeta$ be the formula such that $\rho(\varphi \land \psi; \mathbf{s}, t) = \rho(\zeta; \mathbf{s}, t)$ (in the case of equal robustness, we make an arbitrary but deterministic choice). We define
\begin{equation*}
  \I^-(\varphi \land \psi; \mathbf{s}, t) = \I^-(\zeta; \mathbf{s}, t)
\end{equation*}
and similarly for $\I^+$.

Having defined $\I^-$ and $\I^+$ like above, it is clear that we have constructed the required mappings $\I^-$ and
$\I^+$ for an arbitrary STL formula $\varphi$.

\subsubsection{An Example on Scaling}\label{sssec:example}
Let us take a look at the formula
\begin{equation*}
  \varphi := \always_{[0,10]} (\textup{SPEED} < 50) \lor \eventually_{[0,30]} (\textup{RPM} > 2700),
\end{equation*}
and suppose that $\textup{SPEED}(t) \in [0, 120]$ and $\textup{RPM}(t) \in [0, 4800]$ for all $t$. Let $\mathbf{s}$ be
the vector-valued signal $t \mapsto (\textup{SPEED}(t), \textup{RPM}(t))$. The range of
$\rho(\textup{SPEED} < 50; \mathbf{s}, t)$ is $[-70, 50]$ and the range of
$\rho(\textup{RPM} > 2700; \mathbf{s}, t)$ is $[-2700, 2100]$.

Consider the situation where $\max_{t \leq 10} \textup{SPEED}(t) = 5$ and $\max_{t \leq 30} \textup{RPM}(t) = 1000$.
Clearly
\begin{equation*}
  \rho(\always_{[0,10]} (\textup{SPEED}(t) < 50); \mathbf{s}, 0) = 50 - \max_{t \leq 10} \textup{SPEED}(t) = 45
\end{equation*}
and
\begin{equation*}
  \rho(\eventually_{[0,30]} (\textup{RPM}(t) > 2700); \mathbf{s}, 0) = \max_{t \leq 30} \textup{RPM}(t) - 2700 = -1700.
\end{equation*}
Therefore $\rho(\varphi; \mathbf{s}, 0) = 45$. Now if we naively take the scaling range to be $[-2700, 2100]$, which is
the smallest interval containing the ranges of the predicates, then we end up with a scaled robustness of
$45 / 2100 \approx 0.02$. This seems to indicate that the input signal is close to falsifying the requirement
$\varphi$, but this is obviously a wrong impression as the component $\textup{SPEED}(t)$ is very far away from being a
falsifying signal.

Let us compute the scaled robustness metric as defined in
\autoref{sssec:effective_range}. The formula $\varphi$ is a conjunction of two formulas, and thus
the effective range for $\varphi$ equals the effective range of the subformula whose
robustness at time $t = 0$ evaluates to a larger value. Above we saw that this corresponds to the left subformula.
Moreover, as the effective ranges of propositions are constant, it is irrelevant where the minimum corresponding to
the always operator is attained. Consequently
\begin{equation*}
  \I(\varphi; \mathbf{s}, 0) = \I(\always_{[0,10]} (\textup{SPEED} < 50); \mathbf{s}, 0) = \I(\textup{SPEED} < 50; \mathbf{s}, 0) = [-70, 50].
\end{equation*}
Therefore $\overline{\rho}(\varphi; \mathbf{s}, 0) = 45/50 = 0.9$. This number is much more reasonable and indicates
correctly that the input signal is not close to falsifying $\varphi$.

This example indicates that our scaled robustness metric can overcome problems related to naive scaling, and we propose
to use our method based on the effective range whenever scaling STL robustness values is required.

\section{Research Questions and Evaluation Methodology}\label{sec:evaluation_methodology}
In this section, we describe the details of a practical implementation of OGAN (\autoref{alg1}). In order to evaluate this implementation, we propose several research questions that we answer by conducting computational experiments in which we compare OGAN against state-of-the-art requirement falsification algorithms on standard benchmarks. We also present a novel methodology to analyze the experimental results. The research questions are answered to in \autoref{sec:results}.

\subsection{Research Questions}
A requirement falsification algorithm must obviously be able to falsify a requirement (if it is falsifiable). We call this falsification capability \emph{effectiveness}, and we measure it using falsification rate, which is defined in \autoref{ssec:evaluation_methodology}. Our first research question is thus the following:
\begin{itemize}
    \item[] \textbf{RQ1}. Is OGAN effective on common falsification benchmarks?
\end{itemize}

Two algorithms can achieve the same effectiveness, but one algorithm can use significantly more resources than another. We deem that lowering the number of SUT executions needed for a falsification amounts to increased \emph{efficiency}. We ask:
\begin{itemize}
    \item[] \textbf{RQ2}. Is OGAN efficient on common falsification benchmarks?
\end{itemize}
We study efficiency using methods from survival analysis; cf. \autoref{ssec:evaluation_methodology}.

Recall that initially the OGAN algorithm uses a Monte-Carlo sampling method to obtain the initial training data for the discriminator. It is reasonable to ask how the choice of the sampling strategy affects OGAN's effectiveness and efficiency.
\begin{itemize}
    \item[] \textbf{RQ3}. Does the choice of Monte-Carlo sampling strategy affect OGAN's performance?
\end{itemize}
We address this question by comparing two sampling strategies: uniform random sampling and Latin hypercube sampling.

A falsification algorithm that is effective and efficient in the above senses can use a lot of computational resources, which might be undesirable. OGAN trains neural networks, and this is known to be computationally expensive. This naturally leads to the following question:
\begin{itemize}
    \item[] \textbf{RQ4}. What is the computational overhead of OGAN?
\end{itemize}

The discriminator is trained to learn the relation between SUT inputs and the robustness. The purpose of the discriminator is only to allow the generator to learn to produce tests with low robustness. It is not clear how accurate the discriminator needs to be for a successful falsification. If it is accurate, then the discriminator could be used as a surrogate model in other tasks beyond requirement falsification. Thus, we ask the following question:
\begin{itemize}
    \item[] \textbf{RQ5}. Does OGAN train accurate discriminators?
\end{itemize}

Our final research question concerns the use of the generator to augment the discriminator's training data. It is not clear a priori whether it is a good idea to focus on low-robustness tests produced by the generator or if it is a better idea to use samples independent of it.
\begin{itemize}
    \item[] \textbf{RQ6}. Does augmenting the discriminator's training data by sampling the generator improve OGAN's effectiveness and efficiency?
\end{itemize}
We study this by comparing OGAN against a variant whose generator sampling is disabled.

\subsection{Evaluation Methodology}\label{ssec:evaluation_methodology}
Here we propose a methodology to evaluate OGAN and answer the proposed research questions. We aim to outline a general methodology for studying the effectiveness and efficiency of requirement falsification algorithms. A widely used guide for the evaluation of randomized algorithms is presented in the survey \cite{2012:a_hitchhikers_guide_to_statistical_tests_for_assessing}. Here we go beyond these recommendations and use methods from survival analysis that are not discussed in the survey. To the best of our knowledge, this has not been done in the context of evaluating requirement falsification algorithms.

Consider a benchmark, that is, a SUT and a requirement $\varphi$. A requirement falsification algorithm is naturally evaluated on its capability of being able to falsify $\varphi$ with a given execution budget. As most requirement falsification algorithms are stochastic, this capability is not determined well by a single run of the algorithm, so several independent replicas are required for the assessment. This leads to the concept of \emph{falsification rate} (FR) which is the proportion of successful falsifications over, say, $N$ replicas. We take FR to be our evaluation criterion for effectiveness, and we deem an algorithm $\mathcal{A}$ to be more effective than algorithm $\mathcal{A}'$ if $\mathcal{A}$ has higher FR than $\mathcal{A}'$.

Recall our assumption that evaluating the SUT output is expensive. Thus among two equally effective algorithms, the one that uses fewer executions is better: it is more efficient. A simple statistic $\overline{S}$ for efficiency is the mean of the number of executions required for a falsification. The statistic $\overline{S}$, however, only considers successful replicas, and a noneffective algorithm may achieve low $\overline{S}$. This means that the number $\overline{S}$ should be always interpreted together with the FR.

In order to compare falsification algorithms better, we propose to base the comparison on methods of survival analysis \cite{2002:the_statistical_analysis_of_failure_time_data}. Consider a falsification algorithm $\mathcal{A}$ for the falsification of a requirement $\varphi$ of some SUT. Let $E_{\mathcal{A}}$ be the number of executions needed to falsify $\varphi$ using $\mathcal{A}$, and set $E_{\mathcal{A}} = B^+$ (censored observation, i.e., an unknown value that is at least $B$) if a falsifying input was not found within a given execution budget $B$. Thus $E_{\mathcal{A}}$ is a random variable and it can be interpreted as the survival time of $\varphi$ when $\mathcal{A}$ is used for falsification. Intuitively, the algorithm $\mathcal{A}$ has better performance than an algorithm $\mathcal{A}'$ if the survival times observed from $E_{\mathcal{A}}$ tend to be lower than those observed from $E_{\mathcal{A}'}$.

A natural way to visualize survival times is to plot the survival function $S$ defined by setting
\begin{equation*}
    S(t) = \mathbb{P}(E_\mathcal{A} > t).
\end{equation*}
The standard way to estimate this function is to use the Kaplan–Meier estimator $\hat{S}$ \cite[Sec.~1.4]{2002:the_statistical_analysis_of_failure_time_data} defined as
\begin{equation*}
    \hat{S}(t) = \prod_{1 \leq i \leq t} \left( 1 - \frac{d_i}{n_i} \right)
\end{equation*}
where $d_i$ is the number of patients that have died since time $i-1$ (the number executions of $\mathcal{A}$ that managed to falsify $\varphi$ in exactly $i$ executions) and $n_i$ is the number of patients at risk (the number of executions of $\mathcal{A}$ that have not been censored and have not been able to falsify $\varphi$ with at most $i-1$ executions). See \autoref{fig:survival_1} for example survival functions (the light areas correspond to $95 \%$ confidence intervals; see below).

Here censoring happens only when the execution budget $B$ is exhausted, and in this case it can be easily proved that the final value $\hat{S}(B)$ equals $1 - F$ where $F$ is the falsification rate $\smash[tb]{\sum_{i=1}^B d_i / N}$. This connection is convenient as it allows us to compute confidence intervals for the FR. See \cite[Sec.~1.4]{2002:an_introduction_to_fuzzy_logic_and_fuzzy_sets} for more on confidence intervals for $\hat{S}(t)$ based on the $\log (- \log(\cdot))$ transformation.

Estimating the survival function thus serves two purposes. First, it yields the FR and a confidence interval for it. Second, the survival function can be used for a further comparison of algorithms achieving similar FR values. Indeed, if the survival function for an algorithm $\mathcal{A}$ decreases faster than that of an algorithm $\mathcal{A}'$, then $\mathcal{A}$ tends to falsify the requirement with fewer SUT executions than $\mathcal{A}'$. From this observation, it follows that we should test if the distributions of $E_{\mathcal{A}}$ and $E_{\mathcal{A}'}$ are the same. Under the hypothesis that the distributions are identical, the log-rank statistic \cite[Sec.~1.5]{2002:the_statistical_analysis_of_failure_time_data} is asymptotically $\chi^2_1$-distributed and can be used for computing $p$-values. In this paper, we test the hypothesis of identical distributions, but we do not provide an explicit alternative hypothesis. This would require modeling the random variables $E_{\mathcal{A}}$ and $E_{\mathcal{A}'}$, and we have not attempted to do so. Without an explicit alternative, we do the analysis on case-by-case basis. For example, the $p$-value in \autoref{tbl:pvalues_ogan} supports that the survival functions of OGAN US and OGAN LHS (defined in \autoref{ssec:mc_sampling}) are not identical in the $\ATI$ benchmark. From \autoref{fig:survival_0_1}, it is evident that, even though the FRs are similar, OGAN US tends to require fewer executions than OGAN LHS.

In addition, the survival function can be used to give a baseline assessment on how difficult a given benchmark is. We propose to do this as follows: use uniform random search as the falsification algorithm on two benchmarks and compare the estimated survival functions. The benchmark for which the survival function reaches the lowest value the fastest can be deemed to be the easier benchmark. For example, based on \autoref{fig:survival_0_1}, we consider the benchmark $\ATXVIA$ to be easier than $\ATXVIB$.

Summarizing the above, we evaluate a falsification algorithm as follows: on each benchmark, we report the falsification rate FR and its $95 \%$ confidence interval (CI). We display the estimated survival functions in order to compare the performances of several algorithms on a given benchmark. In addition, we report the mean number $\overline{S}$ of executions required for a falsification.

\subsection{Benchmark Selection}\label{ssec:benchmarks}
For our evaluation, we mainly use the benchmarks of the ARCH-COMP 2023 competition \cite{ARCH23}. In addition, we include some benchmarks from \cite{2021:falsification_of_hybrid_systems_using_adaptive}. These problems represent a varied selection of systems, and they have been used for the evaluation of requirement falsification algorithms on numerous occasions~\cite{ARCH21,ARCH22,ARCH23,2021:falsification_of_hybrid_systems_using_adaptive}.

For brevity, we have excluded from this report benchmarks that we deem as too easy. Our exclusion criterion is as follows: we do not report results for a benchmark if uniform random search can achieve a falsification rate greater than $0.5$ with a budget of $75$ executions over $50$ repetitions. This criterion excludes the benchmarks $\textup{AFC29}$, $\textup{AT2}$, $\textup{AT51}$, $\textup{AT52}$, $\textup{AT53}$, $\textup{AT54}$, $\textup{CC1}$, $\textup{CC2}$, and $\textup{CC5}$ from \cite{ARCH23}. We also exclude benchmarks of \cite{ARCH23} that at most one algorithm could falsify as there is no comparison to be made. This criterion excludes the benchmarks $\textup{SC}$ and $\textup{AFC33}$. We remark that OGAN is not able to falsify the requirements of these two difficult benchmarks. Finally, we exclude the benchmarks $\textup{AT6ABC}$, $\textup{CCX}$, $\textup{NNX}$, and $\textup{AT6ABC}$ because they concern the falsification of multiple requirements, and we only focus on the falsification of a single requirement in this paper.

Next we list the benchmarks used in our experiments. In each benchmark, we have used a sampling rate of $0.01$ for discretizing the input and output signals. \\

\noindent
\textbf{Automotive Powertrain Fuel Control (AFC).} This benchmark is proposed in
\cite{2014:powertrain_control_verification_benchmark}. Given two input signals $\signal{\theta}$ (throttle) and
$\signal{\omega}$ (brake), the model outputs a signal $\signal{\mu}$ (normalized deviation from a reference value). As
in the ARCH-COMP 2023 competition, we assume that
\begin{equation*}
  \signal[t]{\theta} \in [0, 61.2], \quad \signal[t]{\omega} \in [900, 1100], \quad \text{and} \quad \signal[t]{\mu} \in [-1, 1]
\end{equation*}
for all $t$ during the simulation time of $50$ time units. The input signals are piecewise constant signals. The
throttle $\theta$ consists of $10$ segments of $5$ time units and the brake $\omega$ is constant. We use the following
requirement:
\begin{equation*}
  \textup{AFC27} = \always_{[11,50]} \left( (\mathrm{rise} \lor \mathrm{fall}) \rightarrow (\always_{[1,5]} |\signal{\mu}| < \beta) \right) \text{ and} \\
\end{equation*}
where $\beta = 0.008$ and
\begin{align*}
  \mathrm{rise} &= (\signal{\theta} < 8.8) \land (\eventually_{[0,0.05]} \signal{\theta} > 40.0) \text{ and} \\
  \mathrm{fall} &= (\signal{\theta} > 40.0) \land (\eventually_{[0,0.05]} \signal{\theta} < 8.8).
\end{align*}
The requirement expresses that after the throttle has risen or fallen, the
deviation from the reference value should decrease soon and stay low for a while. \\

\noindent
\textbf{Automatic Transmission (AT).} This model is from Mathworks and has been proposed as a benchmark in
\cite{2015:benchmarks_for_temporal_logic_requirements_for_automotive}. Given two input signals $\textup{THROTTLE}$ and
$\textup{BRAKE}$, the model outputs three signals: $v$ (speed of the modeled car), $\omega$ (engine speed of the car in
RPM), and $g$ (currently selected gear). As in the ARCH-COMP 2023 competition, we assume that
\begin{equation*}
  \textup{THROTTLE}(t) \in [0, 100] \quad \text{and} \quad \textup{BRAKE}(t) \in [0, 325]
\end{equation*}
for all $t$ during the simulation time of $30$ time units. We assume that the input signals are piecewise constant
signals consisting of $6$ segments of $5$ time units in accordance with \cite{ARCH23}. For the output signals, we set
the following ranges:
\begin{equation*}
  \signal[t]{v} \in [0, 125], \quad \signal[t]{\omega} \in [0, 4800], \quad \text{and} \quad \signal[t]{g} \in \{1,2,3,4\}
\end{equation*}
for all $t$.

In our experiments, we consider the requirements (with obvious interpretations) from
\cite{ARCH23,2021:falsification_of_hybrid_systems_using_adaptive} defined by the following templates:
\begin{align*}
  \textup{AT1}_T &= \always_{[0,T]} \, \signal{v} < 120, \\
  \textup{AT6}_{T,v,\omega} &= (\always_{[0,30]} \signal{\omega} < \omega) \rightarrow (\always_{[0,T]} \signal{v} < v), \text{ and} \\
  \textup{ATX2} &= \lnot ( \always_{[10,30]} \signal{v} \in [50,60] ).
\end{align*}
\vspace{0.1em}

\noindent
\textbf{Chasing Cars (CC).} This model is derived from \cite{2000:towards_a_theory_of_stochastic_hybrid_systems}. This
is another car model which takes inputs $\textup{THROTTLE}$ and $\textup{BRAKE}$, and outputs the location $Y_1$ of a
car and locations $Y_2$, $Y_3$, $Y_4$, $Y_5$ of cars chasing it. We assume that
\begin{equation*}
  \textup{THROTTLE}(t), \, \textup{BRAKE}(t) \in [0, 1]
\end{equation*}
for all $t$ during the simulation of $100$ time units. The input signals are piecewise constant signals consisting of $20$ segments of $5$ time units as in \cite{ARCH23}. Experimentally we have decided to use respectively the ranges $[-250, 0]$, $[-240, 10]$, $[-230, 20]$, $[-220, 30]$, and $[-210, 40]$ for the signals $Y_1$, $\ldots$, $Y_5$. The following requirements are found in \cite{ARCH23}:
\begin{align*}
  \textup{CC3} &= \always_{[0,80]} \left( ( \always_{[0,20]} \, \signal{Y_2} - \signal{Y_1} \leq 20 ) \lor (\eventually_{[0,20]} \signal{Y_5} - \signal{Y_4} \geq 40) \right), \text{ and} \\
  \textup{CC4} &= \always_{[0,65]} \eventually_{[0,30]} \always_{[0,20]} \, \signal{Y_5} - \signal{Y_4} \geq 8.
\end{align*}
\vspace{0.1em}

\noindent
\textbf{F-16 Ground Collision Avoidance (F16).} A version of this model has been presented in
\cite{2018:verification_challenges_in_f16_ground_collision_avoidance}. The model controls an aircraft and attempts to
avoid ground collisions. The input for the model is a vector (not a signal) with three components
$\textup{ROLL}$, $\textup{PITCH}$, and $\textup{YAW}$ which determine the orientation of the aircraft in the starting
position at the altitude of $4040$ feet. As in \cite{ARCH23}, we set
\begin{equation*}
  \textup{ROLL} \in [0.2\pi, 0.2833\pi], \,\, \textup{PITCH} \in [-0.4\pi, -0.35\pi], \text{ and } \textup{YAW} \in [-0.375\pi, -0.125\pi].
\end{equation*}
The output of the model is a signal $\textup{ALTITUDE}$ representing the altitude of the aircraft during the simulation
of $15$ time units. The only requirement is the following:
\begin{equation*}
  \textup{F16} = \always_{[0,15]} \, \textup{ALTITUDE} > 0.
\end{equation*}
In our experiments, we use the range $[0, 4040]$ for the output signal.

We stress that there are many versions of the F16 model; see for example the Python implementation\footnote{\url{https://github.com/stanleybak/AeroBenchVVPython}}. We have observed different models can have wildly different behavior especially if the input ranges are modified. We use the Matlab implementation and signal ranges presented in \cite{ARCH23}.

Observe that a requirement falsification algorithm that requires online testing is not suitable to falsify the requirement $\textup{F16}$ because the input is not a signal.\\

\noindent
\textbf{Neural Network Controller (NN).} This model is a neural network controller for magnet levitation at a reference
position based on an example from
Mathworks\footnote{\url{https://www.mathworks.com/help/deeplearning/ug/design-narma-l2-neural-controller-in-simulink.html}}.
The input signal is a reference position $R$ and the output $P$ is the current position of the magnet. We use the
following requirement template from \cite{ARCH23}:
\begin{equation*}
  \textup{NN}_\beta  = \always_{[1,37]} ( |\signal{P} - \signal{R}| > \alpha + \beta|\signal{R}| \rightarrow \eventually_{[0,2]} \always_{[0,1]} |\signal{P} - \signal{R}| < \alpha + \beta|\signal{R}|)
\end{equation*}
where $\alpha = 0.005$. The simulation time is $40$ time units, and we assume that the input signal
consists of $3$ segments of $40/3$ time units. For the requirement $\textup{NN}_\beta$, we assume that
$\signal[t]{R} \in [1, 3]$ for all $t$, and we assume that $\signal[t]{R} \in [1.95, 2.05]$ for all $t$ in the case of
the other requirement. For the output position $P$, we use the range $[0, 4]$.\\

\noindent
\textbf{Pacemaker (PM).} This model is a simple controller of a pacemaker device described in \cite{2022:two_simulink_models_with_requirements_for_a_simple}. The output is the $\textup{PACECOUNT}$ signal which counts how many heart paces have been observed since the beginning of the simulation (this is a nondecreasing discrete-valued signal). The input is the lower rate limit signal $\textup{LRL}$. It specifies the minimum amount of paces per minute that the pacemaker should achieve. We assume that $\textup{LRL}(t) \in [50,90]$ for all $t$ and that $\textup{LRL}$ is piecewise constant signal consisting of $5$ segments of $2$ time units for a total a total length of $10$ time units. The input signal values can be arbitrary floating point numbers in the interval $[50,90]$, but they are rounded to the nearest integer internally. We consider the following requirement given in \cite{2022:two_simulink_models_with_requirements_for_a_simple}:
\begin{equation*}
  \textup{PM} = \always_{[0,10]} \, \textup{PACECOUNT} \leq 15 \land \eventually_{[0,10]} \, \textup{PACECOUNT} \geq 8.
\end{equation*}
It specifies that the observed pacing should match the requested LRL. Indeed, $90$ paces in a minute correspond to $15$ paces in $10$ seconds and $50$ paces a minute to $8$ paces in $10$ seconds.

\subsection{OGAN Setup for Experiments}\label{ssec:ogan_setup}
This section describes what choices we have made in setting up \autoref{alg1} for the experiments. We have
refrained from tuning the parameters for each benchmark separately, and we describe a common setup that has good overall
performance. Our description is a distillation of our experience with OGAN so far and can be taken as a good initial setup in subsequent experiments.\\

\noindent
\textbf{The Latent Space.} For all benchmarks, we set the latent space $\latentspace$ to be $[-1, 1]^{d_H}$. The dimensions of the input spaces (see \autoref{ssec:benchmarks}) take values in $\{3, 11, 12, 40\}$, and it is easier to learn a mapping from a higher-dimensional space to a lower-dimensional space, so we compromise and select $D = 20$.\\

\noindent
\textbf{Discriminator and Generator Models.} The generator $\generator$ and the discriminator $\discriminator$ are modeled as neural networks. This is important as it allows $\generator$ to be trained against $\discriminator$ by freezing the weights of $\discriminator$ and applying the usual back-propagation of errors to the weights of $\generator$. The $\generator$ is a neural network mapping $\latentspace$ to inputs in $[-1, 1]^K$ for some $K$, and $\discriminator$ maps $[-1, 1]^K$ to $[0,1]$. Appropriate activation functions are used to ensure that the ranges of $\generator$ and $\discriminator$ are respectively contained in $[-1, 1]^K$ and $[0, 1]$.

Since our execution budgets are low ($300$, $1500$), the training data is rather
small. Thus it would be unreasonable to use large or deep neural network models. In all cases, the generator network is
a fully connected network with three hidden layers of $128$ neurons. We use $\tanh$ activation for the output (since
the tests components are elements of $[-1, 1]$) and leaky ReLU with negative slope $0.01$ for other activations.

In all experiments with input signals, that is, all excluding F16, we use a convolutional  network for the discriminator.
We use two convolutional layers of $16$ feature maps with kernel size $2$, stride $1$, and $0$-padding
$1$ followed by a leaky ReLU activation with negative slope $0.01$ and a maxpool layer with window size $2$ and stride
$2$. The convolutional layer is flattened and mapped to a hidden layer of $128$ neurons (no activation). The output of
this final layer is fed through the sigmoid function to produce a single number.
The input in the F16 benchmark cannot be understood as a signal or a time series, and we use the above generator setup with sigmoid output activation.

We have observed that the proposed models have large enough capacity to overfit to a training data of $300$ samples
(with the number of epochs described below).

In both the generator and the discriminator, we initialize the weights using the Glorot-Bengio initialization
\cite{2010:understanding_the_difficulty_of_training_deep_feedforward} with layers that use a tanh or sigmoid activation and the He et al.\ initialization \cite{2015:delving_deep_into_rectifiers_surpassing_human-level} for layers with ReLU
activations.\\

\noindent
\textbf{Training of the Neural Networks.} We use a modified squared error as the loss function used in training of both
$\discriminator$ and $\generator$. Since our robustness metric takes values in $[0,1]$, the squared error does not
penalize for errors in a correct way. Instead, we propose the following. Let $F(x) = \mathrm{logit}(0.98x + 0.01)$
where $\mathrm{logit(x)} = \log x/(1-x)$ (the constants $0.98$ and $0.01$ are arbitrary, and their purpose is to avoid
singularities when $x = 0$ or $x = 1$). We let
\begin{equation*}
  \mathcal{L}(\hat{y}, y) = (F(\hat{y}) - F(y))^2 + \lambda\left(F\left( \frac12 - \frac{\hat{y} - y}{2}\right) - F\left(\frac12\right) \right)^2
\end{equation*}
with $\lambda = 0.001$. The first summand of the loss function encourages the network to minimize errors especially when
the ground truth is close to $0$ or $1$. This is exactly what we need: the discriminator does not need to be perfect,
it just needs to predict ``large'' when the ground truth is ``large'' and ``small'' when the ground truth is ``small''.
The second summand softly enforces our wish that $y = \hat{y}$. Indeed, the function $F$ is symmetric about $1/2$ so,
in principle, the first term could be minimized by setting $y \approx 1$ when $\hat{y} \approx 0$ (or vice versa). We
have not observed this behavior in our experiments, but it is better to be safe than sorry.

We use the Adam optimizer for training both $\discriminator$ and $\generator$. We use the default values $0.9$ and
$0.999$ for the beta parameters and learning rates $0.005$ and $0.0001$ for the discriminator and the generator
respectively. We used the hyperparameter tuning and found that this combination of
learning rates gave the best overall performance on all benchmarks when the learning rates were selected from the set
$\{0.5, 0.1, 0.05, 0.01, 0.005, 0.001, 0.0005, 0.0001\}$. We recommend to start with the selected learning rates in any
subsequent study. Based on our experience, we suggest that the discriminator learning rate should be larger than the
generator learning rate.

We use $15$ training epochs for the discriminator and, on each epoch, we train on a single batch consisting of the
complete training data so far. We have not considered any mechanisms to prevent harmful overfitting but, based on our
experience with the benchmarks, it seems that this setup is good enough so that the discriminator can generalize and
generators trained on it can accomplish their task. With the generator, we use $375$ epochs with batch size $32$ (i.e.,
$B = 32$ in \autoref{alg1}). This overfits the generator on the discriminator to some degree which means that the
generator will produce tests that are quite similar to each other. As long as the discriminator is of good quality,
this has no practical implications when the objective is to find one single falsifying input as is the case in these benchmarks.\\

\noindent
\textbf{Other.} We set $\alpha = 0.95$ (see the line $24$ of \autoref{alg1}).

\subsection{Baseline and Data Collection}
We compare OGAN to a uniform random search algorithm and a total of eleven tools that participated in the ARCH-COMP 2021 and 2023 competitions \cite{ARCH21,ARCH23}. For brief introductions to these tools, see \autoref{sec:related_work}. The experiment results were collected on a desktop PC running Ubuntu 22.04 with an Intel i9-10900X CPU, a NVIDIA GeForce RTX 3090 GPU, and 64 GB of RAM.

\subsection{Replication of Experiment Results}
OGAN is implemented as a part of the STGEM falsification framework, which was used to obtain the results presented in this paper. STGEM can perform different testing tasks, including requirement falsification, when provided with a suitable executable function for the SUT and the requirements. STGEM supports several search algorithms such as OGAN, WOGAN \cite{WOGAN}, grid search, random sampling algorithms, and several evolutionary algorithms. It can also combine different algorithms such as a random sampling algorithm and OGAN in the same falsification task. This is how the initial training data is collected for OGAN. The different search algorithms are independent of each other and can be applied sequentially or in parallel.

STGEM can tune the hyperparameters of its algorithms. In fact, any of its algorithms can be used for the hyperparameter search. It also has facilities for executing computational experiments and running multiple replicas of a falsification task. Careful use of the random number generator ensures that the results are replicable. STGEM is written in Python and uses high performance libraries, such as Numpy and PyTorch, for mathematical operations and machine learning.

A SUT in STGEM can be defined as a Python class, as a Matlab Simulink model, or as an external operating system process. STGEM has full support for requirements in STL including an implementation of the scaled robustness metric of \autoref{sssec:effective_range}). It is also possible to define a custom requirement monitor and robustness metric.

The STGEM framework and its implementation of the OGAN algorithm is freely available at \url{https://gitlab.abo.fi/stc/stgem}. The benchmarks presented in this paper are available at \url{https://gitlab.abo.fi/stc/experiments/ogan} along with the repeatibility instructions. The results necessary for recreating the tables and figures of this article are available at \url{https://dx.doi.org/10.5281/zenodo.8382845}.

\section{Experiment Results}\label{sec:results}

\subsection{RQ1 and RQ2: Effectiveness and Efficiency}
Recall that the effectiveness of a requirement falsification algorithm measures its capability of being able to falsify a requirement. RQ1 asks if OGAN is effective on common falsification benchmarks. RQ2 asks the same for efficiency; increased efficiency means fewer needed SUT executions. We answer RQ1 and RQ2 by comparing OGAN's effectiveness and efficiency to those of the requirement falsification algorithms from the ARCH-COMP 2021 and 2023 competitions and \cite{2021:falsification_of_hybrid_systems_using_adaptive}. We omit ARCH-COMP 2022 results since in this iteration of the competition the execution budget was unlimited.

\subsubsection{Comparison to ARCH-COMP 2023 Tools}
\begin{figure*}
\centering
\includegraphics[trim=35 5 35 0,clip,width=0.41\textwidth]{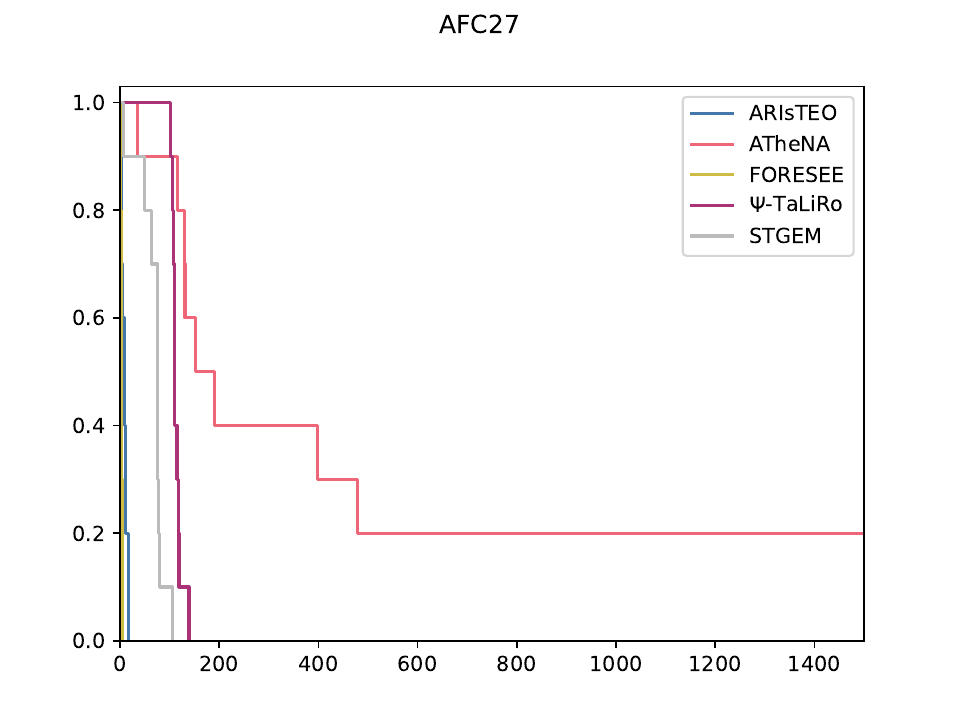}
\includegraphics[trim=35 5 35 0,clip,width=0.41\textwidth]{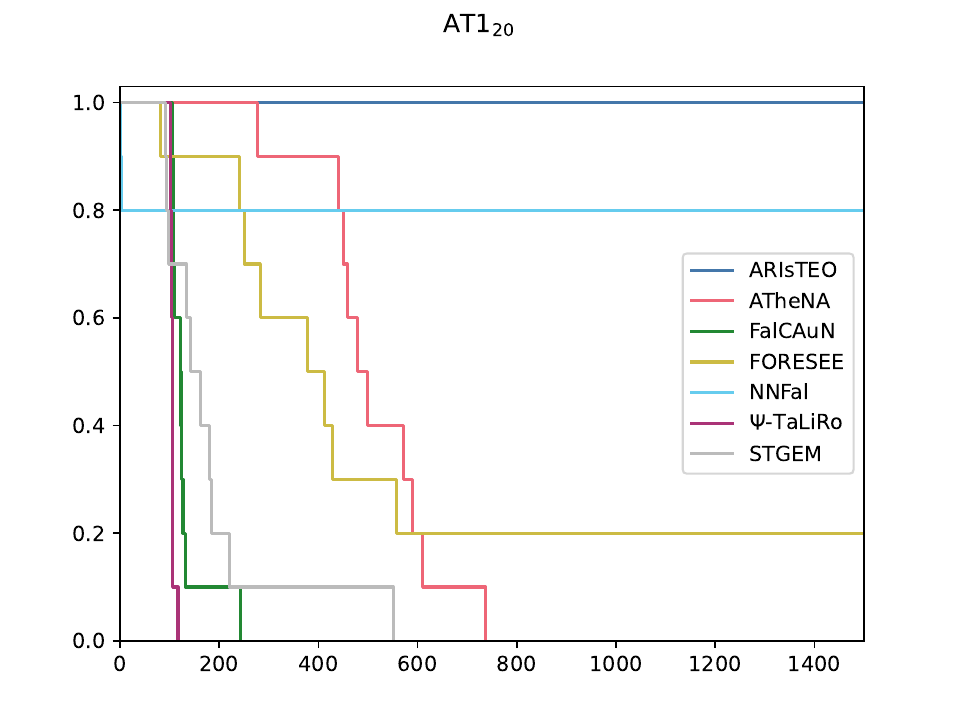}
\includegraphics[trim=35 5 35 0,clip,width=0.41\textwidth]{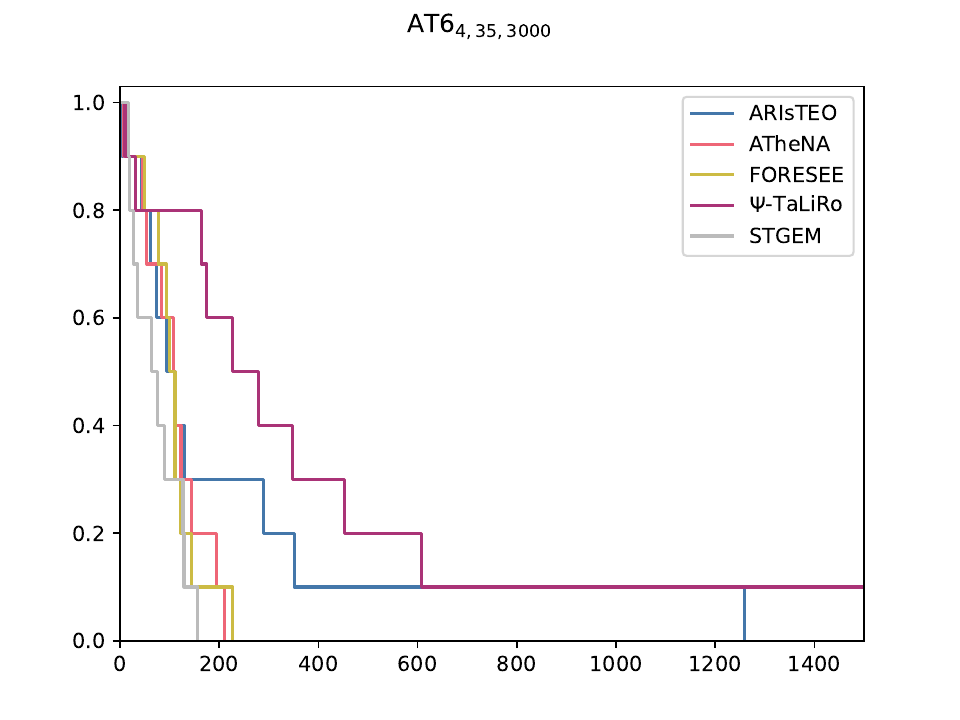}
\includegraphics[trim=35 5 35 0,clip,width=0.41\textwidth]{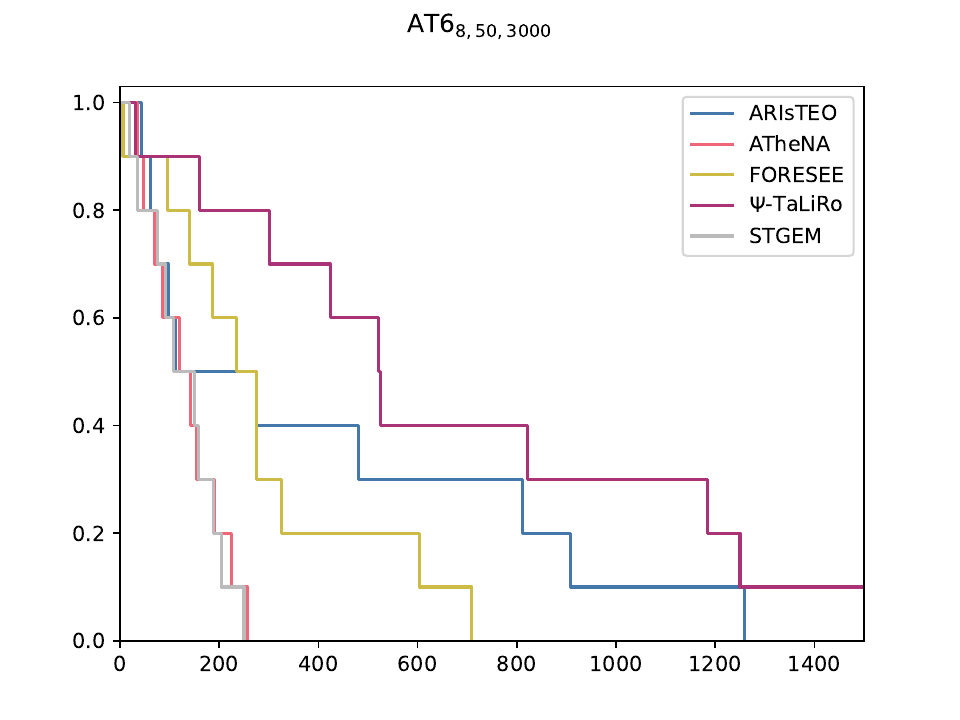}
\includegraphics[trim=35 5 35 0,clip,width=0.41\textwidth]{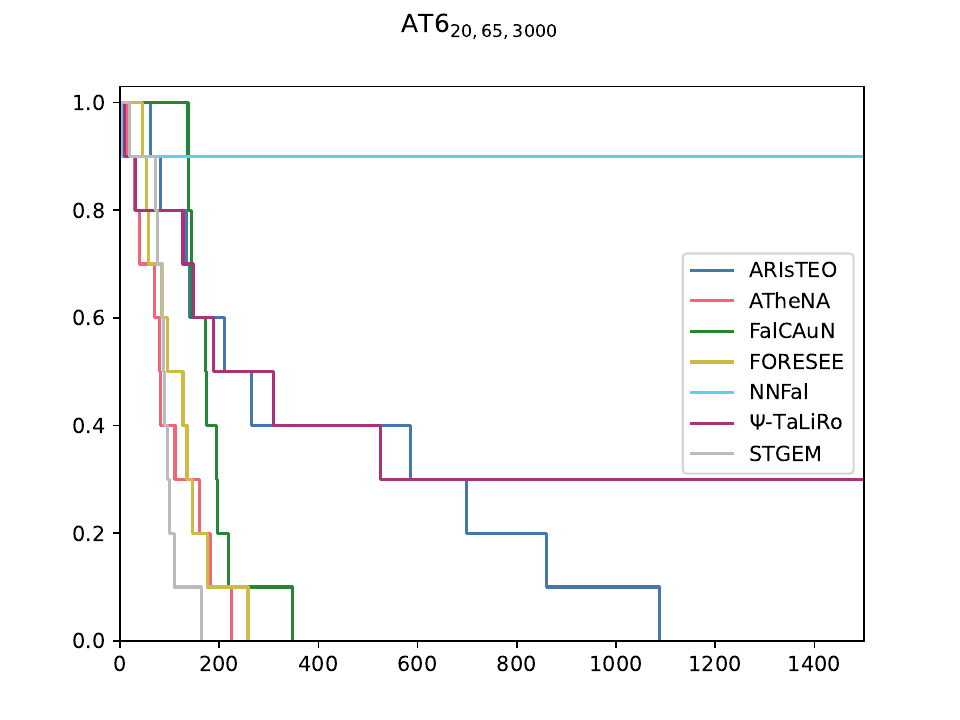}
\includegraphics[trim=35 5 35 0,clip,width=0.41\textwidth]{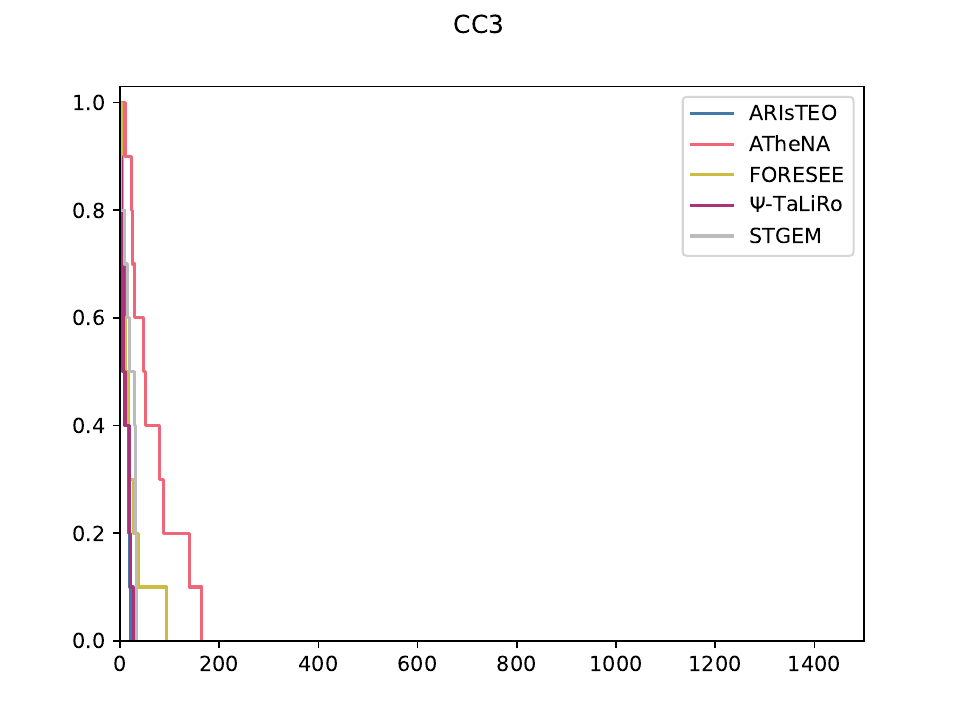}
\includegraphics[trim=35 5 35 0,clip,width=0.41\textwidth]{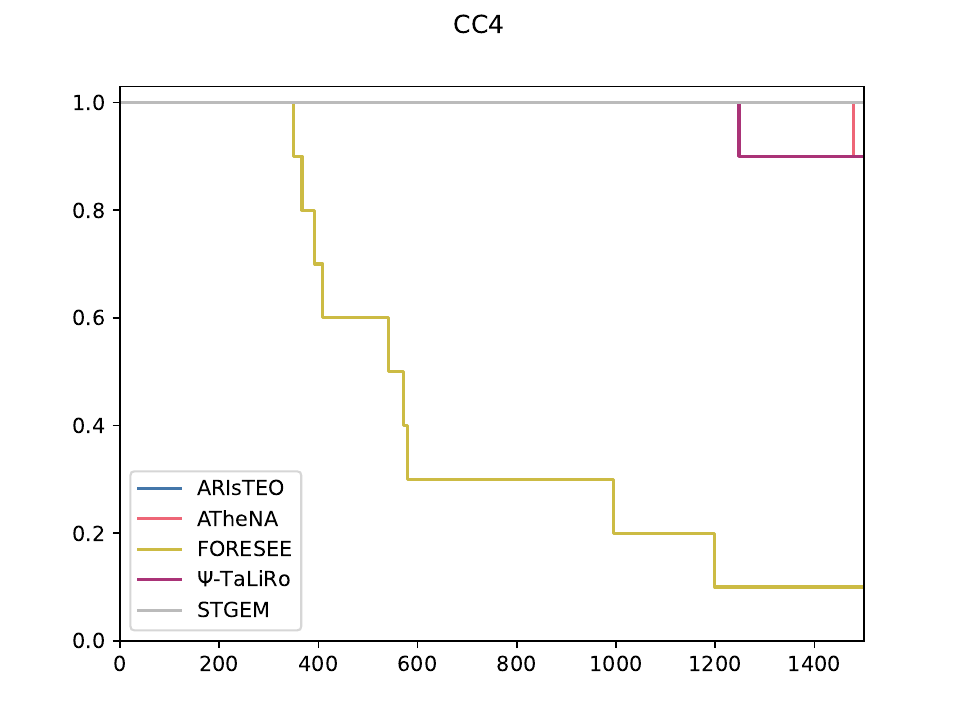}
\includegraphics[trim=35 5 35 0,clip,width=0.41\textwidth]{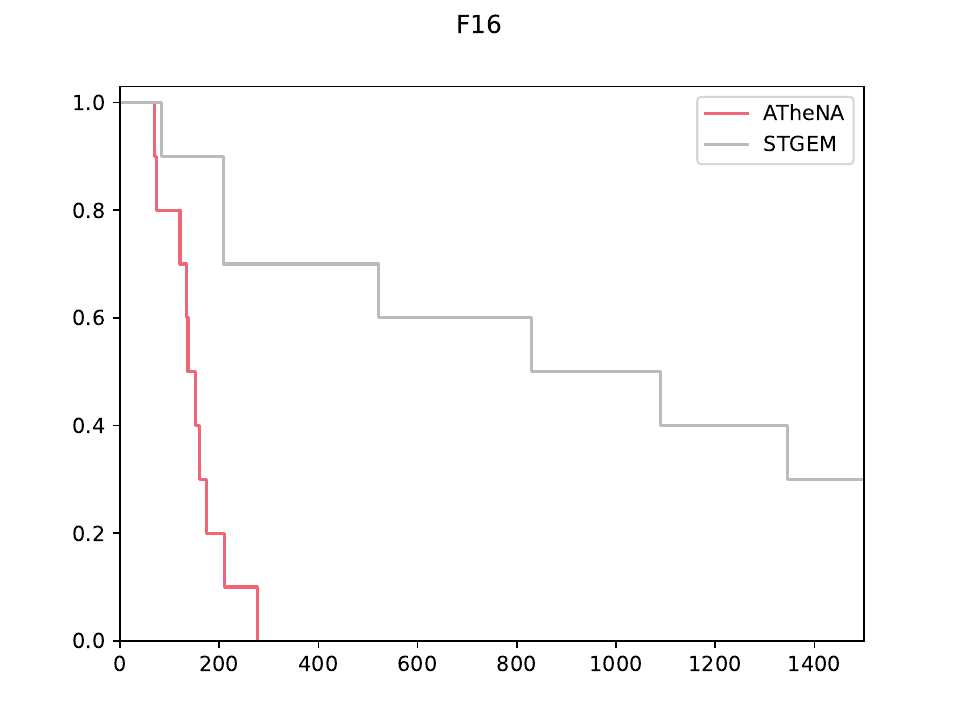}
\caption{Survival functions for ARCH-COMP 2023 tools on benchmarks from \autoref{ssec:benchmarks}.}\label{fig:arch23_survival_0}
\end{figure*}

\begin{figure*}
\centering
\includegraphics[trim=35 5 35 0,clip,width=0.41\textwidth]{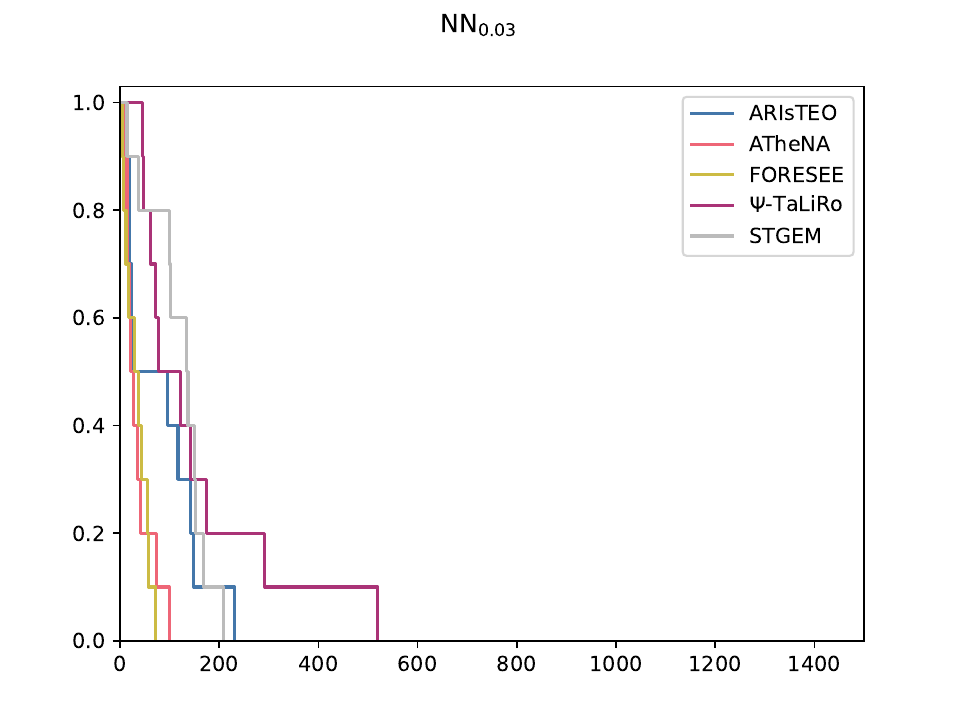}
\includegraphics[trim=35 5 35 0,clip,width=0.41\textwidth]{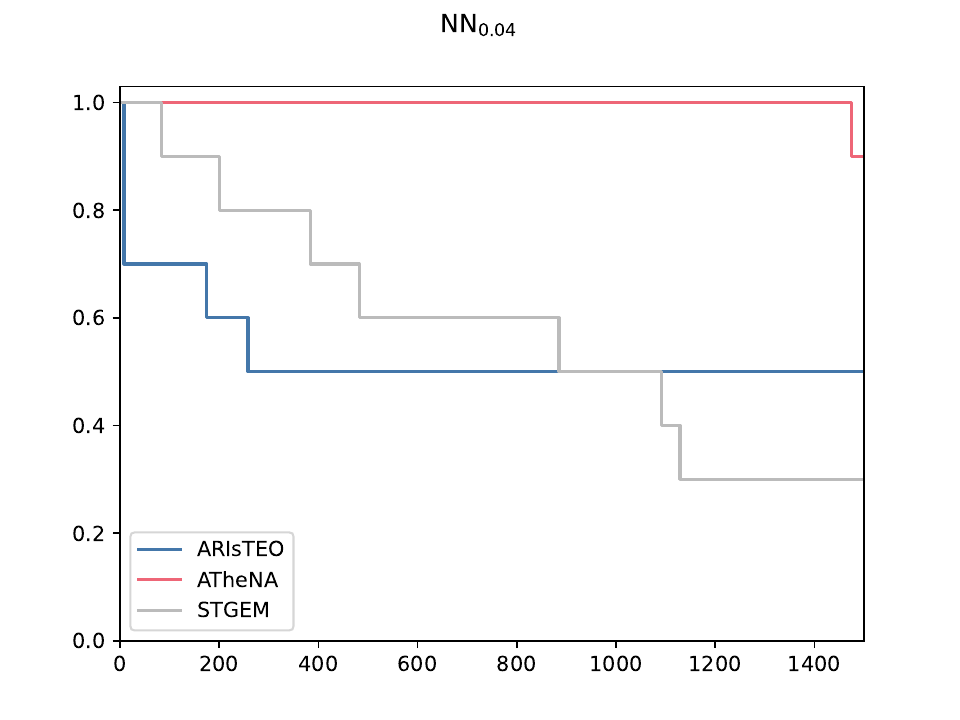}
\includegraphics[trim=35 5 35 0,clip,width=0.41\textwidth]{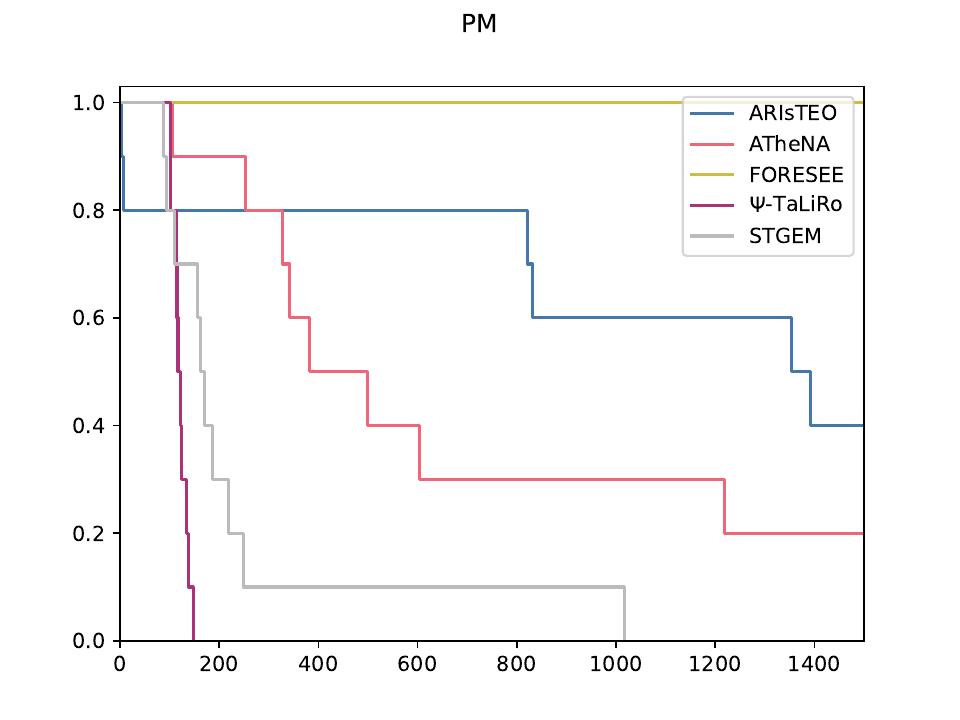}
\includegraphics[trim=35 5 35 0,clip,width=0.41\textwidth]{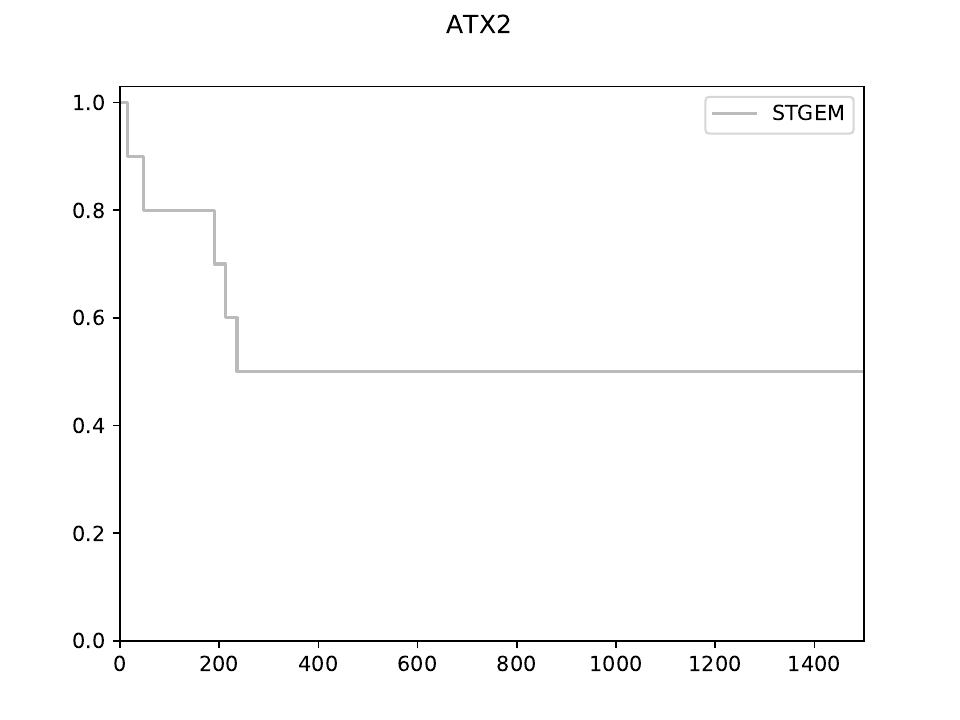}
\includegraphics[trim=35 5 35 0,clip,width=0.41\textwidth]{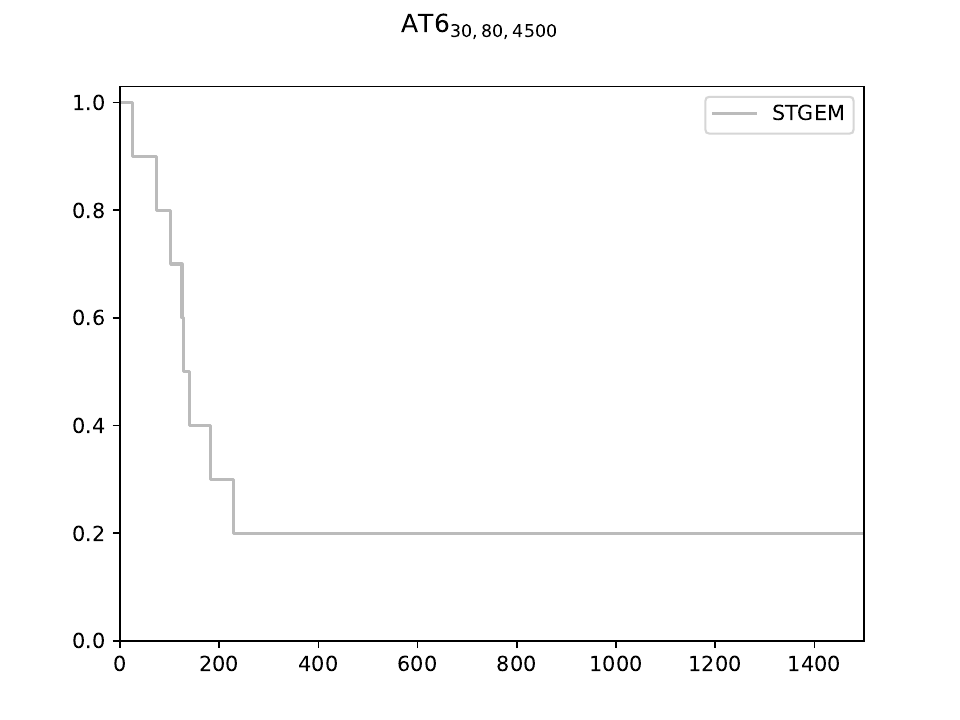}
\includegraphics[trim=35 5 35 0,clip,width=0.41\textwidth]{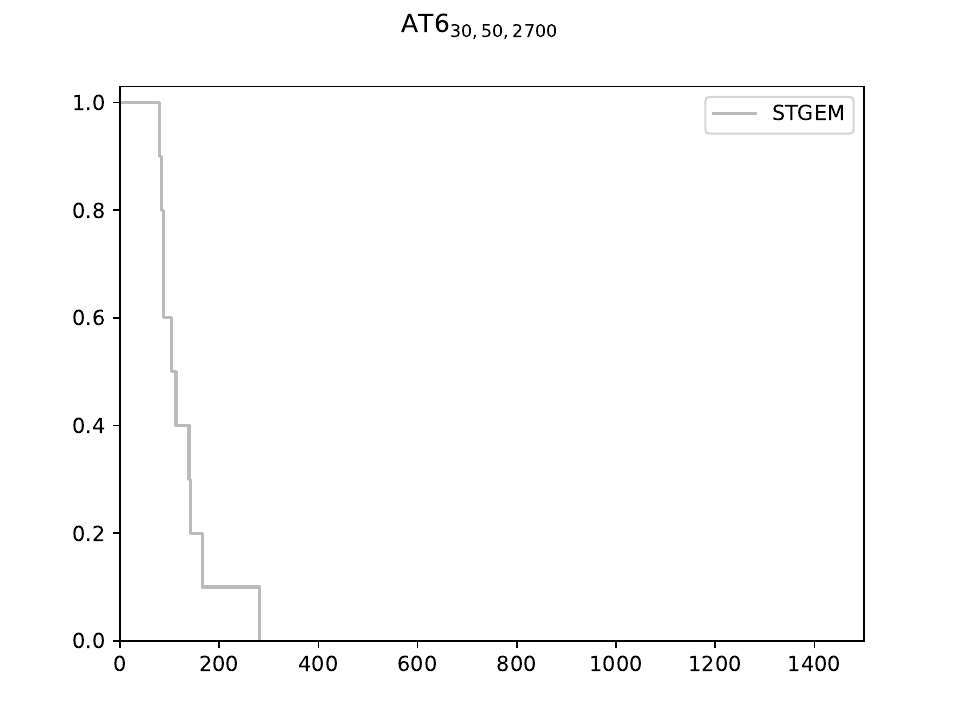}
\caption{\autoref{fig:arch23_survival_0} continued.}\label{fig:arch23_survival_1}
\end{figure*}

\autoref{tbl:arch23_results} and Figures \ref{fig:arch23_survival_0} and \ref{fig:arch23_survival_1} report the results of the ARCH-COMP 2023 competition \cite{ARCH23} for the benchmarks of \autoref{ssec:benchmarks}. The reported data is based on the results collected by the competition organizers\footnote{\url{https://dx.doi.org/10.5281/zenodo.8024426}}. The competition report only states the statistics FR and $\overline{S}$. We estimated the confidence intervals and survival functions from the data using the methods described in \autoref{ssec:evaluation_methodology}. The results are based on only $10$ replicas and the reported confidence intervals are quite wide due to the small sample size. For completeness sake, we include the results for the benchmarks $\ATXII$, $\ATXVIA$, and $\ATXVIB$ which were not part of the competition.

The algorithms of ARCH-COMP 2023 are described in more detail in \autoref{sec:related_work}. \autoref{tbl:atomic} indicates which tools use offline test execution and which of them use precollected data or models. Recall that OGAN uses offline test execution and does not use any previous data or models.

The ARCH-COMP 2023 used an execution budget of $1500$ tests. The input parametrization was the same as described in \autoref{ssec:ogan_setup}. OGAN was configured as described in \autoref{ssec:ogan_setup}. The initial Monte Carlo sampling was chosen to be uniform random sampling with an execution budget of $75$. The sampling probability $P$ was set to $0.21$ which means that, on average, $375$ ($25 \%$ of $1500$) uniform random samples were executed during the total budget of $1500$ executions.

From the results, we see that, with the exception of the benchmarks $\CCIV$, $\FA$, and $\NN_{0.04}$, OGAN has full FR and the survival function decay is among the fastest. The benchmark $\CCIV$ appears impossible for OGAN to falsify. This appears to be the case for all algorithms except FORESEE. Data for the $\FA$ benchmark is reported by only two algorithms: ATheNA and OGAN. ATheNA clearly wins OGAN in effectiveness and efficiency. We remark that ATheNA uses handcrafted robustness metrics for each benchmark \cite{2023:search_based_software_testing_driven}. This can be seen as use of precollected information, and it is unclear how it affects ATheNA's performance. Finally, on the $\NN_{0.04}$ benchmark OGAN achieves the best FR. The sample size is small, so we cannot confidently say that OGAN is better than ARIsTEO on this benchmark.

\begin{sidewaystable}
\resizebox{\textwidth}{!}{
\begin{tabular}{lrrrrrrrrrrrrrrrrrrrrr}
  \toprule
       & \multicolumn{3}{c}{ARIsTEO:}
       & \multicolumn{3}{c}{ATheNA:}
       & \multicolumn{3}{c}{FalCAuN:}
       & \multicolumn{3}{c}{FORESEE:}
       & \multicolumn{3}{c}{NNFal:}
       & \multicolumn{3}{c}{$\Psi$-TaLiRo:}
       & \multicolumn{3}{c}{STGEM:} \\
       & \multicolumn{3}{c}{ARX-2}
       & \multicolumn{3}{c}{}
       & \multicolumn{3}{c}{}
       & \multicolumn{3}{c}{}
       & \multicolumn{3}{c}{}
       & \multicolumn{3}{c}{ConBO}
       & \multicolumn{3}{c}{OGAN} \\
  \midrule \midrule
  Req.
       & \multicolumn{1}{c}{FR} & \multicolumn{1}{c}{$95 \%$ CI} & \multicolumn{1}{c}{$\overline{S}$}
       & \multicolumn{1}{c}{FR} & \multicolumn{1}{c}{$95 \%$ CI} & \multicolumn{1}{c}{$\overline{S}$}
       & \multicolumn{1}{c}{FR} & \multicolumn{1}{c}{$95 \%$ CI} & \multicolumn{1}{c}{$\overline{S}$}
       & \multicolumn{1}{c}{FR} & \multicolumn{1}{c}{$95 \%$ CI} & \multicolumn{1}{c}{$\overline{S}$}
       & \multicolumn{1}{c}{FR} & \multicolumn{1}{c}{$95 \%$ CI} & \multicolumn{1}{c}{$\overline{S}$}
       & \multicolumn{1}{c}{FR} & \multicolumn{1}{c}{$95 \%$ CI} & \multicolumn{1}{c}{$\overline{S}$}
       & \multicolumn{1}{c}{FR} & \multicolumn{1}{c}{$95 \%$ CI} & \multicolumn{1}{c}{$\overline{S}$} \\
  \cmidrule(r){1-1}
  \cmidrule(lr){2-4}
  \cmidrule(lr){5-7}
  \cmidrule(lr){8-10}
  \cmidrule(lr){11-13}
  \cmidrule(lr){14-16}
  \cmidrule(lr){17-19}
  \cmidrule(lr){20-22}
  $\textup{AFC27}$ &
  $1.00$ &
  $[1.00, 1.00]$ &
  $8.5$ &
  $0.80$ &
  $[0.53, 0.97]$ &
  $204.2$ &
   &
   &
   &
  $1.00$ &
  $[1.00, 1.00]$ &
  $2.6$ &
   &
   &
   &
  $1.00$ &
  $[1.00, 1.00]$ &
  $113.2$ &
  $1.00$ &
  $[1.00, 1.00]$ &
  $69.4$ \\
  \cmidrule(r){1-1}
  \cmidrule(lr){2-4}
  \cmidrule(lr){5-7}
  \cmidrule(lr){8-10}
  \cmidrule(lr){11-13}
  \cmidrule(lr){14-16}
  \cmidrule(lr){17-19}
  \cmidrule(lr){20-22}
  $\textup{AT1}_{20}$ &
  $0.00$ &
  $[0.00, 0.00]$ &
  -- &
  $1.00$ &
  $[1.00, 1.00]$ &
  $511.0$ &
  $1.00$ &
  $[1.00, 1.00]$ &
  $130.1$ &
  $0.80$ &
  $[0.53, 0.97]$ &
  $329.1$ &
  $0.20$ &
  $[0.05, 0.59]$ &
  $1.5$ &
  $1.00$ &
  $[1.00, 1.00]$ &
  $105.4$ &
  $1.00$ &
  $[1.00, 1.00]$ &
  $185.8$ \\
  $\textup{AT6}_{4,35,3000}$ &
  $1.00$ &
  $[1.00, 1.00]$ &
  $241.7$ &
  $1.00$ &
  $[1.00, 1.00]$ &
  $108.1$ &
   &
   &
   &
  $1.00$ &
  $[1.00, 1.00]$ &
  $104.8$ &
   &
   &
   &
  $0.90$ &
  $[0.64, 0.99]$ &
  $255.0$ &
  $1.00$ &
  $[1.00, 1.00]$ &
  $74.2$ \\
  $\textup{AT6}_{8,50,3000}$ &
  $1.00$ &
  $[1.00, 1.00]$ &
  $412.4$ &
  $1.00$ &
  $[1.00, 1.00]$ &
  $132.7$ &
   &
   &
   &
  $1.00$ &
  $[1.00, 1.00]$ &
  $285.2$ &
   &
   &
   &
  $0.90$ &
  $[0.64, 0.99]$ &
  $580.0$ &
  $1.00$ &
  $[1.00, 1.00]$ &
  $128.3$ \\
  $\textup{AT6}_{20,65,3000}$ &
  $1.00$ &
  $[1.00, 1.00]$ &
  $412.4$ &
  $1.00$ &
  $[1.00, 1.00]$ &
  $132.7$ &
  $1.00$ &
  $[1.00, 1.00]$ &
  $186.6$ &
  $1.00$ &
  $[1.00, 1.00]$ &
  $117.8$ &
  $0.10$ &
  $[0.01, 0.53]$ &
  $3.0$ &
  $0.70$ &
  $[0.42, 0.93]$ &
  $191.1$ &
  $1.00$ &
  $[1.00, 1.00]$ &
  $89.6$ \\
  \cmidrule(r){1-1}
  \cmidrule(lr){2-4}
  \cmidrule(lr){5-7}
  \cmidrule(lr){8-10}
  \cmidrule(lr){11-13}
  \cmidrule(lr){14-16}
  \cmidrule(lr){17-19}
  \cmidrule(lr){20-22}
  $\textup{CC3}$ &
  $1.00$ &
  $[1.00, 1.00]$ &
  $12.8$ &
  $1.00$ &
  $[1.00, 1.00]$ &
  $66.0$ &
   &
   &
   &
  $1.00$ &
  $[1.00, 1.00]$ &
  $22.4$ &
   &
   &
   &
  $1.00$ &
  $[1.00, 1.00]$ &
  $11.7$ &
  $1.00$ &
  $[1.00, 1.00]$ &
  $20.3$ \\
  $\textup{CC4}$ &
  $0.00$ &
  $[0.00, 0.00]$ &
  -- &
  $0.10$ &
  $[0.01, 0.53]$ &
  $1479.0$ &
   &
   &
   &
  $0.90$ &
  $[0.64, 0.99]$ &
  $600.6$ &
   &
   &
   &
  $0.10$ &
  $[0.01, 0.53]$ &
  $1248.0$ &
  $0.00$ &
  $[0.00, 0.00]$ &
  -- \\
  \cmidrule(r){1-1}
  \cmidrule(lr){2-4}
  \cmidrule(lr){5-7}
  \cmidrule(lr){8-10}
  \cmidrule(lr){11-13}
  \cmidrule(lr){14-16}
  \cmidrule(lr){17-19}
  \cmidrule(lr){20-22}
  $\textup{F16}$ &
   &
   &
   &
  $1.00$ &
  $[1.00, 1.00]$ &
  $151.1$ &
   &
   &
   &
   &
   &
   &
   &
   &
   &
   &
   &
   &
  $0.70$ &
  $[0.42, 0.93]$ &
  $612.1$ \\
  \cmidrule(r){1-1}
  \cmidrule(lr){2-4}
  \cmidrule(lr){5-7}
  \cmidrule(lr){8-10}
  \cmidrule(lr){11-13}
  \cmidrule(lr){14-16}
  \cmidrule(lr){17-19}
  \cmidrule(lr){20-22}
  $\textup{NN}_{0.03}$ &
  $1.00$ &
  $[1.00, 1.00]$ &
  $83.7$ &
  $1.00$ &
  $[1.00, 1.00]$ &
  $35.8$ &
   &
   &
   &
  $1.00$ &
  $[1.00, 1.00]$ &
  $33.6$ &
   &
   &
   &
  $1.00$ &
  $[1.00, 1.00]$ &
  $155.5$ &
  $1.00$ &
  $[1.00, 1.00]$ &
  $120.4$ \\
  $\textup{NN}_{0.04}$ &
  $0.50$ &
  $[0.25, 0.82]$ &
  $91.4$ &
  $0.10$ &
  $[0.01, 0.53]$ &
  $1475.0$ &
  &
  &
  &
  &
  &
  &
  &
  &
  &
  &
  &
  &
  $0.70$ &
  $[0.42, 0.93]$ &
  $607.7$ \\
  \cmidrule(r){1-1}
  \cmidrule(lr){2-4}
  \cmidrule(lr){5-7}
  \cmidrule(lr){8-10}
  \cmidrule(lr){11-13}
  \cmidrule(lr){14-16}
  \cmidrule(lr){17-19}
  \cmidrule(lr){20-22}
  $\textup{PM}$ &
  $0.60$ &
  $[0.33, 0.88]$ &
  $734.7$ &
  $0.80$ &
  $[0.53, 0.97]$ &
  $466.4$ &
   &
   &
   &
  $0.00$ &
  $[0.00, 0.00]$ &
  -- &
   &
   &
   &
  $1.00$ &
  $[1.00, 1.00]$ &
  $121.5$ &
  $1.00$ &
  $[1.00, 1.00]$ &
  $166.5$
\end{tabular}
}
\caption{Falsification results over $10$ independent replicas for ARCH-COMP 2023 tools. We report falsification rate (FR), i.e., the ratio of successful falsifications out of $10$ trials, its $95 \%$ confidence interval, and mean number $\overline{S}$ of executions required for a falsification.}\label{tbl:arch23_results}
\vspace{3em}
\resizebox{\textwidth}{!}{
\begin{tabular}{lccccccccccccc}
  \toprule
       & ARIsTEO
       & ATheNA
       & Breach
       & FalCAuN
       & falsify
       & FALSTAR
       & FORESEE
       & NNFal
       & $\Psi$-TaLiRo
       & S-TaLiRo
       & STGEM
       & zlscheck \\
  \midrule
  \midrule
  Atomic test execution &
  $\pyes$ &
  $\pyes$ &
  $\pyes$ &
  $\pyes$ &
  $\pno$ &
  $\pno$ &
  $\pyes$ &
  $\pyes$ &
  $\pyes$ &
  $\pyes$ &
  $\pyes$ &
  $\pno$ \\
  \hline
  No precollected data &
  $\pyes$ &
  $\pno$ &
  $\pyes$ &
  $\pyes$ &
  $\pno$ &
  $\pyes$ &
  $\pyes$ &
  $\pno$ &
  $\pyes$ &
  $\pyes$ &
  $\pyes$ &
  $\pno$
\end{tabular}
}
\caption{Information on which of the considered falsification algorithms use atomic testing and which use previously collected data or models.}\label{tbl:atomic}
\end{sidewaystable}

\subsubsection{Comparison to ARCH-COMP 2021 Tools}\label{sssec:arch21_comparison}
Tables \ref{tbl:arch21_results} and \ref{tbl:alvts_results} report the results of the ARCH-COMP 2021 competition \cite{ARCH21} and the article \cite{2021:falsification_of_hybrid_systems_using_adaptive} together with the corresponding results for the OGAN algorithm. Both sources have a common execution budget of $300$ executions. Unfortunately no data is available to compute confidence intervals or survival functions, so we merely report the statistics FR and $\overline{S}$. The input parametrization is the same as in \autoref{ssec:benchmarks}. The reported results are based on $50$ independent replicas, so the results are more reliable than those reported above.

The OGAN setup is as in \autoref{ssec:ogan_setup}. We use $75$ initial random samples obtained via uniform random sampling. The sampling probability $P$ is set to $0$. 

The results show that OGAN achieves full or almost full FR on all benchmarks except $\ATXII$, $\CCIV$, $\FA$, and $\NN_{0.04}$. $\CCIV$ is again very hard for most algorithms to falsify. On the $\ATXII$ benchmark, OGAN loses clearly to Breach and FALSTAR. Even though OGAN can achieve a higher FR with a bigger budget, as is evident from \autoref{fig:arch23_survival_1}, it appears to need at least an execution budget of $300$ to achieve a FR at least $0.5$; the algorithms Breach and FALSTAR are much more efficient. On the $\FA$ benchmark OGAN achieves the best FR. On the $\NN_{0.04}$ benchmark, OGAN is the second best, but we remark that the astonishing performance of zlscheck is likely due to its ability of being able to access the internals of the SUT \cite{ARCH21}.

Comparing algorithm efficiency is now much harder in the absence of survival function estimates. The algorithm achieving the lowest $\overline{S}$ varies from benchmark to benchmark. It is however safe to say that on all benchmarks OGAN does not have a markedly higher $\overline{S}$ than the other algorithms. The only exception to this is perhaps the $\AFC$ benchmark where all algorithms except OGAN and S-TaLiRo achieve a very low $\overline{S}$. Nevertheless, we interpret the data to indicate that OGAN has efficiency that is comparable to the other algorithms.

Notice that if the OGAN survival functions of Figures \ref{fig:arch23_survival_0} and \ref{fig:arch23_survival_1} are truncated to $300$ executions, the resulting FRs match quite closely those reported in Tables \ref{tbl:arch21_results} and \ref{tbl:alvts_results}. This means that the two sampling strategies considered here ($75$ initial random samples, $P = 0.21$ and $75$ random samples, $P = 0$) yield equivalent behavior up to $300$ executions. Therefore increasing the execution budget and increasing exploration by raising $P$ did not hamper OGAN's ability to falsify with fewer executions than is in the budget.

\begin{table}
\resizebox{\textwidth}{!}{
\begin{tabular}{lrrrrrrrrrrrrrrrrrr}
  \toprule
       & \multicolumn{2}{c}{ARIsTEO:}
       & \multicolumn{2}{c}{Breach:}
       & \multicolumn{2}{c}{FalCAuN:}
       & \multicolumn{2}{c}{falsify:}
       & \multicolumn{2}{c}{FALSTAR:}
       & \multicolumn{2}{c}{FORESEE:}
       & \multicolumn{2}{c}{S-TaLiRo:}
       & \multicolumn{2}{c}{STGEM:}
       & \multicolumn{2}{c}{zlscheck:} \\
       & \multicolumn{2}{c}{ARX-2}
       & \multicolumn{2}{c}{GNM}
       & \multicolumn{2}{c}{}
       & \multicolumn{2}{c}{A3C}
       & \multicolumn{2}{c}{aLVTS}
       & \multicolumn{2}{c}{}
       & \multicolumn{2}{c}{SOAR}
       & \multicolumn{2}{c}{OGAN}
       & \multicolumn{2}{c}{GD} \\
  \midrule \midrule
  Req.
       & \multicolumn{1}{c}{FR} & \multicolumn{1}{c}{$\overline{S}$}
       & \multicolumn{1}{c}{FR} & \multicolumn{1}{c}{$\overline{S}$}
       & \multicolumn{1}{c}{FR} & \multicolumn{1}{c}{$\overline{S}$}
       & \multicolumn{1}{c}{FR} & \multicolumn{1}{c}{$\overline{S}$}
       & \multicolumn{1}{c}{FR} & \multicolumn{1}{c}{$\overline{S}$}
       & \multicolumn{1}{c}{FR} & \multicolumn{1}{c}{$\overline{S}$}
       & \multicolumn{1}{c}{FR} & \multicolumn{1}{c}{$\overline{S}$}
       & \multicolumn{1}{c}{FR} & \multicolumn{1}{c}{$\overline{S}$}
       & \multicolumn{1}{c}{FR} & \multicolumn{1}{c}{$\overline{S}$} \\
  \cmidrule(r){1-1}
  \cmidrule(lr){2-3}
  \cmidrule(lr){4-5}
  \cmidrule(lr){6-7}
  \cmidrule(lr){8-9}
  \cmidrule(lr){10-11}
  \cmidrule(lr){12-13}
  \cmidrule(lr){14-15}
  \cmidrule(lr){16-17}
  \cmidrule(lr){18-19}
  $\textup{AFC27}$ &
  $1.00$ &
  $2.3$ &
  $1.00$ &
  $3.0$ &
  &
  &
  $1.00$ &
  $1.6$ &
  $1.00$ &
  $3.9$ &
  $1.00$ &
  $2.8$ &
  $1.00$ &
  $70.3$ &
  $1.00$ &
  $76.8$ &
  $1.00$ &
  $1.0$ \\
  \cmidrule(r){1-1}
  \cmidrule(lr){2-3}
  \cmidrule(lr){4-5}
  \cmidrule(lr){6-7}
  \cmidrule(lr){8-9}
  \cmidrule(lr){10-11}
  \cmidrule(lr){12-13}
  \cmidrule(lr){14-15}
  \cmidrule(lr){16-17}
  \cmidrule(lr){18-19}
  $\textup{AT1}_{20}$ &
  $0.00$ &
  -- &
  $0.00$ &
  -- &
  $0.92$ &
  $171.2$ &
  $0.78$ &
  $125.8$ &
  $1.00$ &
  $33.0$ &
  $0.58$ &
  $252.3$ &
  $1.00$ &
  $170.3$ &
  $0.82$ &
  $119.1$ &
  $1.00$ &
  $2.3$ \\
  $\textup{AT6}_{4,35,3000}$ &
  $0.94$ &
  $103.1$ &
  $0.00$ &
  -- &
  &
  &
  &
  &
  $1.00$ &
  $76.1$ &
  $0.98$ &
  $117.0$ &
  $0.88$ &
  $130.4$ &
  $1.00$ &
  $88.0$ &
  $1.00$ &
  $42.7$ \\
  $\textup{AT6}_{8,50,3000}$ &
  $0.90$ &
  $164.7$ &
  $0.00$ &
  -- &
  &
  &
  &
  &
  $1.00$ &
  $82.4$ &
  $0.78$ &
  $180.2$ &
  $0.78$ &
  $207.2$ &
  $1.00$ &
  $132.4$ &
  $0.10$ &
  $129.5$ \\
  $\textup{AT6}_{20,65,3000}$ &
  $0.98$ &
  $89.1$ &
  $0.00$ &
  -- &
  $1.00$ &
  $214.1$ &
  &
  &
  $0.00$ &
  -- &
  $0.94$ &
  $124.6$ &
  $0.84$ &
  $197.5$ &
  $0.96$ &
  $92.3$ &
  $0.04$ &
  $261.7$
  \\
  \cmidrule(r){1-1}
  \cmidrule(lr){2-3}
  \cmidrule(lr){4-5}
  \cmidrule(lr){6-7}
  \cmidrule(lr){8-9}
  \cmidrule(lr){10-11}
  \cmidrule(lr){12-13}
  \cmidrule(lr){14-15}
  \cmidrule(lr){16-17}
  \cmidrule(lr){18-19}
  $\textup{CC3}$ &
  $1.00$ &
  $18.9$ &
  $1.00$ &
  $28.6$ &
  &
  &
  $0.88$ &
  $13.5$ &
  $0.08$ &
  $207.5$ &
  $1.00$ &
  $14.4$ &
  $1.00$ &
  $22.4$ &
  $1.00$ &
  $23.7$ &
  $1.00$ &
  $23.4$ \\
  $\textup{CC4}$ &
  $0.00$ &
  -- &
  $0.00$ &
  -- &
  &
  &
  $0.18$ &
  $120.4$ &
  $0.00$ &
  -- &
  $0.26$ &
  $311.1$ &
  $0.00$ &
  -- &
  $0.00$ &
  -- &
  $0.64$ &
  $124.6$ \\
  \cmidrule(r){1-1}
  \cmidrule(lr){2-3}
  \cmidrule(lr){4-5}
  \cmidrule(lr){6-7}
  \cmidrule(lr){8-9}
  \cmidrule(lr){10-11}
  \cmidrule(lr){12-13}
  \cmidrule(lr){14-15}
  \cmidrule(lr){16-17}
  \cmidrule(lr){18-19}
  $\textup{F16}$ &
  $0.00$ &
  -- &
  $0.02$ &
  $297.0$ &
  &
  &
  &
  &
  &
  &
  &
  &
  $0.14$ &
  $127.6$ &
  $0.36$ &
  $183.3$ &
  $0.00$ &
  -- \\
  \cmidrule(r){1-1}
  \cmidrule(lr){2-3}
  \cmidrule(lr){4-5}
  \cmidrule(lr){6-7}
  \cmidrule(lr){8-9}
  \cmidrule(lr){10-11}
  \cmidrule(lr){12-13}
  \cmidrule(lr){14-15}
  \cmidrule(lr){16-17}
  \cmidrule(lr){18-19}
  $\textup{NN}_{0.03}$ &
  $1.00$ &
  $62.8$ &
  $1.00$ &
  $6.0$ &
  &
  &
  $1.00$ &
  $1.00$ &
  $0.52$ &
  $177.0$ &
  $1.00$ &
  $44.2$ &
  $0.98$ &
  $68.0$ &
  $0.98$ &
  $76.0$ &
  $1.00$ &
  $1.3$ \\
  $\textup{NN}_{0.04}$ &
  &
  &
  &
  &
  &
  &
  &
  &
  &
  &
  &
  &
  $0.08$ &
  $193.0$ &
  $0.24$ &
  $168.5$ &
  $1.00$ &
  $1.1$ \\
\end{tabular}
}
\caption{Data is from \cite[Tables~2,3]{ARCH21} except for OGAN. Blanks indicate that data was not available. For the performance of uniform random search, see \autoref{tbl:ogan_results}.}\label{tbl:arch21_results}
\end{table}

\begin{table}
\centering
\begin{tabular}{lrrrrrr}
  \toprule
       & \multicolumn{2}{c}{Breach:}
       & \multicolumn{2}{c}{FALSTAR:}
       & \multicolumn{2}{c}{STGEM:} \\
       & \multicolumn{2}{c}{CMA-ES}
       & \multicolumn{2}{c}{aLVTS}
       & \multicolumn{2}{c}{OGAN} \\
  \midrule \midrule
  Req.
       & \multicolumn{1}{c}{FR} & \multicolumn{1}{c}{$\overline{S}$}
       & \multicolumn{1}{c}{FR} & \multicolumn{1}{c}{$\overline{S}$}
       & \multicolumn{1}{c}{FR} & \multicolumn{1}{c}{$\overline{S}$} \\
  \cmidrule(r){1-1}
  \cmidrule(lr){2-3}
  \cmidrule(lr){4-5}
  \cmidrule(lr){6-7}
  $\textup{ATX2}$ &
  $1.00$ &
  $145.2$ &
  $1.00$ &
  $86.3$ &
  $0.48$ &
  $159.2$
  \\
  $\textup{AT6}_{30,80,4500}$ &
  $1.00$ &
  $97.0$ &
  $1.00$ &
  $22.8$ &
  $0.88$ &
  $103.8$
  \\
  $\textup{AT6}_{30,50,2700}$ &
  $0.98$ &
  $46.7$ &
  $1.00$ &
  $47.6$ &
  $0.92$ &
  $120.2$
\end{tabular}
\caption{Data is from \cite[Table~2]{2021:falsification_of_hybrid_systems_using_adaptive} except for OGAN.}\label{tbl:alvts_results}
\end{table}

\subsubsection{The Answer}
In the preceding two subsections, we argued that OGAN's effectiveness and efficiency is comparable to the best-performing algorithms of ARCH-COMP 2021, \cite{2021:falsification_of_hybrid_systems_using_adaptive}, and ARCH-COMP 2023 on the majority of the considered benchmarks. OGAN is not uniformly best, but neither is any other algorithm. Since the goal of the ARCH-COMP competition is to compare state-of-the-art tools for requirement falsification, we answer RQ1 and RQ2 as follows.

\begin{itemize}
    \item[] \textbf{Answer to RQ1 and RQ2.} OGAN has state-of-the-art requirement falsification efficiency and effectiveness.
\end{itemize}

\subsection{RQ3: The Effect of the Chosen Monte-Carlo Sampling}\label{ssec:mc_sampling}
In this subsection, we consider RQ3 which asks how the choice of the Monte-Carlo sampling strategy affects OGAN's performance. We answer the question by considering two OGAN variants: OGAN US which uses uniform random sampling and OGAN LHS which uses Latin hypercube sampling (LHS). We also report how well pure uniform random search fares on the benchmarks. This provides information on how difficult the benchmarks are.

We use an execution budget of $300$ (used in \cite{ARCH21,2021:falsification_of_hybrid_systems_using_adaptive}) and an initial random sampling budget $N = 75$ ($25 \%$ of $300$). We set the sampling probability $P$ to $0$ which means that we study how the initial random sampling strategy affects OGAN. Other settings are as in \autoref{ssec:ogan_setup}. Notice that the setup for OGAN US is the same as in \autoref{sssec:arch21_comparison}, so the results for OGAN from that section match those of OGAN US. We ran $50$ independent replicas of each falsification task. We used the same random seeds for the uniform random search and OGAN US, which means that the first $75$ executions are identical for both algorithms. 

\begin{figure*}
\centering
\includegraphics[trim=35 5 35 0,clip,width=0.41\textwidth]{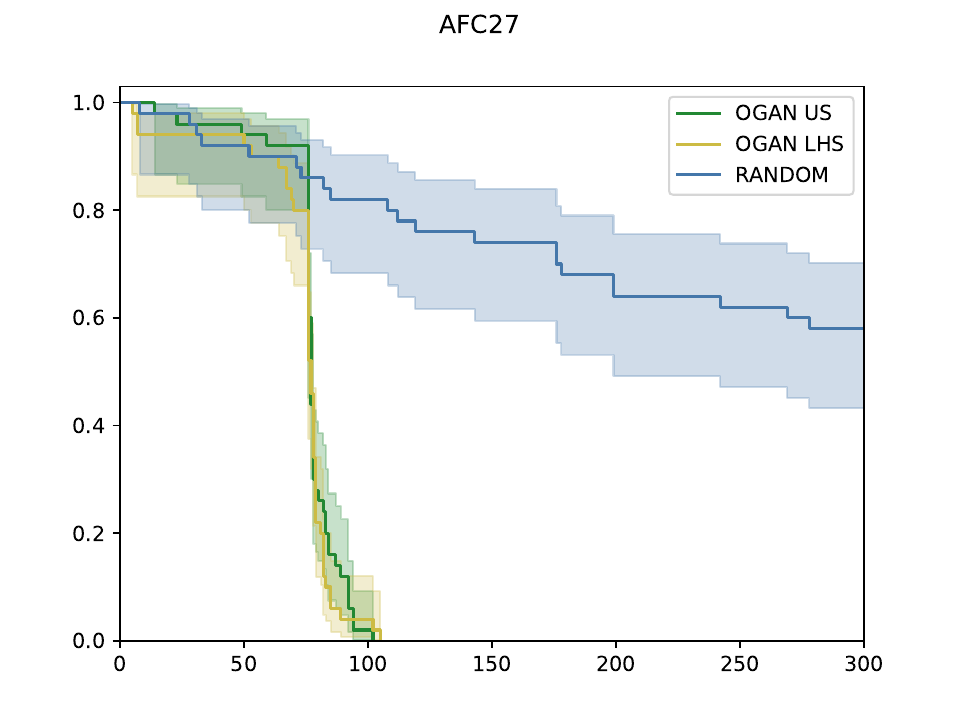}
\includegraphics[trim=35 5 35 0,clip,width=0.41\textwidth]{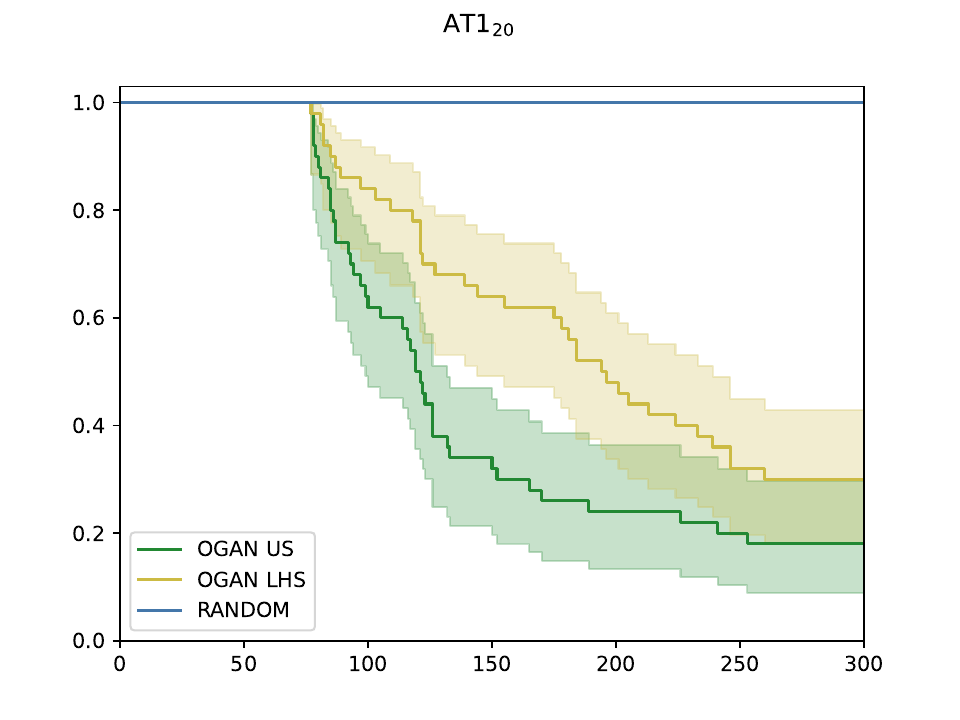}
\includegraphics[trim=35 5 35 0,clip,width=0.41\textwidth]{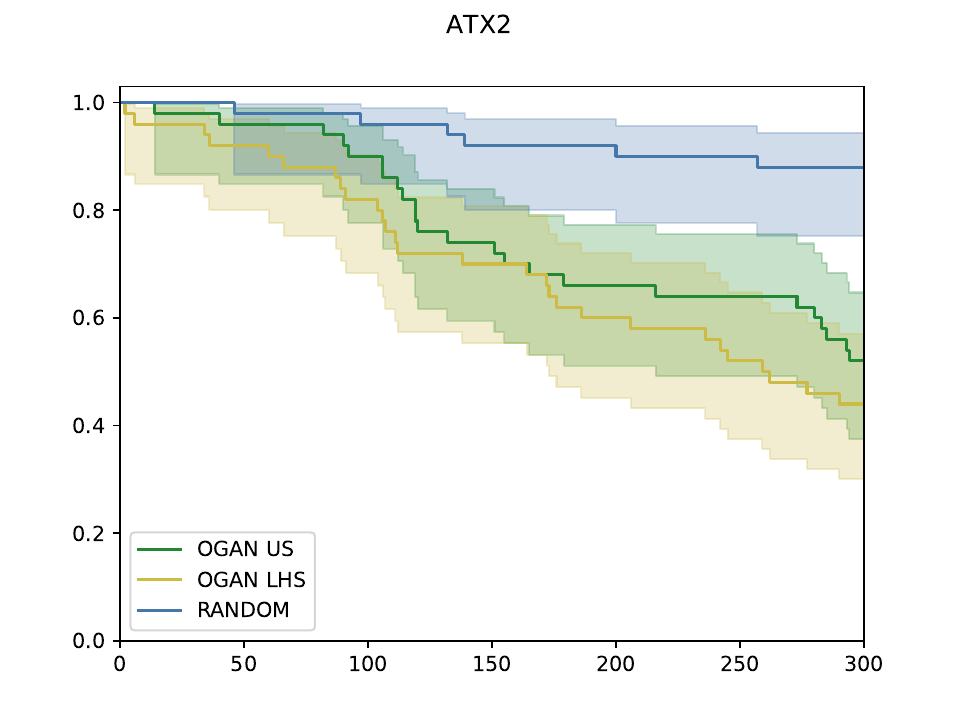}
\includegraphics[trim=35 5 35 0,clip,width=0.41\textwidth]{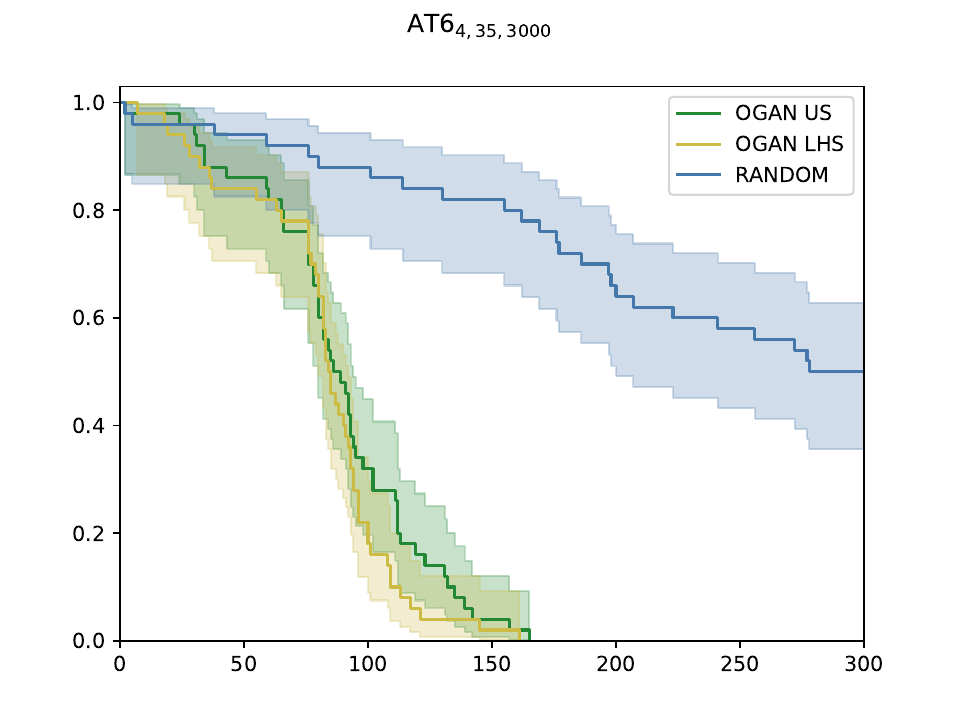}
\includegraphics[trim=35 5 35 0,clip,width=0.41\textwidth]{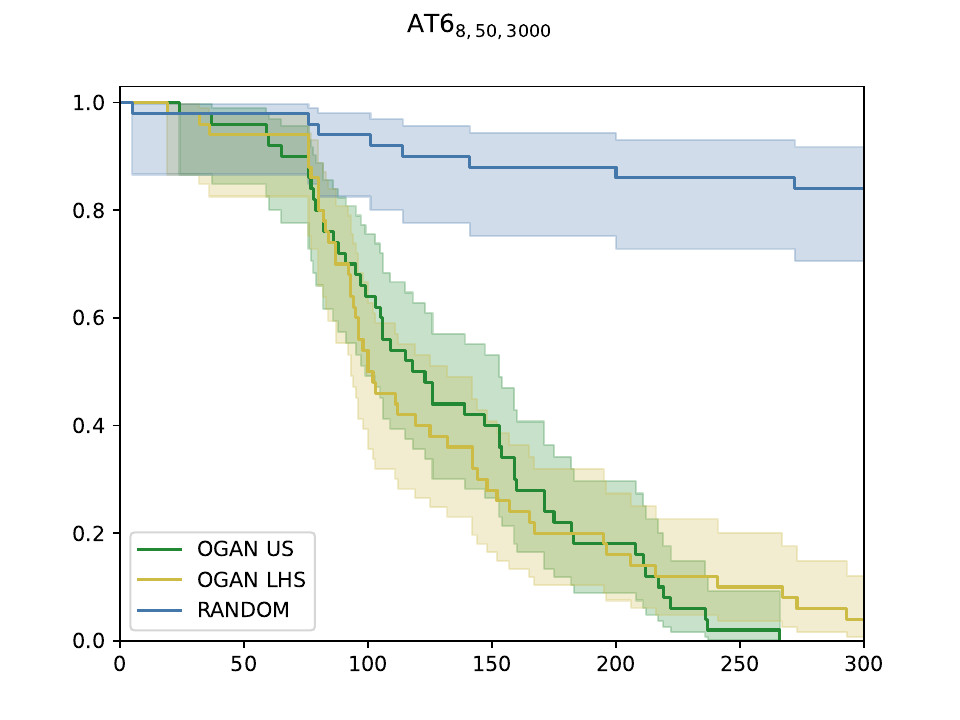}
\includegraphics[trim=35 5 35 0,clip,width=0.41\textwidth]{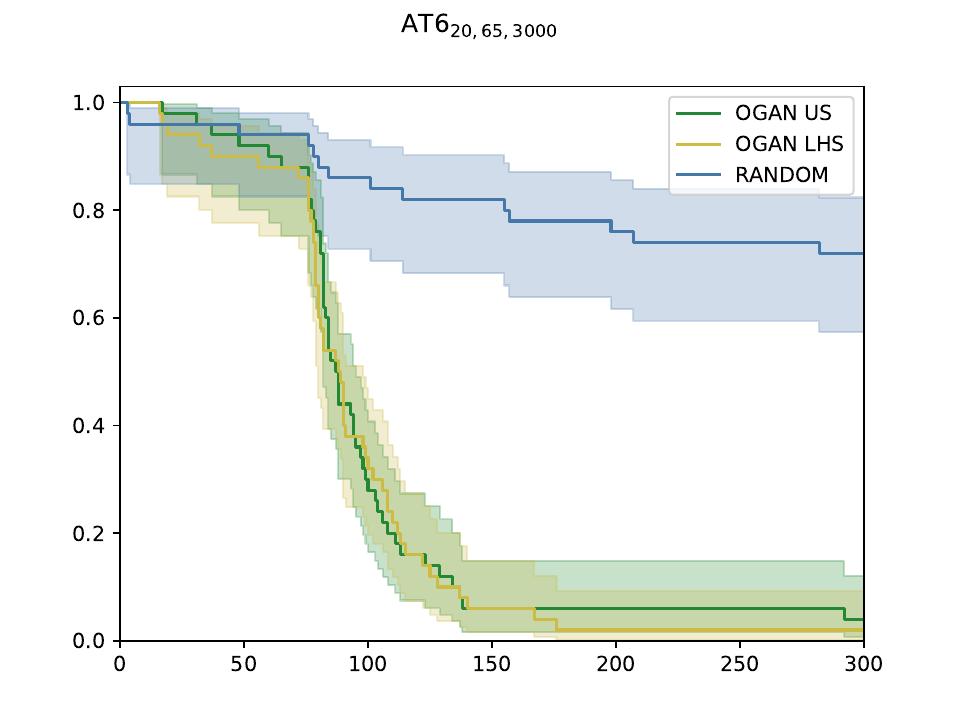}
\includegraphics[trim=35 5 35 0,clip,width=0.41\textwidth]{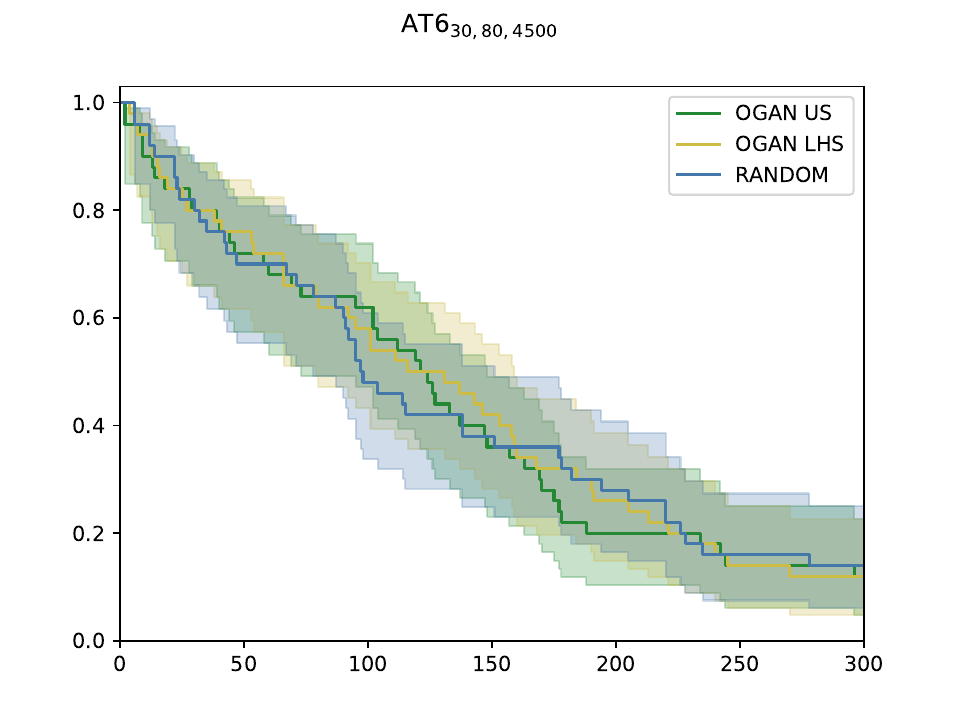}
\includegraphics[trim=35 5 35 0,clip,width=0.41\textwidth]{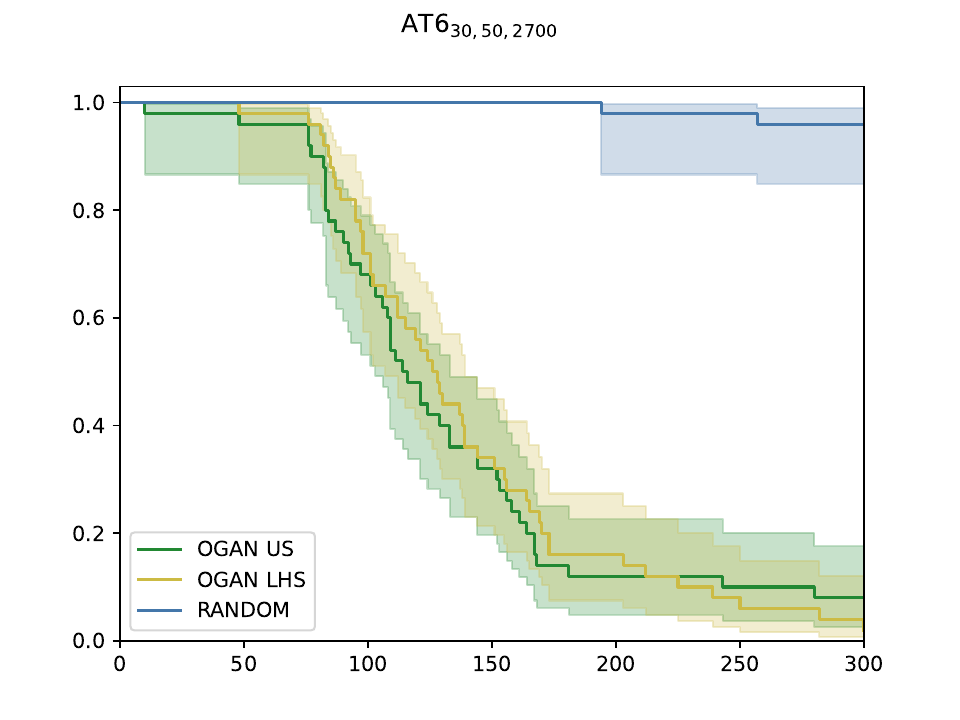}
\caption{Survival functions for RANDOM, OGAN US, and OGAN LHS for the benchmarks of \autoref{ssec:benchmarks}. The light areas correspond to $95 \%$ confidence intervals.}\label{fig:survival_0_1}
\end{figure*}

\begin{figure*}
\centering
\includegraphics[trim=35 5 35 0,clip,width=0.41\textwidth]{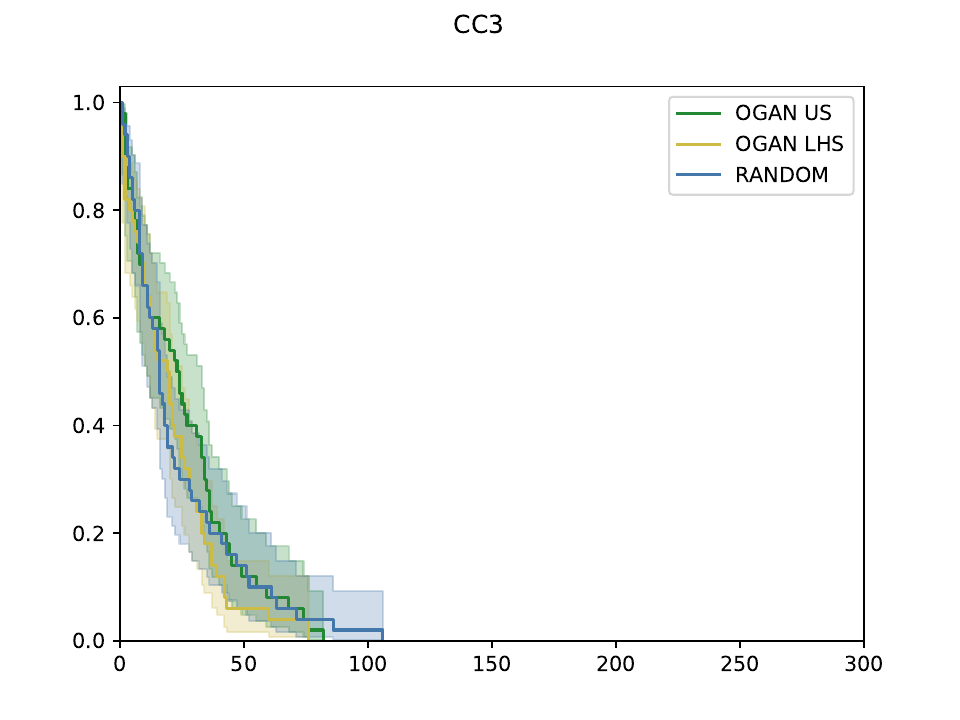}
\includegraphics[trim=35 5 35 0,clip,width=0.41\textwidth]{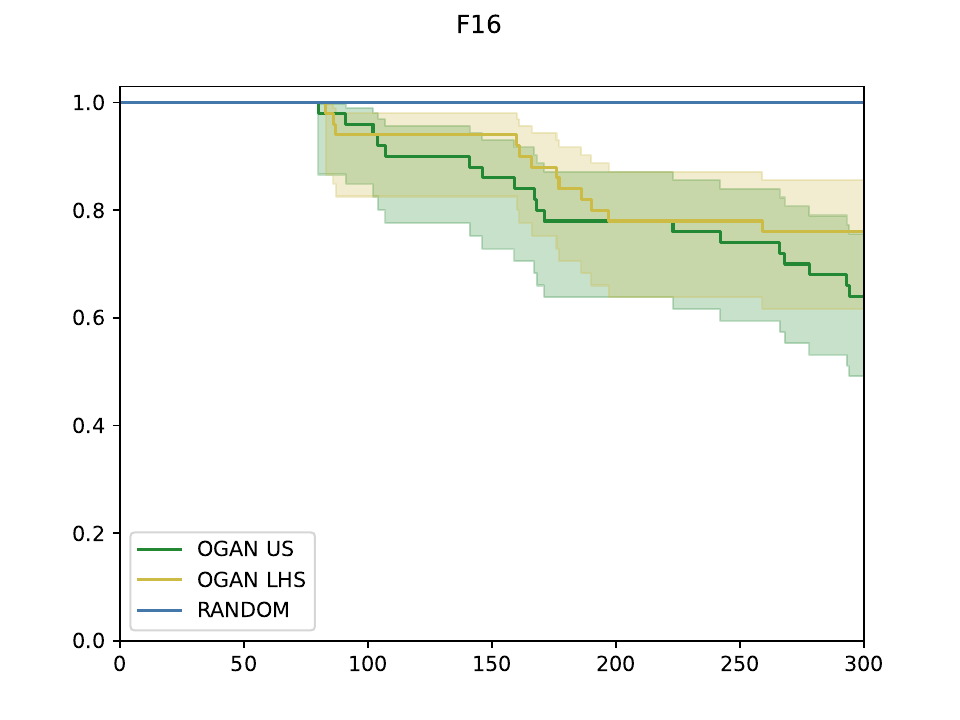}
\includegraphics[trim=35 5 35 0,clip,width=0.41\textwidth]{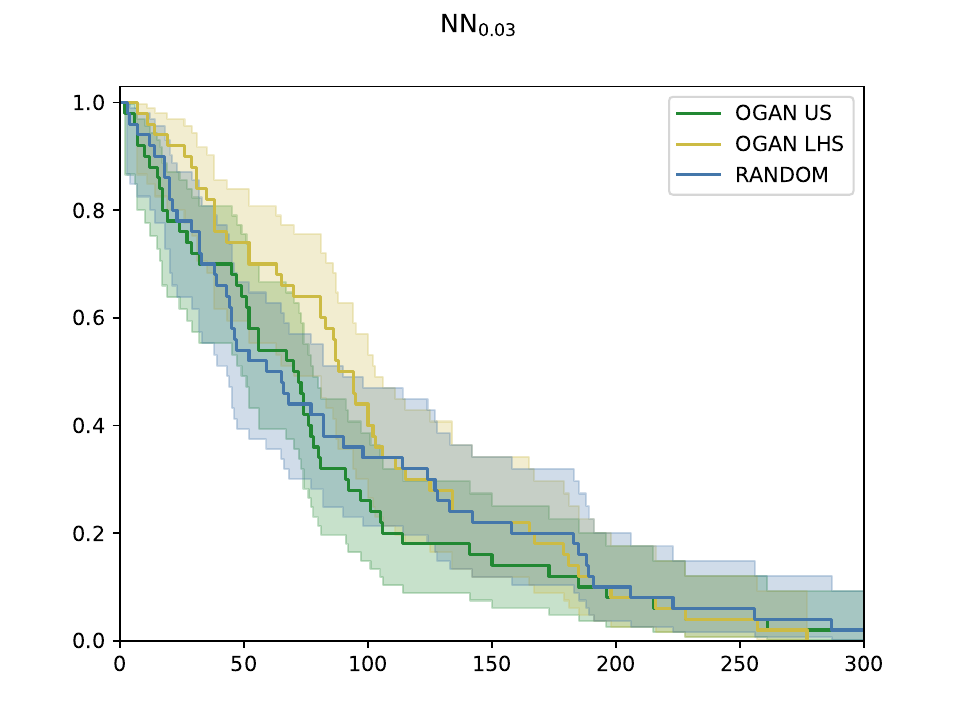}
\includegraphics[trim=35 5 35 0,clip,width=0.41\textwidth]{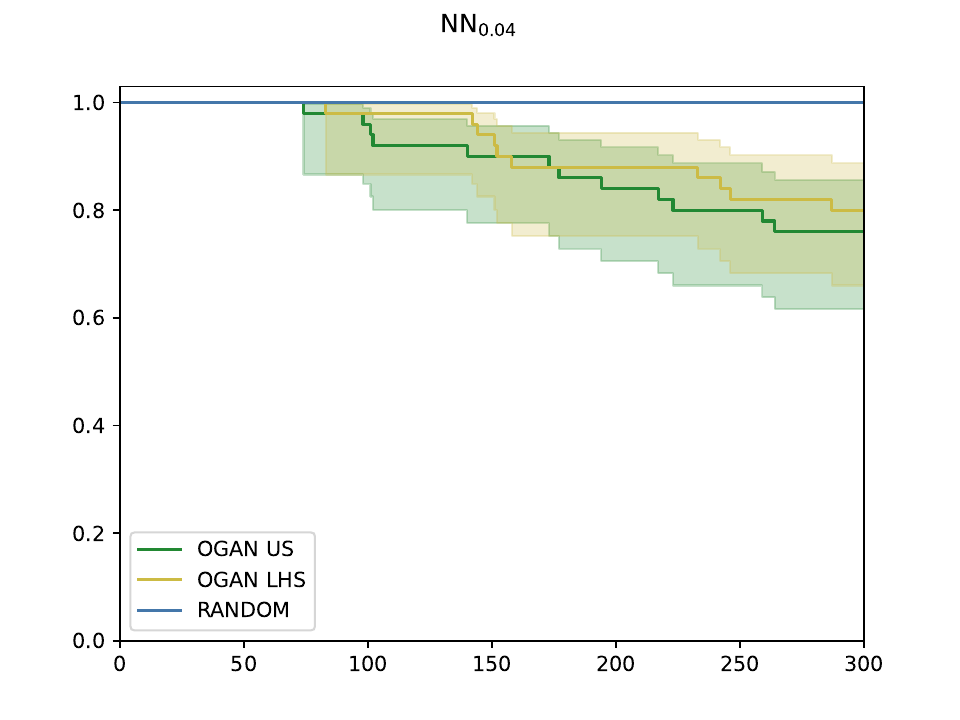}
\includegraphics[trim=35 5 35 0,clip,width=0.41\textwidth]{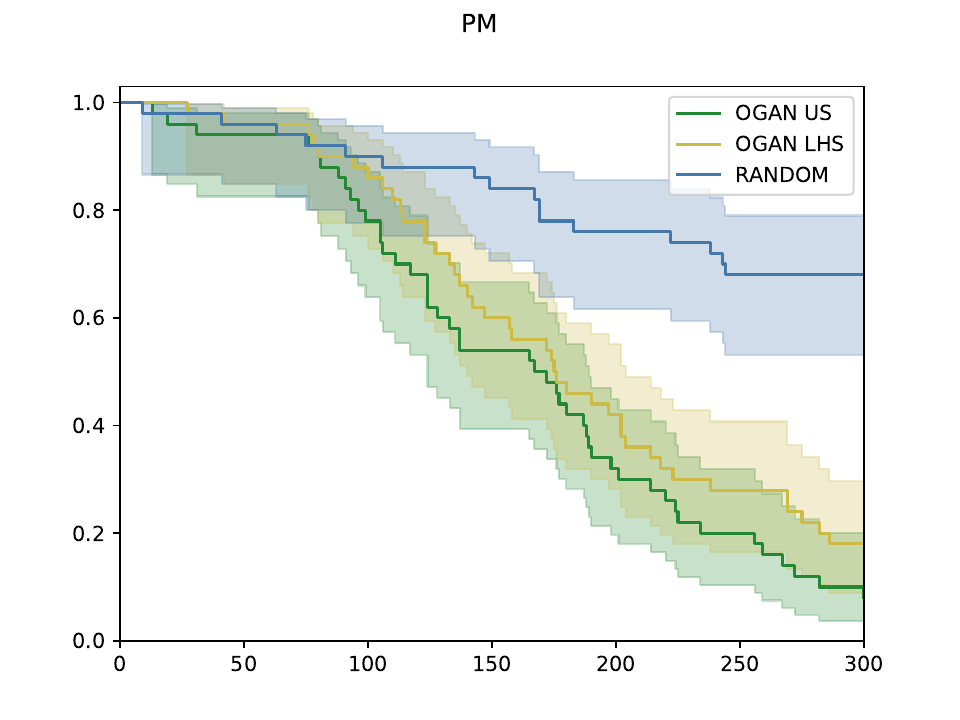}
\caption{\autoref{fig:survival_0_1} continued.}\label{fig:survival_0_2}
\end{figure*}

\autoref{tbl:ogan_results} and Figures \ref{fig:survival_0_1}, \ref{fig:survival_0_2} report the statistics chosen in \autoref{sec:evaluation_methodology} (the survival function for the benchmark $\CCIV$ is omitted as there is no survival functions to display). \autoref{tbl:pvalues_ogan} reports the $p$-values of the logrank test under the null hypothesis that the survival functions of the two OGAN variants represent the same distribution. The column with heading RANDOM reports how uniform random search performs and thus establishes the baseline difficulty of the benchmarks.

{\scriptsize
\begin{table}
\begin{tabular}{lrrrrrrrrr}
  \toprule
       & \multicolumn{3}{c}{STGEM:}
       & \multicolumn{3}{c}{STGEM:}
       & \multicolumn{3}{c}{STGEM:} \\
       & \multicolumn{3}{c}{RANDOM}
       & \multicolumn{3}{c}{OGAN US}
       & \multicolumn{3}{c}{OGAN LHS} \\
  \midrule \midrule
  Req.
       & \multicolumn{1}{c}{FR} & \multicolumn{1}{c}{$95 \%$ CI} & \multicolumn{1}{c}{$\overline{S}$} 
       & \multicolumn{1}{c}{FR} & \multicolumn{1}{c}{$95 \%$ CI} & \multicolumn{1}{c}{$\overline{S}$}
       & \multicolumn{1}{c}{FR} & \multicolumn{1}{c}{$95 \%$ CI} & \multicolumn{1}{c}{$\overline{S}$} \\
  \cmidrule(r){1-1}
  \cmidrule(lr){2-4}
  \cmidrule(lr){5-7}
  \cmidrule(lr){8-10}
  $\textup{AFC27}$ &
  $0.42$ &
  $[0.30, 0.57]$ &
  $126.8$ &
  $1.00$ &
  $[1.00, 1.00]$ &
  $76.8$ &
  $1.00$ &
  $[1.00, 1.00]$ &
  $73.1$ \\
  \cmidrule(r){1-1}
  \cmidrule(lr){2-4}
  \cmidrule(lr){5-7}
  \cmidrule(lr){8-10}
  $\textup{AT1}_{20}$ &
  $0.00$ &
  $[0.00, 0.00]$ &
  -- &
  $0.82$ &
  $[0.70, 0.91]$ &
  $119.1$ &
  $0.70$ &
  $[0.57, 0.82]$ &
  $154.8$ \\
  $\textup{ATX2}$ &
  $0.12$ &
  $[0.06, 0.25]$ &
  $68.2$ &
  $0.48$ &
  $[0.35, 0.63]$ &
  $159.2$ &
  $0.56$ &
  $[0.43, 0.70]$ &
  $144.2$ \\
  $\textup{AT6}_{4,35,3000}$ &
  $0.50$ &
  $[0.37, 0.64]$ &
  $159.2$ &
  $1.00$ &
  $[1.00, 1.00]$ &
  $88.0$ &
  $1.00$ &
  $[1.00, 1.00]$ &
  $81.5$ \\
  $\textup{AT6}_{8,50,3000}$ &
  $0.16$ &
  $[0.08, 0.29]$ &
  $123.6$ &
  $1.00$ &
  $[1.00, 1.00]$ &
  $132.4$ &
  $0.96$ &
  $[0.88, 0.99]$ &
  $123.3$ \\
  $\textup{AT6}_{20,65,3000}$ &
  $0.28$ &
  $[0.18, 0.43]$ &
  $113.4$ &
  $0.96$ &
  $[0.88, 0.99]$ &
  $92.3$ &
  $0.98$ &
  $[0.91, 1.00]$ &
  $89.6$ \\
  $\textup{AT6}_{30,80,4500}$ &
  $0.86$ &
  $[0.75, 0.94]$ &
  $103.1$ &
  $0.88$ &
  $[0.77, 0.95]$ &
  $105.7$ &
  $0.88$ &
  $[0.77, 0.95]$ &
  $110.1$ \\
  $\textup{AT6}_{30,50,2700}$ &
  $0.04$ &
  $[0.01, 0.15]$ &
  $225.5$ &
  $0.92$ &
  $[0.82, 0.97]$ &
  $120.2$ &
  $0.98$ &
  $[0.91, 1.00]$ &
  $137.0$ \\
  \cmidrule(r){1-1}
  \cmidrule(lr){2-4}
  \cmidrule(lr){5-7}
  \cmidrule(lr){8-10}
  $\textup{CC3}$ &
  $1.00$ &
  $[1.00, 1.00]$ &
  $23.5$ &
  $1.00$ &
  $[1.00, 1.00]$ &
  $25.7$ &
  $1.00$ &
  $[1.00, 1.00]$ &
  $21.0$ \\
  $\textup{CC4}$ &
  $0.00$ &
  $[0.00, 0.00]$ &
  -- &
  $0.00$ &
  $[0.00, 0.00]$ &
  -- &
  $0.00$ &
  $[0.00, 0.00]$ &
  -- \\
  \cmidrule(r){1-1}
  \cmidrule(lr){2-4}
  \cmidrule(lr){5-7}
  \cmidrule(lr){8-10}
  $\textup{F16}$ &
  $0.00$ &
  $[0.00, 0.00]$ &
  -- &
  $0.36$ &
  $[0.24, 0.51]$ &
  $183.3$ &
  $0.24$ &
  $[0.14, 0.38]$ &
  $160.7$ \\
  \cmidrule(r){1-1}
  \cmidrule(lr){2-4}
  \cmidrule(lr){5-7}
  \cmidrule(lr){8-10}
  $\textup{NN}_{0.03}$ &
  $0.86$ &
  $[0.75, 0.94]$ &
  $85.8$ &
  $0.98$ &
  $[0.91, 1.00]$ &
  $76.0$ &
  $1.00$ &
  $[1.00, 1.00]$ &
  $101.1$ \\
  $\textup{NN}_{0.04}$ &
  $0.00$ &
  $[0.00, 0.00]$ &
  -- &
  $0.24$ &
  $[0.14, 0.38]$ &
  $168.5$ &
  $0.20$ &
  $[0.11, 0.34]$ &
  $183.8$ \\
  \cmidrule(r){1-1}
  \cmidrule(lr){2-4}
  \cmidrule(lr){5-7}
  \cmidrule(lr){8-10}
  $\textup{PM}$ &
  $0.32$ &
  $[0.21, 0.47]$ &
  $144.5$ &
  $0.92$ &
  $[0.82, 0.97]$ &
  $155.7$ &
  $0.82$ &
  $[0.70, 0.91]$ &
  $161.2$
\end{tabular}
\caption{Falsification results over $50$ independent replicas for uniform random search, OGAN with uniform random sampling as initial random search, and OGAN with Latin hypercube sampling as initial random search for the benchmarks of \autoref{ssec:benchmarks}.}\label{tbl:ogan_results}
\end{table}
}

\begin{table}
\begin{minipage}{.49\textwidth}
\centering
{\setlength{\tabcolsep}{14pt}
\begin{tabular}{lc}
  Req. & $p$-value \\
  \cmidrule(r){1-1}
  \cmidrule(lr){2-2}
  $\textup{AFC27}$ & $0.428$ \\
  \cmidrule(r){1-1}
  \cmidrule(lr){2-2}
  $\textup{AT1}_{20}$ & $0.014$ \\
  $\textup{ATX2}$ & $0.337$ \\
  $\textup{AT6}_{4,35,3000}$ & $0.208$ \\
  $\textup{AT6}_{8,50,3000}$ & $0.822$ \\
  $\textup{AT6}_{20,65,3000}$ & $0.814$ \\
  $\textup{AT6}_{30,80,4500}$ & $0.848$ \\
  $\textup{AT6}_{30,50,2700}$ & $0.889$ \\
  \cmidrule(r){1-1}
  \cmidrule(lr){2-2}
  $\textup{CC3}$ & $0.218$ \\
  $\textup{CC4}$ & $1.000$ \\
  \cmidrule(r){1-1}
  \cmidrule(lr){2-2}
  $\textup{F16}$ & $0.217$ \\
  \cmidrule(r){1-1}
  \cmidrule(lr){2-2}
  $\textup{NN}_{0.03}$ & $0.209$ \\
  $\textup{NN}_{0.04}$ & $0.605$ \\
  \cmidrule(r){1-1}
  \cmidrule(lr){2-2}
  $\textup{PM}$ & $0.157$
\end{tabular}
\caption{Log-rank test $p$-values under the hypothesis that the survival functions of OGAN US and OGAN LHS represent the same distribution.}\label{tbl:pvalues_ogan}
}%
\end{minipage}%
\hfill%
\begin{minipage}{.49\textwidth}
\centering
{\setlength{\tabcolsep}{14pt}
\begin{tabular}{lcc}
  \toprule
       & US:
       & LHS: \\
  \midrule
  \midrule
  Req. & $p$-value & $p$-value \\
  \cmidrule(r){1-1}
  \cmidrule(lr){2-2}
  \cmidrule(lr){3-3}
  $\textup{AFC27}$ & $0.98$ & $0.93$ \\
  \cmidrule(r){1-1}
  \cmidrule(lr){2-2}
  \cmidrule(lr){3-3}
  $\textup{AT1}_{20}$ & $0.79$ & $0.02$ \\
  $\textup{ATX2}$ & $0.00$ & $0.00$ \\
  $\textup{AT6}_{4,35,3000}$ & $0.39$ & $0.03$  \\
  $\textup{AT6}_{8,50,3000}$ & $0.54$ & $0.84$  \\
  $\textup{AT6}_{20,65,3000}$ & $0.85$ & $0.70$ \\
  $\textup{AT6}_{30,80,4500}$ & $0.00$ & $0.01$ \\
  $\textup{AT6}_{30,50,2700}$ & $0.22$ & $0.30$ \\
  \cmidrule(r){1-1}
  \cmidrule(lr){2-2}
  \cmidrule(lr){3-3}
  $\textup{CC3}$ & $0.92$ & $0.65$ \\
  $\textup{CC4}$ & $1.00$ & $1.00$ \\
  \cmidrule(r){1-1}
  \cmidrule(lr){2-2}
  \cmidrule(lr){3-3}
  $\textup{F16}$ & $0.08$ & $0.19$ \\
  \cmidrule(r){1-1}
  \cmidrule(lr){2-2}
  \cmidrule(lr){3-3}
  $\textup{NN}_{0.03}$ & $0.99$ & $0.05$ \\
  $\textup{NN}_{0.04}$ & $0.95$ & $0.15$ \\
  \cmidrule(r){1-1}
  \cmidrule(lr){2-2}
  \cmidrule(lr){3-3}
  $\textup{PM}$ & $0.04$ & $0.72$
\end{tabular}
\caption{Log-rank test $p$-values under the hypothesis that the survival functions of adaptive OGAN and nonadaptive OGAN represent the same distribution.}\label{tbl:pvalues}
}%
\end{minipage}
\end{table}

It is immediate that OGAN consistently beats uniform random search by a large margin in effectiveness and efficiency. The only exceptions are the benchmarks $\ATXVIA$, $\CCIII$, and $\NN_{0.03}$ where all three algorithms have equal performance as indicated by Figures \ref{fig:survival_0_1}, \ref{fig:survival_0_2}. This observation is supported by the fact that the corresponding logrank test $p$-values under the hypothesis that OGAN US and random search have identical survival functions are $0.869$, $0.736$, and $0.608$. When comparing OGAN LHS and random search, the respective $p$-values are $0.984$, $0.580$, and $0.738$. Nevertheless, OGAN always performs as well as uniform random search and typically much better.

As is evident from \autoref{tbl:pvalues_ogan}, OGAN US and OGAN LHS yield similar performance. Only in the benchmark $\ATI$ is there a marked difference. Given that the performance is similar on the other benchmarks, this is likely explained by the characteristics of this benchmark, but we have not studied it further. We conclude that using LHS over uniform random sampling does not yield a marked difference in OGAN's performance.

\begin{itemize}
    \item[] \textbf{Answer to RQ3.} OGAN performance is not affected significantly by the chosen Monte-Carlo sampling strategy.
\end{itemize}

\subsection{RQ4: OGAN Computational Overhead}
In order to study RQ4, we have collected the total time $T$ to complete each replica and the times $t_G$, $t_T$, and $t_E$ which respectively are the time to generate a single test, model training time per each generated test, and test execution time (including both the SUT execution and the robustness computation). We expect $t_G$ to be negligible for uniform random search algorithm ($t_T$ is zero). For OGAN, the time $t_G$ corresponds to the execution of the function SAMPLE\_TEST of \autoref{alg1}, and we expect it to be higher than $t_G$ for uniform random search but still lower than $t_T$ for OGAN due to the training of neural networks. For OGAN, the numbers $t_G$ and $t_T$ are influenced by the chosen neural network architectures and the dimensions of the latent space and the input space (larger dimensions are expected to increase the times). The training time $t_T$ also depends on the number of training data, so $t_T$ tends to be higher on benchmarks that take longer to falsify in terms of executed tests. The test execution time $t_E$ is independent of the used falsification algorithm. Whether or not the OGAN computational overhead $t_G + t_T$ is relevant depends on $t_E$. To make the execution time more concrete, we report the ratio $R$ defined as the ratio of the total time used for execution and the total time (totals are computed over all replicas). The data we use is from the replicas of \autoref{ssec:mc_sampling}.

{\scriptsize
\begin{table}
\centering
\begin{tabular}{lrrrrrrrrrr}
  \toprule
       & \multicolumn{5}{c}{STGEM:}
       & \multicolumn{5}{c}{STGEM:} \\
       & \multicolumn{5}{c}{RANDOM}
       & \multicolumn{5}{c}{OGAN US} \\
  \midrule \midrule
  Req.
       & \multicolumn{1}{c}{$\overline{t}_G$} & \multicolumn{1}{c}{$\overline{t}_T$} & \multicolumn{1}{c}{$\overline{t}_E$} & \multicolumn{1}{c}{$\overline{T}$} & \multicolumn{1}{c}{$R$}
       & \multicolumn{1}{c}{$\overline{t}_G$} & \multicolumn{1}{c}{$\overline{t}_T$} & \multicolumn{1}{c}{$\overline{t}_E$} & \multicolumn{1}{c}{$\overline{T}$} & \multicolumn{1}{c}{$R$} \\
  \cmidrule(r){1-1}
  \cmidrule(lr){2-6}
  \cmidrule(lr){7-11}
  $\textup{AFC27}$ &
  $*$ &
  $*$ &
  $1.96$ &
  $444.6$ &
  $1.00$ &
  $0.00$ &
  $0.12$ &
  $2.38$ &
  $192.5$ &
  $0.95$ \\
  \cmidrule(r){1-1}
  \cmidrule(lr){2-6}
  \cmidrule(lr){7-11}
  $\textup{AT1}_{20}$ &
  $*$ &
  $*$ &
  $0.47$ &
  $142.0$ &
  $1.00$ &
  $0.00$ &
  $1.22$ &
  $0.67$ &
  $286.8$ &
  $0.35$ \\
  $\textup{ATX2}$ &
  $*$ &
  $*$ &
  $0.49$ &
  $137.5$ &
  $1.00$ &
  $0.00$ &
  $1.85$ &
  $0.54$ &
  $557.0$ &
  $0.23$ \\
  $\textup{AT6}_{4,35,3000}$ &
  $*$ &
  $*$ &
  $0.53$ &
  $122.4$ &
  $1.00$ &
  $0.00$ &
  $0.53$ &
  $0.95$ &
  $130.3$ &
  $0.64$ \\
  $\textup{AT6}_{8,50,3000}$ &
  $*$ &
  $*$ &
  $0.49$ &
  $134.5$ &
  $1.00$ &
  $0.01$ &
  $1.07$ &
  $0.71$ &
  $236.0$ &
  $0.40$ \\
  $\textup{AT6}_{20,65,3000}$ &
  $*$ &
  $*$ &
  $0.51$ &
  $127.2$ &
  $1.00$ &
  $0.00$ &
  $0.70$ &
  $0.83$ &
  $154.5$ &
  $0.54$ \\
  $\textup{AT6}_{30,80,4500}$ &
  $*$ &
  $*$ &
  $0.70$ &
  $91.5$ &
  $1.00$ &
  $0.00$ &
  $1.36$ &
  $0.72$ &
  $269.3$ &
  $0.34$ \\
  $\textup{AT6}_{30,50,2700}$ &
  $*$ &
  $*$ &
  $0.48$ &
  $141.1$ &
  $1.00$ &
  $0.01$ &
  $1.11$ &
  $0.70$ &
  $244.6$ &
  $0.38$ \\
  \cmidrule(r){1-1}
  \cmidrule(lr){2-6}
  \cmidrule(lr){7-11}
  $\textup{CC3}$ &
  $*$ &
  $*$ &
  $2.89$ &
  $67.9$ &
  $1.00$ &
  $*$ &
  $0.01$ &
  $2.73$ &
  $70.6$ &
  $1.00$ \\
  $\textup{CC4}$ &
  $*$ &
  $*$ &
  $1.09$ &
  $327.2$ &
  $1.00$ &
  $0.00$ &
  $3.27$ &
  $1.23$ &
  $1348.6$ &
  $0.27$ \\
  \cmidrule(r){1-1}
  \cmidrule(lr){2-6}
  \cmidrule(lr){7-11}
  $\textup{F16}$ &
  $*$ &
  $*$ &
  $0.08$ &
  $25.0$ &
  $1.00$ &
  $0.00$ &
  $1.20$ &
  $0.09$ &
  $334.3$ &
  $0.07$ \\
  \cmidrule(r){1-1}
  \cmidrule(lr){2-6}
  \cmidrule(lr){7-11}
  $\textup{NN}_{0.03}$ &
  $*$ &
  $*$ &
  $1.36$ &
  $122.3$ &
  $1.00$ &
  $0.00$ &
  $0.92$ &
  $1.42$ &
  $188.1$ &
  $0.61$ \\
  $\textup{NN}_{0.04}$ &
  $*$ &
  $*$ &
  $1.07$ &
  $320.3$ &
  $1.00$ &
  $0.00$ &
  $2.04$ &
  $1.12$ &
  $847.8$ &
  $0.35$ \\
  \cmidrule(r){1-1}
  \cmidrule(lr){2-6}
  \cmidrule(lr){7-11}
  $\textup{PM}$ &
  $*$ &
  $*$ &
  $0.50$ &
  $126.1$ &
  $1.00$ &
  $0.00$ &
  $1.15$ &
  $0.59$ &
  $291.9$ &
  $0.34$
\end{tabular}
\caption{Time measurements over $50$ independent replicas for uniform random search and OGAN US. We report the means $\overline{t}_G$, $\overline{t}_T$, $\overline{t}_E$ of test generation time, training time, and execution time, the mean total time $\overline{T}$, and the ratio $R$ of total execution time to the sum of total times. All times are measured in wall time and in seconds. Asterisk $*$ indicates that the measurement was smaller than $\num{1e-4}$.}\label{tbl:ogan_time_results}
\end{table}
}

The collected time measurements are displayed in \autoref{tbl:ogan_time_results}. As expected, the uniform random search algorithm spends all of its time executing tests, and its computational overhead is negligible. It is clear that OGAN has a significant computational overhead on almost all benchmarks. The overheads for $\AFC$ and $\CCIII$ are low, but this is explained by the fact that often already the initial random sampling manages to find a falsifying input: the neural networks are trained only a few times, and this accounts to little overhead. It is worth remarking that on the $\AFC$ benchmark OGAN manages to achieve a significantly lower mean total time than the random search algorithm. We believe this is because it is very easy for the discriminator to learn from its training data what is a falsifying input. Thus OGAN gains a significant advantage.

The ratios $R$ indicate that typically OGAN uses less than half of the time for test execution. Since we have assumed that test execution is expensive, it follows that using OGAN is even more expensive timewise. Thus if time is the main concern, then OGAN might not be the tool of choice for the selected benchmarks. However, notice that the execution time is independent of the OGAN algorithm. If the execution took several minutes, then the ratio $R$ would be automatically very high provided that the OGAN setup is not altered. This would make OGAN a much more appealing approach. If effectiveness and efficiency are also important, as they should be, OGAN is a viable choice even when SUT execution is fast. In general, it is difficult to meaningfully compare computational overheads between algorithms that are not equally effective and efficient.

\begin{itemize}
    \item[] \textbf{Answer to RQ4.} OGAN incurs a significant computational overhead over random search on the selected benchmarks. This is mainly due to the training of neural networks. The severity of the overhead depends on the SUT execution time, which is independent of OGAN.
\end{itemize}

\subsection{RQ5: OGAN Discriminator Accuracy}
The OGAN discriminator $\discriminator$ functions as a surrogate model for the composition of the SUT $\sut$ and the robustness metric $\rho$. Ideally, the discriminator accurately models the mapping $x \mapsto \rho(\sut(x))$ where $x$ is in the input space $\inputspace$. This is, however, not a strict requirement for successful falsification as the discussion of \autoref{ssec:falsification_generative_models} merely assumes that low-robustness areas of the input space are modeled sufficiently well. In this section, we study the accuracy of the OGAN discriminators learned during the experiments of \autoref{ssec:mc_sampling} and answer RQ5.

Consider a benchmark from \autoref{ssec:benchmarks}. Let $\mathcal{B}$ be a balanced validation set, obtained
independently of the results of \autoref{ssec:mc_sampling}, such that, for all $i = 0, \ldots, 9$, the size of the set
\begin{equation*}
  \left\{ x \in \mathcal{B} : \rho(\sut(x)) \in \left[\frac{M}{10} i, \frac{M}{10} (i+1)\right) \right\}
\end{equation*}
is $25$ whenever it is nonempty (nonemptiness is determined empirically based on $10000$ uniformly randomly selected
inputs). Here $M$ is the empirical maximum of $\rho \circ \sut$ computed from $10000$ uniformly randomly selected
inputs. For a given OGAN discriminator $\discriminator$, we compute an accuracy score $M(\discriminator)$ as the mean absolute
error:
\begin{equation*}
  M(\discriminator) = \frac{1}{|\mathcal{B}|} \sum_{x \in \mathcal{B}} |\discriminator(x) - \rho(\sut(x))|.
\end{equation*}
Observe that the loss of \autoref{ssec:ogan_setup} that was used to train the discriminators is different from the mean absolute error. We use the mean absolute error here as it is more interpretable.

\autoref{tbl:mae} reports the means and standard errors for the accuracy scores for each benchmark from
\autoref{ssec:benchmarks} and both OGAN variants of \autoref{ssec:mc_sampling}. The selected discriminators were the OGAN discriminators trained when the falsification
tasks of \autoref{ssec:mc_sampling} terminated. Thus a sample size of $50$ discriminators was used to compute each statistic. It
is evident that typically the mean absolute error is rather large especially given that the
range of $\rho \circ \sut$ is $[0,M]$. The exceptional results for the benchmark $\textup{CC4}$ are explained by the
fact that the empirical maximum $0.008$ is a small number. Notice also that the benchmark $\textup{CC3}$ is, on average, falsified by the initial random search. It follows that in this case the majority of the discriminators were never trained, and the scores mainly reflect the accuracy of a discriminator with randomly initialized weights.

\begin{table}
\centering
\begin{tabular}{lcccc}
  \toprule
       & \multicolumn{2}{c}{OGAN US:}
       & \multicolumn{2}{c}{OGAN LHS:} \\
  \midrule
  \midrule
  Req. & \multicolumn{1}{c}{Mean} & \multicolumn{1}{c}{SD} & \multicolumn{1}{c}{Mean} & \multicolumn{1}{c}{SD} \\
  \cmidrule(r){1-1}
  \cmidrule(lr){2-3}
  \cmidrule(lr){4-5}
  $\textup{AFC27}$ & $0.28$ & $0.02$ & $0.27$ & $0.02$ \\
  \cmidrule(r){1-1}
  \cmidrule(lr){2-3}
  \cmidrule(lr){4-5}
  $\textup{AT1}_{20}$ & $0.13$ & $0.03$ & $0.14$ & $0.04$ \\
  $\textup{ATX2}$ & $0.21$ & $0.01$ & $0.21$ & $0.01$ \\
  $\textup{AT6}_{4,35,3000}$ & $0.29$ & $0.04$ & $0.30$ & $0.04$ \\
  $\textup{AT6}_{8,50,3000}$ & $0.29$ & $0.03$ & $0.30$ & $0.03$ \\
  $\textup{AT6}_{20,65,3000}$ & $0.30$ & $0.03$ & $0.29$ & $0.04$ \\
  $\textup{AT6}_{30,80,4500}$ & $0.21$ & $0.01$ & $0.21$ & $0.01$ \\
  $\textup{AT6}_{30,50,2700}$ & $0.30$ & $0.04$ & $0.30$ & $0.04$ \\
  \cmidrule(r){1-1}
  \cmidrule(lr){2-3}
  \cmidrule(lr){4-5}
  $\textup{CC3}$ & $0.46$ & $0.11$ & $0.46$ & $0.11$ \\
  $\textup{CC4}$ & $0.00$ & $0.00$ & $0.00$ & $0.00$ \\
  \cmidrule(r){1-1}
  \cmidrule(lr){2-3}
  \cmidrule(lr){4-5}
  $\textup{F16}$ & $0.05$ & $0.03$ & $0.05$ & $0.03$ \\
  \cmidrule(r){1-1}
  \cmidrule(lr){2-3}
  \cmidrule(lr){4-5}
  $\textup{NN}_{0.03}$ & $0.24$ & $0.03$ & $0.26$ & $0.03$ \\
  $\textup{NN}_{0.04}$ & $0.26$ & $0.02$ & $0.27$ & $0.02$ \\
  \cmidrule(r){1-1}
  \cmidrule(lr){2-3}
  \cmidrule(lr){4-5}
  $\textup{PM}$ & $0.14$ & $0.06$ & $0.14$ & $0.05$
\end{tabular}
\caption{Averages and standard deviations of the mean absolute errors for the OGAN discriminators from
\autoref{ssec:mc_sampling}.}\label{tbl:mae}
\end{table}

We conclude that the OGAN discriminators trained with a limited budget (300 tests) are not globally accurate models, but they model the low-robustness regions well-enough since they perform better than random search as explained in \autoref{ssec:mc_sampling}. Manual inspection of several discriminators reveals that discriminator estimates are often biased in the sense that their range is much narrower than the expected range of $[0,1]$. Since falsification nevertheless succeeds, this is not a problem: it is enough that the lower end of the discriminator range maps reasonably well to a low-robustness area of $\rho \circ \sut$.

\begin{itemize}
    \item[] \textbf{Answer to RQ5.} The trained discriminators are not globally accurate models. However, they estimate the robustness well enough to guide OGAN towards falsifying inputs.
\end{itemize}

\subsection{RQ6: The Role of Generator Sampling}\label{ssec:survival}
In \autoref{ssec:falsification_generative_models}, we proposed to use the generator to augment the discriminator's training data. Here we study the effect of this compared to using Monte-Carlo sampling independent of the generator. We do this by considering a variant of OGAN, which we call nonadaptive OGAN, and compare it to the unmodified OGAN which we call here adaptive OGAN. Otherwise we use the benchmarks and setup of \autoref{sssec:arch21_comparison}. We consider two Monte-Carlo sampling strategies: uniform random sampling and Latin hypercube sampling. To be clear, the sampling probability $P$ is set to $0$ in adaptive OGAN.

In the nonadaptive variant, we sample two tests to be executed: test $t_1$ using the first branch of the if statement in the function $\textup{SAMPLE\_TEST}$ of \autoref{alg1} and another test $t_2$ using the other branch. The test $t_1$ and its robustness are added to the discriminator's training data whereas $t_2$ and its robustness are saved but not used for training purposes. After the algorithm has terminated, we report the number of executions needed for a falsification based on the tests $t_2$ not the tests $t_1$. Essentially, the training of the discriminator is based on randomly sampled tests and the success of the falsification is determined by the tests sampled from the generator. We emphasize that we keep the execution budget the same but count the execution of the tests $t_1$ and $t_2$ as a single execution. Without this, the comparison cannot be meaningfully done. We also remark that the initial random search phase is the same in both variants.

\begin{figure*}
\centering
\includegraphics[trim=35 5 35 0,clip,width=0.41\textwidth]{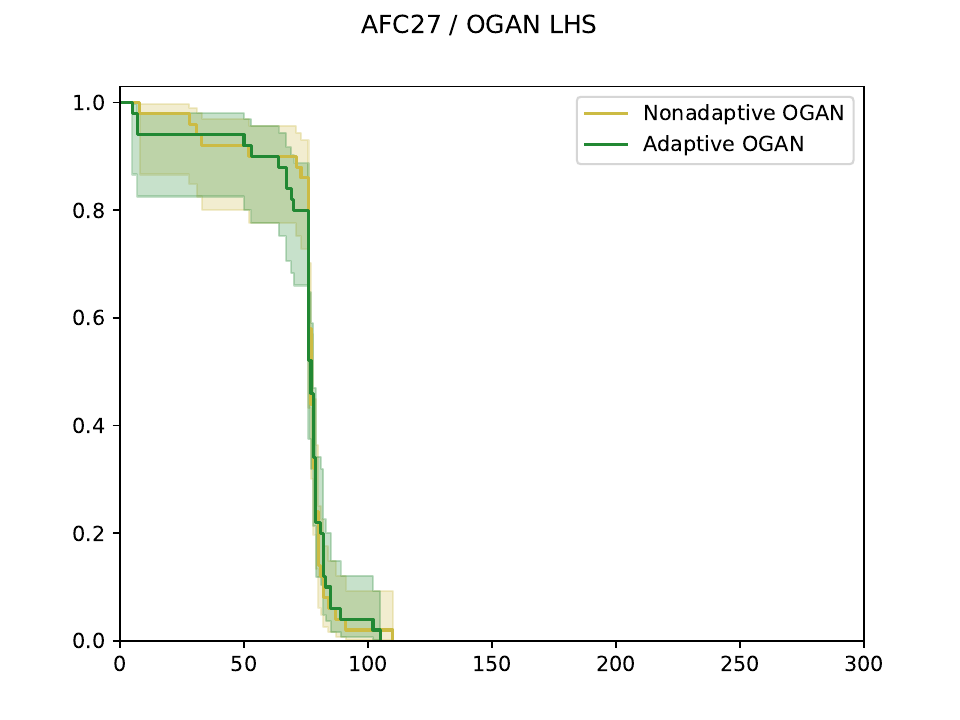}
\includegraphics[trim=35 5 35 0,clip,width=0.41\textwidth]{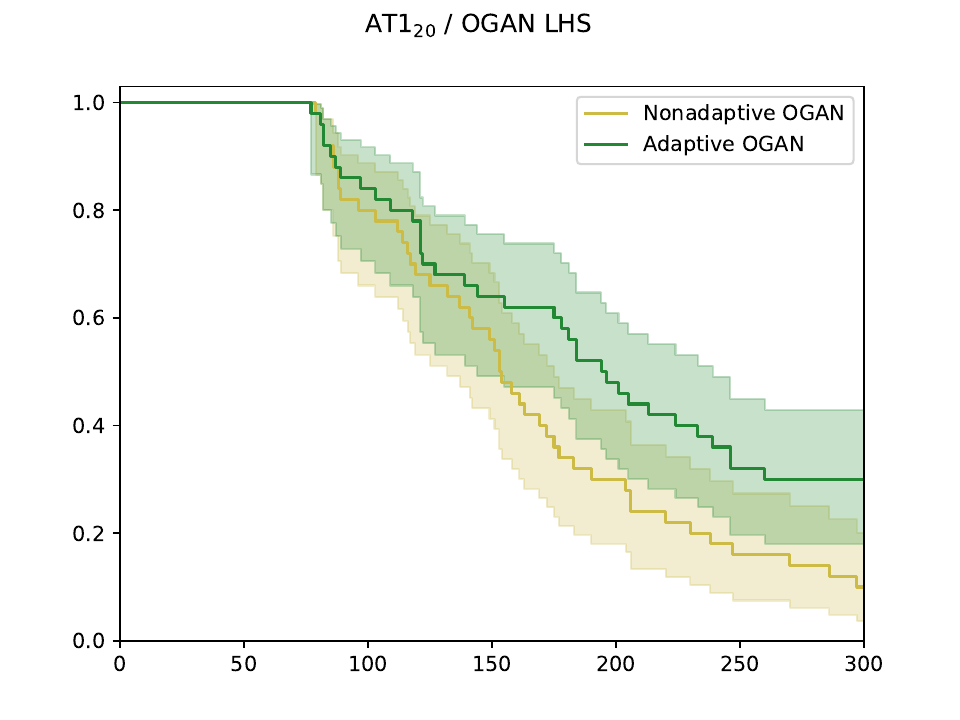}
\includegraphics[trim=35 5 35 0,clip,width=0.41\textwidth]{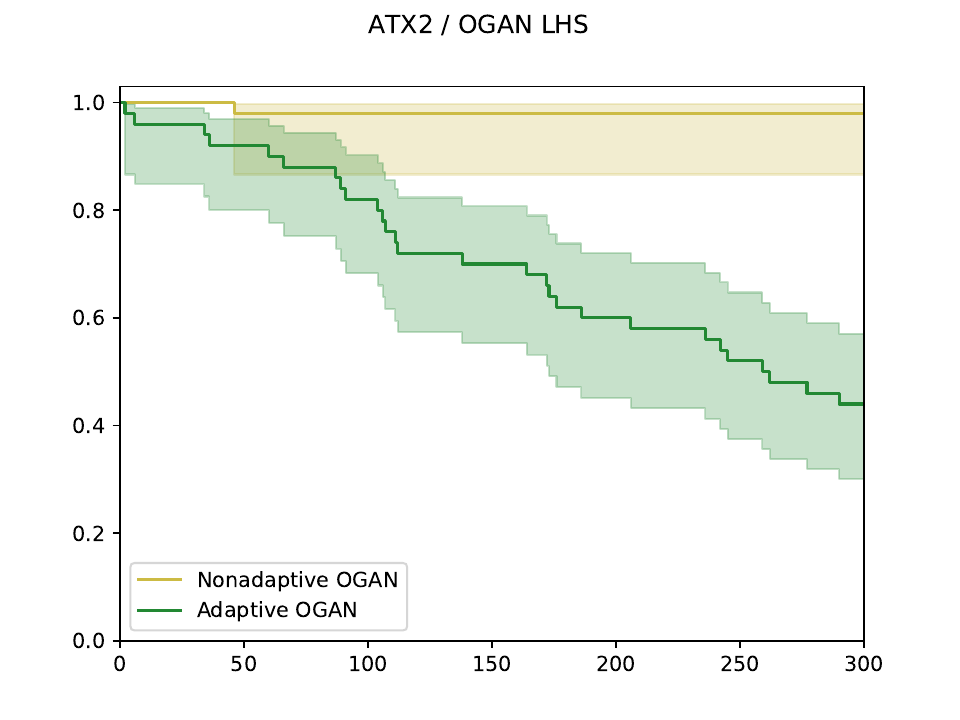}
\includegraphics[trim=35 5 35 0,clip,width=0.41\textwidth]{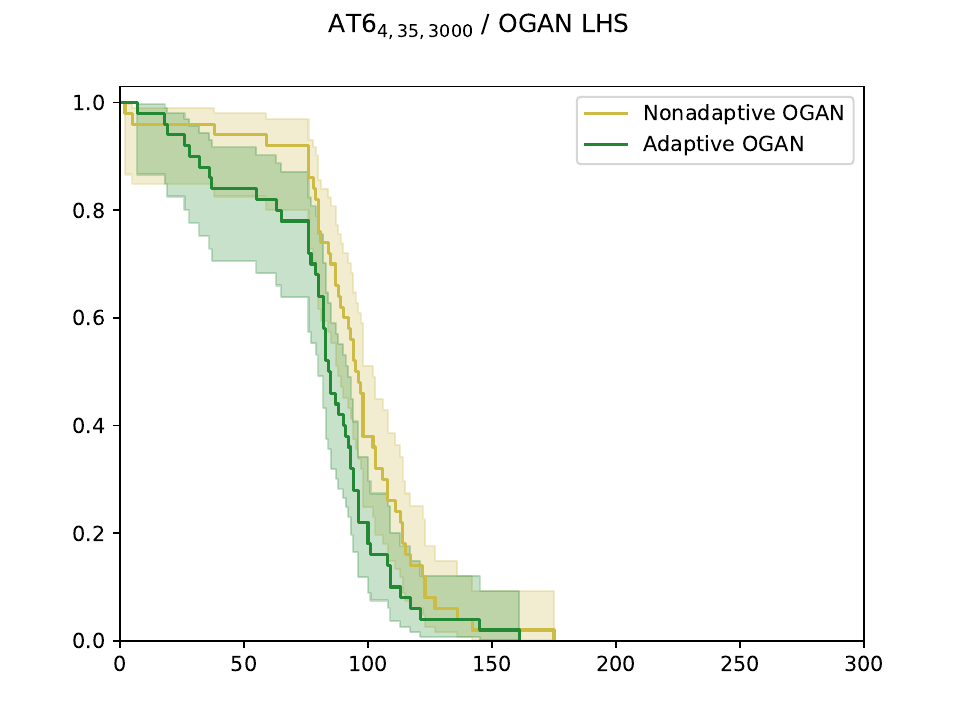}
\includegraphics[trim=35 5 35 0,clip,width=0.41\textwidth]{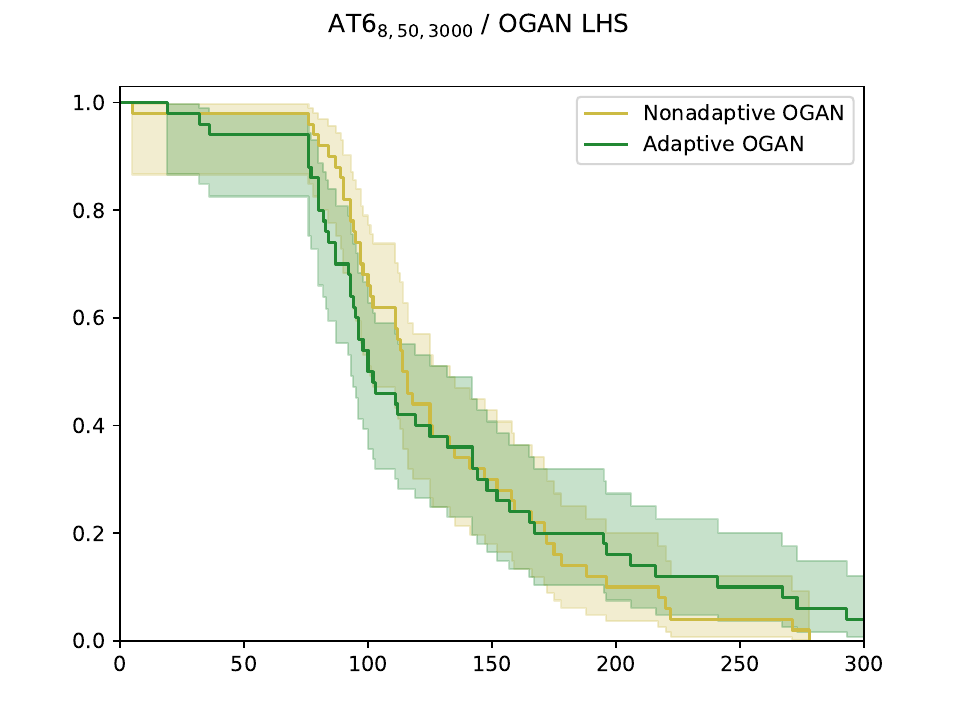}
\includegraphics[trim=35 5 35 0,clip,width=0.41\textwidth]{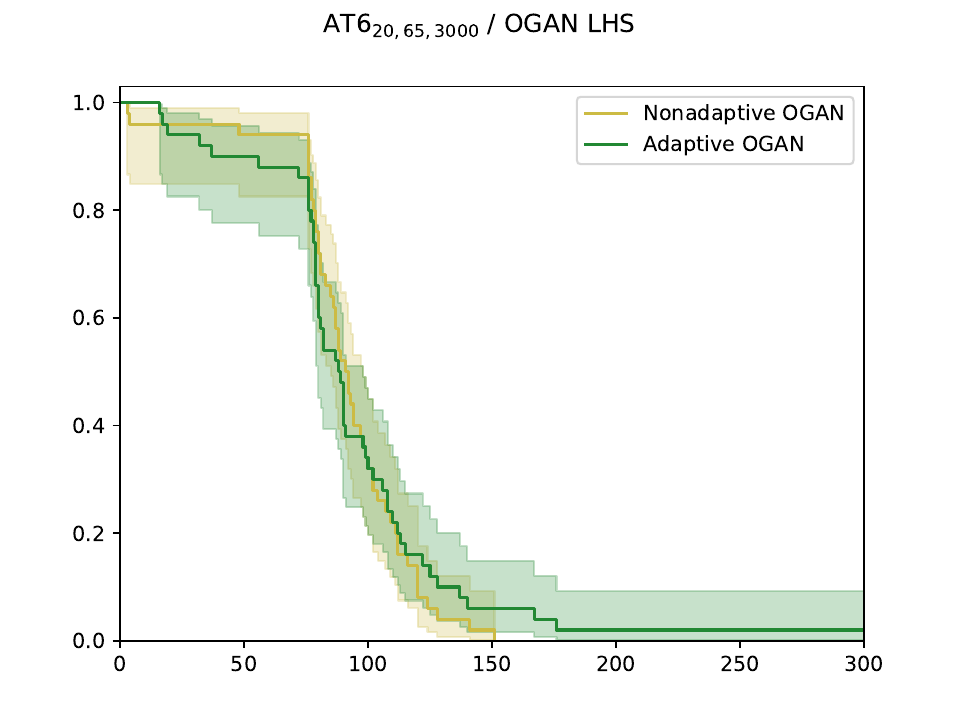}
\includegraphics[trim=35 5 35 0,clip,width=0.41\textwidth]{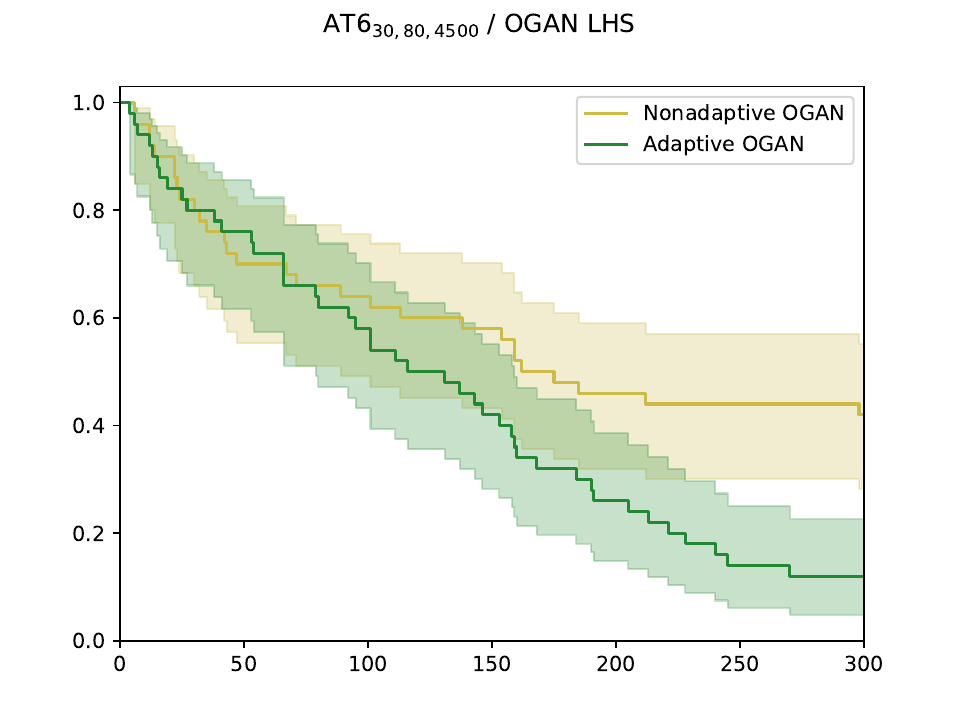}
\includegraphics[trim=35 5 35 0,clip,width=0.41\textwidth]{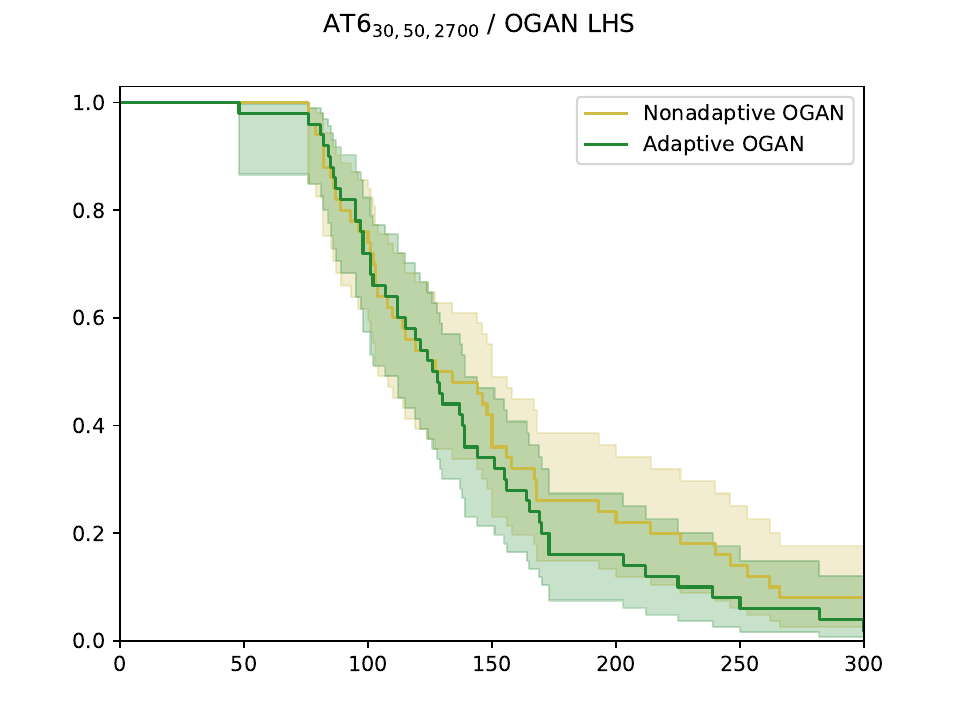}
\caption{Survival functions for adaptive OGAN and nonadaptive OGAN using uniform random sampling for the benchmarks of \autoref{ssec:benchmarks}.}\label{fig:survival_1}
\end{figure*}

\begin{figure*}
\centering
\includegraphics[trim=35 5 35 0,clip,width=0.41\textwidth]{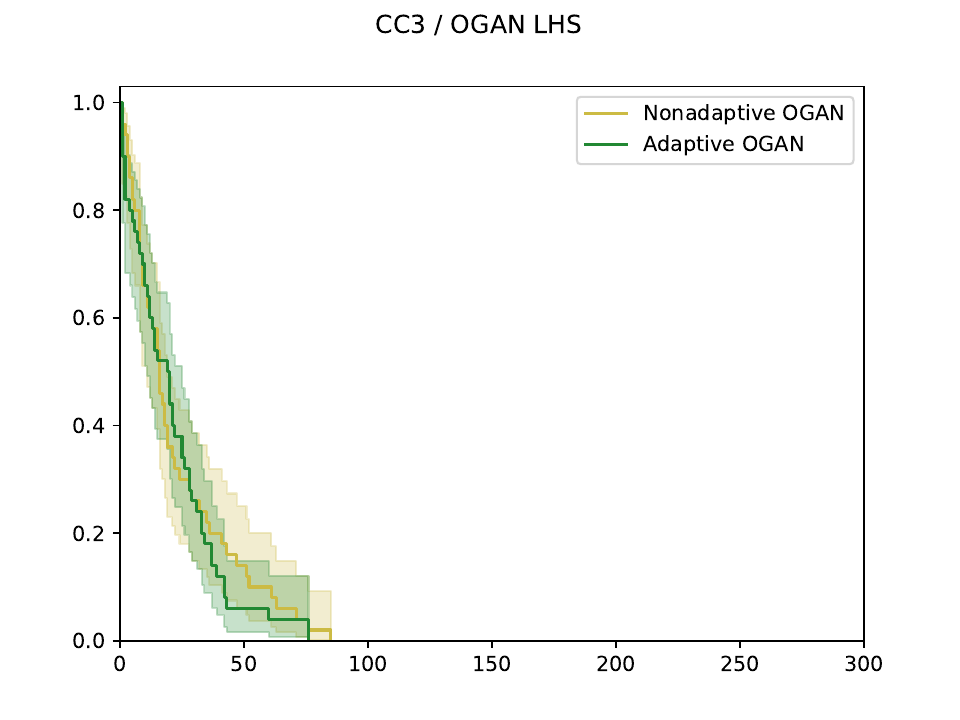}
\includegraphics[trim=35 5 35 0,clip,width=0.41\textwidth]{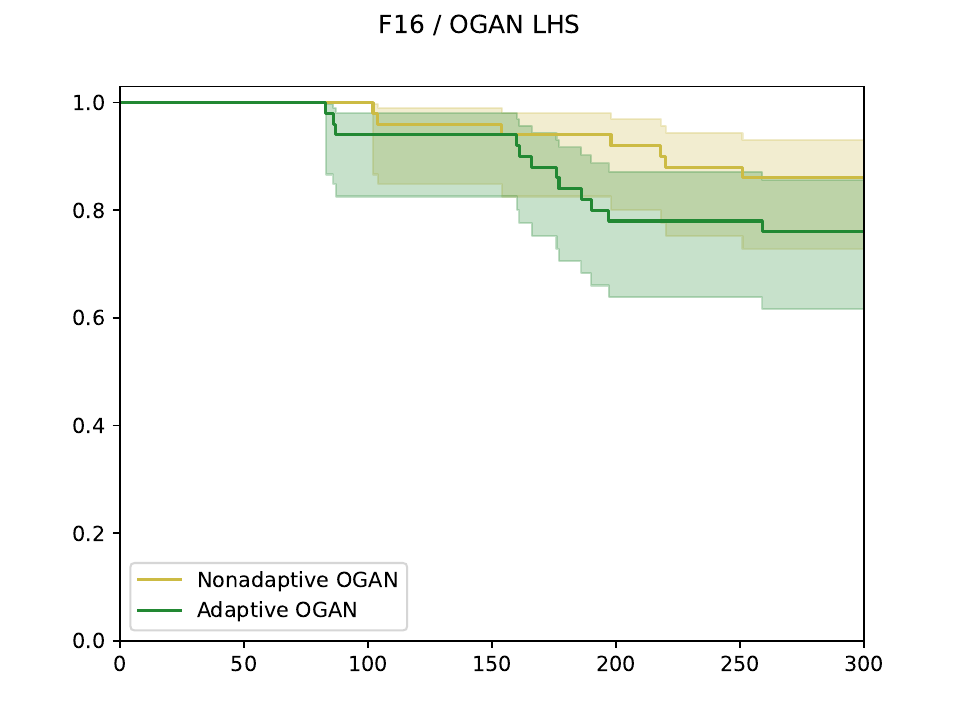}
\includegraphics[trim=35 5 35 0,clip,width=0.41\textwidth]{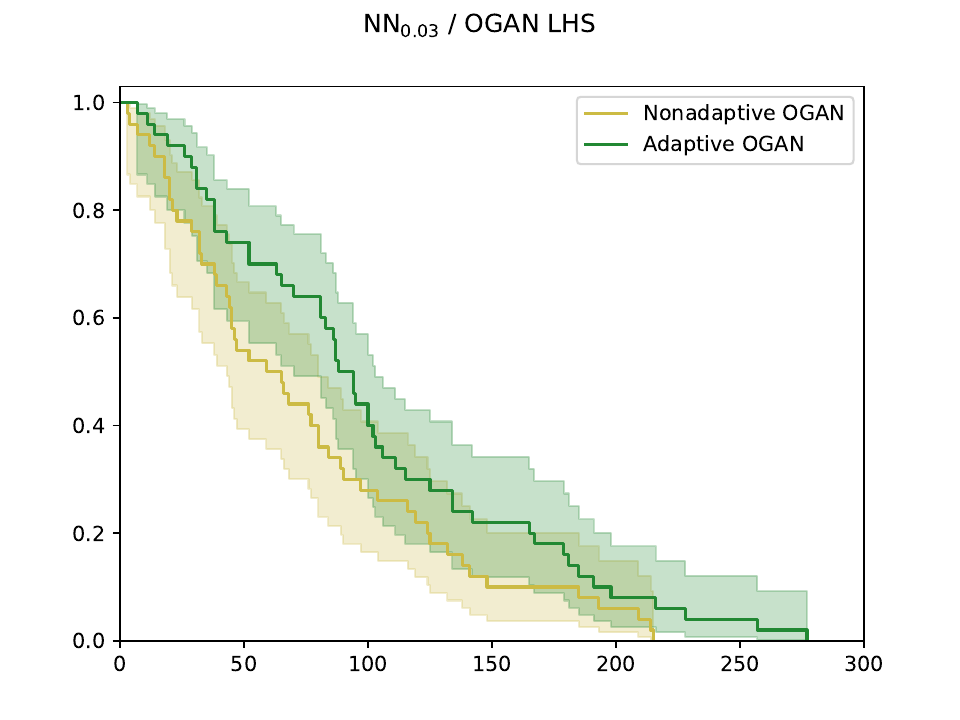}
\includegraphics[trim=35 5 35 0,clip,width=0.41\textwidth]{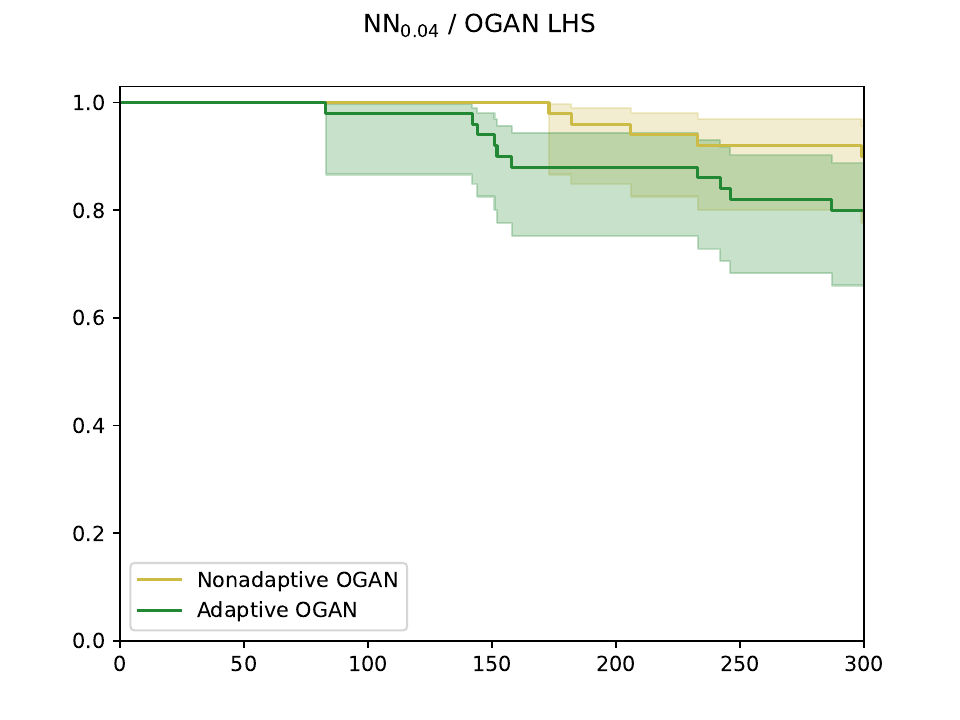}
\includegraphics[trim=35 5 35 0,clip,width=0.41\textwidth]{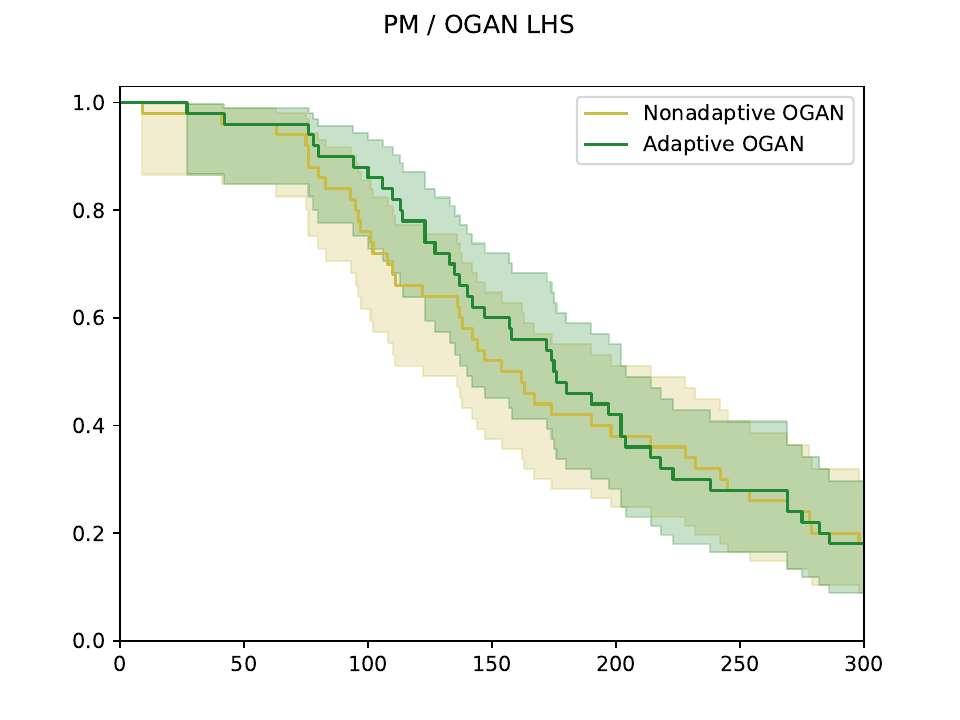}
\caption{\autoref{fig:survival_1} continued.}\label{fig:survival_2}
\end{figure*}

For the benchmarks of \autoref{ssec:benchmarks}, \autoref{tbl:survival} reports the FRs and their $95$ \% confidence intervals for adaptive OGAN and nonadaptive OGAN with both uniform random sampling (US) and LHS. Each statistic is based on $50$ replicas. Figures \ref{fig:survival_1}, \ref{fig:survival_2} depict the estimated survival functions for the two variants that use LHS. The survival functions are very similar for uniform random sampling, so we omit them. Again, the benchmark $\CCIV$ is omitted. \autoref{tbl:pvalues} reports the $p$-values of the log-rank test under the hypothesis that the number of executions required for finding a falsifying test for adaptive OGAN and nonadaptive OGAN are from the same distribution. Rejection of the hypothesis means that the two algorithms perform differently on a given benchmark. Notice that the falsification results of the adaptive OGAN match those described in \autoref{sssec:arch21_comparison}.

\begin{table}
\scalebox{0.93}{
\begin{tabular}{lcccccccc}
  \toprule
       & \multicolumn{2}{c}{Adaptive OGAN:}
       & \multicolumn{2}{c}{Nonadaptive OGAN:}
       & \multicolumn{2}{c}{Adaptive OGAN:}
       & \multicolumn{2}{c}{Nonadaptive OGAN:} \\
       & \multicolumn{2}{c}{US}
       & \multicolumn{2}{c}{US}
       & \multicolumn{2}{c}{LHS}
       & \multicolumn{2}{c}{LHS} \\
  \midrule
  \midrule
  Req. & \multicolumn{1}{c}{FR} & \multicolumn{1}{c}{$95$ \% CI} & \multicolumn{1}{c}{FR} & \multicolumn{1}{c}{$95$ \% CI} & \multicolumn{1}{c}{FR} & \multicolumn{1}{c}{$95$ \% CI} & \multicolumn{1}{c}{FR} & \multicolumn{1}{c}{$95$ \% CI} \\
  \cmidrule(r){1-1}
  \cmidrule(lr){2-3}
  \cmidrule(lr){4-5}
  \cmidrule(lr){6-7}
  \cmidrule(lr){8-9}
  $\textup{AFC27}$ & $1.00$ & $[1.00, 1.00]$ & $1.00$ & $[1.00, 1.00]$ & $1.00$ & $[1.00, 1.00]$ & $1.00$ & $[1.00, 1.00]$ \\
  \cmidrule(r){1-1}
  \cmidrule(lr){2-3}
  \cmidrule(lr){4-5}
  \cmidrule(lr){6-7}
  \cmidrule(lr){8-9}
  $\textup{AT1}_{20}$ & $0.82$ & $[0.70, 0.91]$ & $0.90$ & $[0.80, 0.96]$ & $0.70$ & $[0.57, 0.82]$ & $0.90$ & $[0.80, 0.96]$ \\
  $\textup{ATX2}$ & $0.48$ & $[0.35,0.62]$ & $0.04$ & $[0.01, 0.15]$ & $0.56$ & $[0.43, 0.70]$ & $0.02$ & $[0.00, 0.13]$ \\
  $\textup{AT6}_{4,35,3000}$ & $1.00$ & $[1.00,1.00]$ & $1.00$ & $[1.00, 1.00]$ & $1.00$ & $[1.00, 1.00]$ & $1.00$ & $[1.00, 1.00]$ \\
  $\textup{AT6}_{8,50,3000}$ & $1.00$ & $[1.00, 1.00]$ & $0.96$ & $[0.88, 0.99]$ & $0.96$ & $[0.88, 0.99]$ & $1.00$ & $[1.00, 1.00]$ \\
  $\textup{AT6}_{20,65,3000}$ & $0.96$ & $[0.88, 0.99]$ & $1.00$ & $[1.00, 1.00]$ & $0.98$ & $[0.91, 1.00]$ & $1.00$ & $[1.00, 1.00]$ \\
  $\textup{AT6}_{30,80,4500}$ & $0.88$ & $[0.77, 0.95]$ & $0.54$ & $[0.41, 0.68]$ & $0.88$ & $[0.77, 0.95]$ & $0.58$ & $[0.45, 0.72]$ \\
  $\textup{AT6}_{30,50,2700}$ & $0.92$ & $[0.82, 0.97]$ & $0.88$ & $[0.77, 0.95]$ & $0.98$ & $[0.91, 1.00]$ & $0.92$ & $[0.82, 0.97]$ \\
  \cmidrule(r){1-1}
  \cmidrule(lr){2-3}
  \cmidrule(lr){4-5}
  \cmidrule(lr){6-7}
  \cmidrule(lr){8-9}
  $\textup{CC3}$ & $1.00$ & $[1.00, 1.00]$ & $1.00$ & $[1.00, 1.00]$ & $1.00$ & $[1.00, 1.00]$ & $1.00$ & $[1.00, 1.00]$ \\
  $\textup{CC4}$ & $0.00$ & $[0.00, 0.00]$ & $0.00$ & $[0.00, 0.00]$ & $0.00$ & $[0.00, 0.00]$ & $0.00$ & $[0.00, 0.00]$ \\
  \cmidrule(r){1-1}
  \cmidrule(r){1-1}
  \cmidrule(lr){2-3}
  \cmidrule(lr){4-5}
  \cmidrule(lr){6-7}
  \cmidrule(lr){8-9}
  $\textup{F16}$ & $0.36$ & $[0.24, 0.51]$ & $0.20$ & $[0.11, 0.34]$ & $0.24$ & $[0.14, 0.38]$ & $0.14$ & $[0.07, 0.27]$ \\
  \cmidrule(r){1-1}
  \cmidrule(lr){2-3}
  \cmidrule(lr){4-5}
  \cmidrule(lr){6-7}
  \cmidrule(lr){8-9}
  $\textup{NN}_{0.03}$ & $0.98$ & $[0.91, 1.00]$ & $0.98$ & $[0.91, 1.00]$ & $1.00$ & $[1.00, 1.00]$ & $1.00$ & $[1.00, 1.00]$ \\
  $\textup{NN}_{0.04}$ & $0.24$ & $[0.14, 0.38]$ & $0.24$ & $[0.14, 0.38]$ & $0.20$ & $[0.11, 0.34]$ & $0.10$ & $[0.04, 0.22]$ \\
  \cmidrule(r){1-1}
  \cmidrule(lr){2-3}
  \cmidrule(lr){4-5}
  \cmidrule(lr){6-7}
  \cmidrule(lr){8-9}
  $\textup{PM}$ & $0.92$ & $[0.82, 0.97]$ & $0.80$ & $[0.68, 0.89]$ & $0.82$ & $[0.70, 0.91]$ & $0.82$ & $[0.70, 0.91]$
\end{tabular}
}
\caption{Falsification rates and their $95 \%$ confidence intervals for adaptive OGAN and nonadaptive OGAN for the benchmarks of \autoref{ssec:benchmarks}.}\label{tbl:survival}
\end{table}

From \autoref{tbl:survival}, we see that on the benchmarks $\AFC$, $\ATVIA$, $\ATVIB$, $\ATVIC$, $\ATXVIB$, $\CCIII$, and $\NN_{0.03}$ both adaptive OGAN and nonadaptive OGAN have full or almost full FR without much difference between the two variants. The corresponding $p$-values in \autoref{tbl:pvalues} are also high indicating no difference between the variants except in the case of $\ATVIA$ and $\NN_{0.03}$ with LHS sampling. The explanation is found from Figures \ref{fig:survival_1} and \ref{fig:survival_2}: one variant consistently falsifies faster than the other variant. The survival functions are thus different, but the difference in effectiveness is not large. If we accept the difference of FRs as an effect size, the effect size is $0.00$ for both benchmarks indicating no effect.

The adaptive variant is a clear winner on the benchmarks $\ATXII$, $\ATXVIA$, and $\FA$. On the first two benchmarks, the nonadaptive variant performs very poorly. In the case of $\FA$, the $p$-values do not indicate a statistically significant difference, but the $p$-values are nevertheless on the smaller side. The effect sizes $0.16$ (US) and $0.10$ (LHS) to the favor of the adaptive variant are also quite large. On the benchmarks $\NN_{0.04}$ and $\PM$, the nonadaptive variant suffers a big loss of effectiveness but only on one of the two sampling strategies. In the case of $\NN_{0.04}$, a part of the loss is perhaps explained by the fact that the adaptive variant with LHS is slightly worse than the adaptive variant with US. For $\PM$, we do not observe similar associations.

The only benchmark on which the adaptive variant loses to the nonadaptive variant is $\ATI$. The effect sizes are $-0.08$ (US) and $-0.20$ (LHS) and the corresponding $p$-values are $0.79$ and $0.02$ indicating a statistically significant difference only in the latter case.

Based on the above discussion, we answer RQ6 as follows.

\begin{itemize}
    \item[] \textbf{Answer to RQ6.} Augmenting the discriminator's training data by adaptively sampling the generator is a good feature. Disabling adaptive sampling can decrease effectiveness drastically even to the point of being unable to find falsifying inputs. Generally, adaptive sampling has at least as good effectiveness and efficiency as nonadaptive sampling.
\end{itemize}

\section{Related Work}\label{sec:related_work}

Robustness-based falsification is a widely studied problem in the research literature \cite{2018:specification_based_monitoring_of_cyber_physical,DBLP:journals/jair/CorsoMKLK21}. The ARCH-COMP friendly research competition \cite{ARCH21,ARCH22,ARCH23} comprises several CPS requirement falsification benchmarks that can be considered standard in this research area.

The tools and algorithms that participated in the ARCH-COMP competitions are briefly described below; for detailed information, see the competition reports \cite{ARCH21,ARCH23}. The majority of the entries to the ARCH-COMP competition can be considered as general black-box optimization algorithms, but the competition also includes approaches like reinforcement-learning-based falsification \cite{2021:falsification_of_cyber_physical_systems_using} and falsification using parametric surrogate models \cite{DBLP:conf/icse/MenghiNBP20}. 

The following tools, and OGAN, participated in ARCH COMP 2023.

\begin{itemize}
    \item \textbf{ARIsTEO} \cite{DBLP:conf/icse/MenghiNBP20}. This tool builds and refines a surrogate model. The surrogate is subjected to black-box testing using the S-TaLiRo falsification tool \cite{staliro-tool-paper} with default settings.
    \item \textbf{ATheNA} \cite{2023:search_based_software_testing_driven}. Under the hood, ATheNA uses S-TaLiRo and simulated annealing. The novelty of this tool is the use of custom robustness functions which combine the usual STL robustness function with a benchmark-specific robustness function provided by the user. Therefore, ATheNA makes use of some a priori knowledge about the SUT.
    \item \textbf{FalCAuN} \cite{2020:falsification_of_cyber_physical_systems_with_robustness}. This method actively learns an automaton representing the SUT and the requirement. Model checking is performed on the automaton to find a counterexample.
    \item \textbf{FORESEE} \cite{2021:effective_hybrid_system_falsification_using_monte}. FORESEE is built on top of the Breach falsification tool \cite{2010:breach_a_toolbox_for_verification_and_parameter_synthesis}, and it uses CMA-ES \cite{2016:the_cma_evolution_strategy_a_tutorial} as the optimization algorithm. The novelty of the tool is in the use QB-robustness as robustness metric.
    \item \textbf{NNFal}\footnote{\url{https://gitlab.com/Atanukundu/NNFal}}. The NNFal tool assumes that there exists a pretrained neural network model for the SUT. The falsification problem can then be approached as an adversarial attack on the network. See \cite{2021:reducing_dnn_properties_to_enable_falsification} for the adversarial attack approach on which NNFal ultimately relies.
    \item \textbf{$\Psi$-TaLiRo} \cite{2021:psy_taliro_a_python_toolbox_for_search}. $\Psi$-TaLiRo \cite{2021:psy_taliro_a_python_toolbox_for_search} is a Python implementation of S-TaLiRo \cite{staliro-tool-paper} with a different set of optimization algorithms. It supports the ConBO algorithm which is based on Bayesian optimization. It also supports the Part-X algorithm which estimates failure probabilities \cite{2021:part_x_a_family_of_stochastic_algorithms}. 
\end{itemize}

There are five tools that participated in ARCH-COMP 2021 that are not described earlier.
\begin{itemize}
    \item \textbf{Breach} \cite{2010:breach_a_toolbox_for_verification_and_parameter_synthesis}. Breach is a falsification pipeline supporting various algorithms for requirement falsification. Here, Breach has been used with the global Nelder-Mead algorithm (GNM) \cite{1965:a_simplex_method_for_function_minimization} and CMA-ES \cite{2016:the_cma_evolution_strategy_a_tutorial}. Both algorithms aim to minimize a robustness metric.
    \item \textbf{falsify} \cite{2021:falsification_of_cyber_physical_systems_using}. This tool trains in advance (using the A3C algorithm) a reinforcement learning agent that attempts to build a falsifying input piece-by-piece. This requires online test execution, i.e., the SUT must be observable and controllable during test execution. Moreover, the proposed method is applicable only to certain STL requirements of the form $\always \varphi$.
    \item \textbf{FALSTAR} \cite{2021:falsification_of_hybrid_systems_using_adaptive}. FALSTAR is a falsification pipeline. Here it is used with the adaptive Las Vegas tree search (aLVTS) algorithm. This algorithm is based on partitioning the search space into finer and finer parts based on certain probabilistic criteria. FALSTAR uses online test execution to build its inputs signals a test executes.
    \item \textbf{S-TaLiRo} \cite{staliro-tool-paper}. Like Breach and FALSTAR, S-TaLiRo is a falsification pipeline. In ARCH-COMP 2021, it was configured to use the SOAR optimization algorithm \cite{2021:stochastic_optimization_with_adaptive_restart_a_framework} which uses Bayesian optimization.
    \item \textbf{zlscheck}\footnote{\url{https://github.com/ismailbennani/zlscheck}}. The zlscheck tool assumes that a benchmark has been rewritten in the Z\'{e}lus programming language. This allows zlscheck to observe the system output during the simulation and to access the internals of the SUT. The robustness minimization and internal mode targeting are performed using gradient descent.
\end{itemize}

Most of the above falsification algorithms use a direct search and thus do not create a model for the robustness response of the SUT. The OGAN algorithm trains its discriminator as a surrogate model that is refined over time as more date is available. Methods based on Bayesian optimization \cite{2021:stochastic_optimization_with_adaptive_restart_a_framework,DBLP:journals/tecs/DeshmukhHJMP17} are the most similar to our approach as they use and refine Gaussian process surrogates. The ARIsTEO tool \cite{DBLP:conf/icse/MenghiNBP20} supports surrogates defined as Matlab AI models.

There are two other important design decisions for requirement falsification algorithms that can affect their applicability and performance. The first is whether or not the execution of a test is an atomic operation, and the second is whether or not knowledge collected before the start of the falsification task can be used.  A test is executed atomically if its input signals must be provided entirely before the test execution can start. In contrast, a requirement falsification algorithm using nonatomic testing can interact with the SUT while the test executes and generate the input signals incrementally base on the current system state. This is not possible when using atomic test execution. The precollected knowledge can include surrogate models, hand-crafted system monitors, or highly customized hyperparameters. \autoref{tbl:atomic} describes which tools use atomic test execution and which utilize precollected data or models.

The OGAN algorithm is inspired by generative adversarial networks (GANs) \cite{goodfellow2014generative}, but it is in fact quite different. A GAN has two models, a generator $G$ and a discriminator $D$. As in OGAN, the generator $G$ aims to synthesize samples from a target distribution $p_t$ by transforming noise sampled from a latent space. The discriminator $D$ is different as its task is to represent the probability that a sample is from $p_t$ rather than the generator's distribution $d_g$. In other words, $D$ is trained to distinguish between the real samples $p_t$ and the fake samples $d_g$. The models participate in an adversarial game where $G$ attempts to create samples that $D$ is unable to distinguish from target data (thus driving the samples of $G$ towards true data samples) and $D$ attempts to detect differences between true data and samples of $G$ (thus signaling $G$ about the discrepancies between the current samples and true data). The GAN training assumes that there exists a prior dataset of samples from $p_t$ The GAN training thus does not fit our purposes as we lack a dataset, and we need to generate the tests online. In the OGAN algorithm and in GANs, the discriminator guides the generator to sample the desired data. In GANs, the discriminator is trained against the generator but, in OGAN, the discriminator training is based on training data that is only indirectly related to the generator via the online training data augmentation.

To our knowledge, there is only one other requirement falsification algorithm that uses generative models: the WOGAN algorithm \cite{WOGAN,wogan_sbst22,wogan_sbft23}. WOGAN uses Wasserstein generative adversarial networks and,  in contrast to OGAN, is designed to find multiple and diverse counterexamples for a given requirement.

Finally, we should note that there are other generative models in addition to generative adversarial networks. Recently, diffusion models have become very popular especially in image generation \cite{2021:diffusion_models_beat_gans_on_image}. Diffusion models can also be used for black-box optimization \cite{2023:diffusion_models_for_black_box_optimization}, but we do not know of any attempt to use them for CPS requirement falsification.

\section{Conclusions}\label{sec:conclusions}
We have described in detail the OGAN algorithm for robustness-based falsification. OGAN can be used to find inputs that serve as counterexamples for the correctness of real-time cyber-physical systems. OGAN is based on the concept of offline test execution, and it can be used with systems that cannot be observed or controlled during the execution of a test.

OGAN trains a generative model using of two neural networks: a generator and a discriminator. The training is performed tabula rasa without knowledge of previous execution traces. At each iteration of the algorithm, the discriminator uses the results of previously executed tests to learn the mapping from inputs to robustness values while the generator learns to map noise to inputs that the discriminator estimates to have low robustness. The generator is then sampled for an input, and the selected input is used as a test against the actual SUT. It is expected that this process eventually generates a test with actual low robustness and a counterexample for the correctness of the SUT is found. To our knowledge, OGAN  is the first algorithm that uses generative machine learning models for requirement falsification.

In addition to presenting the OGAN algorithm, we have proposed a novel way to transform an STL robustness metric into a scaled robustness metric taking values in the interval $[0,1]$. Using a scaled robustness metric has the benefit that all neural network models used in the OGAN algorithm can be agnostic about the actual scales of the SUT inputs and outputs. We also believe that scaling contributes to the falsification performance as it is commonly accepted that scaling and normalization help to improve convergence in neural network training \cite{2012:efficient_backprop}.

We consider that the design of OGAN is sound. The study of its Monte Carlo sampling strategy shows that while OGAN requires a source of randomness for its exploration of the input space, OGAN's performance does not depend on the sampling method used (RQ3). Also, it is not necessary that the OGAN discriminator has high prediction accuracy; it is enough that the discriminator can direct the search towards falsifying inputs (RQ5). This indicates that OGAN is robust against minor modeling errors and prediction errors. This may be a further indication that OGAN can also perform well when the SUT is not completely deterministic. Another key element in the design of OGAN is its adaptive sampling method driven by its generator. We have designed an experiment that compares OGAN against a variant that collects training data using random sampling. We found out that the variant can perform very poorly compared to OGAN and that the adaptive sampling method is a good design choice without apparent drawbacks (RQ6).

We evaluated OGAN on several standard CPS benchmarks from the falsification track of the ARCH-COMP competition. For this, we proposed an evaluation method based on survival analysis. We believe that survival analysis has not been used for performance evaluation in the context of requirement falsification. Our opinion is that survival analysis is a useful tool as it provides both statistical tests and easily interpretable visual graphs.

In our evaluation we found that OGAN exhibits state-of-the-art CPS falsification efficiency and effectiveness (RQ1 and RQ2). OGAN has a good performance on many different benchmarks. However, due to the computational overhead introduced by OGAN (RQ4), OGAN is mainly intended for systems where testing has a significant cost (time or other resources).

\section*{Acknowledgments}
We thank the organizers and participants of the falsification track of the ARCH-COMP competitions for creating a standard set of CPS falsification benchmarks with publicly available data. We thank Valentin Soloviev for fruitful discussions regarding the experiment evaluation.

This research was supported by the ECSEL Joint Undertaking (JU) under grant agreement No 101007350 and by Business Finland under grant agreements AIDOaRT 42682/31/2020 and VST 7187/31/2023. The ECSEL JU received support from the European Union’s Horizon 2021 research and innovation program and Sweden, Austria, Czech Republic, Finland, France, Italy, Spain. 

\bibliographystyle{plainurl}
\bibliography{article}

\end{document}